\documentclass[10pt,twocolumn,letterpaper]{article}

\usepackage[pagenumbers]{cvpr}

\newcommand{\ci}[2]{\ensuremath{[#1, #2]}}
\newcommand{\nosign}[1]{\expandafter\nosignaux#1\relax}
\def\nosignaux#1#2\relax{\ifx+#1#2\else#1#2\fi}
\newcommand{\magn}[1]{\expandafter\magnaux#1\relax}
\def\magnaux#1#2\relax{\ifx-#1#2\else\ifx+#1#2\else#1#2\fi\fi}
\newcommand{\figbox}[3]{\fbox{\parbox[c][#2][c]{\dimexpr #1-2\fboxsep-2\fboxrule\relax}{\centering\footnotesize #3}}}
\newcommand{\figslot}[4]{\IfFileExists{#1}{\includegraphics[width=#2]{#1}}{\figbox{#2}{#3}{#4}}}
\newcommand{\pmstd}[1]{{\scriptsize\color{black!60}$\pm$#1}}
\newcommand{\msd}[2]{#1\pmstd{#2}}
\newcommand{\nostd}{\phantom{\pmstd{0.0}}}
\usepackage{pifont}
\newcommand{\cmark}{\ding{51}}
\newcommand{\xmark}{\ding{55}}
\usepackage{tikz}
\usetikzlibrary{arrows.meta,calc}
\definecolor{ourblue}{HTML}{0F4D92}\definecolor{midblue}{HTML}{3775BA}\definecolor{basered}{HTML}{B64342}\definecolor{lightred}{HTML}{E9A6A1}\definecolor{darkred}{HTML}{7A2B2A}\definecolor{refgray}{HTML}{8A8A8A}\usepackage{pgfplots}
\pgfplotsset{compat=1.17}
\tikzset{
  pics/budget/.style 2 args={code={
    \draw[#1!80!black, fill=white, line width=0.4pt, rounded corners=0.8pt] (-0.06cm,-0.10cm) rectangle (0.15cm,0.15cm);
    \draw[#1!80!black, fill=#1!30!white, line width=0.45pt, rounded corners=0.8pt] (-0.13cm,-0.15cm) rectangle (0.08cm,0.10cm);
    \node[inner sep=0pt, font=\sffamily\bfseries\fontsize{6}{6}\selectfont, text=#1!70!black] at (-0.025cm,-0.025cm) {#2};
  }},
  pics/phonetext/.style={code={
    \draw[black!60, fill=white, line width=0.5pt, rounded corners=1.5pt] (-0.15cm,-0.25cm) rectangle (0.15cm,0.25cm);
    \foreach \yy in {0.12,0.04,-0.04,-0.12} \draw[black!45, line width=0.6pt] (-0.09cm,\yy cm) -- (0.09cm,\yy cm);
  }},
  pics/phoneshot/.style={code={
    \draw[black!60, fill=white, line width=0.5pt, rounded corners=1.5pt] (-0.15cm,-0.25cm) rectangle (0.15cm,0.25cm);
    \fill[midblue] (-0.11cm,0.10cm) rectangle (0.11cm,0.20cm);
    \fill[black!25] (-0.11cm,-0.06cm) rectangle (-0.01cm,0.05cm) (0.01cm,-0.06cm) rectangle (0.11cm,0.05cm)
                    (-0.11cm,-0.20cm) rectangle (0.11cm,-0.09cm);
  }}}
\tikzset{plotfont/.style={font=\sffamily\fontsize{7}{8.2}\selectfont},
         plottick/.style={font=\sffamily\fontsize{6.5}{7.5}\selectfont}}
\pgfplotsset{paperplot/.style={
  font=\sffamily\fontsize{7}{8.2}\selectfont,
  every tick label/.append style={font=\sffamily\fontsize{6.5}{7.5}\selectfont},
  label style={font=\sffamily\fontsize{7}{8.2}\selectfont},
  axis x line*=bottom, axis y line*=left,
  axis line style={black!75, line width=0.45pt},
  every tick/.style={black!75, line width=0.45pt},
  major tick length=2pt, tick align=outside, clip=false,
  xticklabel={\pgfmathprintnumber[assume math mode=true]{\tick}},
  yticklabel={\pgfmathprintnumber[assume math mode=true]{\tick}},
  scale only axis}}
\newcommand{\fsign}[1]{\expandafter\fsignaux#1\relax}
\def\fsignaux#1#2\relax{\ifx-#1\textminus#2\else#1#2\fi}
\newcommand{\numnc}[1]{\expandafter\numncaux#1\relax}
\def\numncaux#1,#2\relax{#1#2}
\usepackage{colortbl}
\usepackage{tabularx}
\newcolumntype{L}{>{\raggedright\arraybackslash}X}\definecolor{upgreen}{HTML}{1E7B45}
\definecolor{rowblue}{HTML}{E6EDF7}
\newcommand{\fneg}[1]{\expandafter\fnegaux#1\relax}
\def\fnegaux#1#2\relax{\ifx+#1\textminus#2\else\ifx-#1+#2\else\textminus#1#2\fi\fi}
\newcommand{\dsub}[2]{\,\textsubscript{\color{#1}\bfseries #2}}
\newcommand{\dup}[1]{\dsub{upgreen}{\fsign{#1}}}
\newcommand{\ddown}[1]{\dsub{basered}{\fneg{#1}}}
\newcommand{\dgray}[1]{\dsub{black!62}{\fsign{#1}}}
\newcommand{\tabblock}[2]{\multicolumn{#1}{@{}l}{\textit{#2}}}
\newcommand{\vup}[1]{{\color{upgreen}\boldmath$#1$}}
\newcommand{\vdown}[1]{{\color{basered}\boldmath$#1$}}
\newcommand{\vgray}[1]{{\color{black!62}$#1$}}
\DeclareRobustCommand{\injyes}{\tikz[baseline=-0.6ex]\fill[basered] (0,0) circle[radius=1.6pt];}\DeclareRobustCommand{\injno}{\tikz[baseline=-0.6ex]\filldraw[draw=refgray, fill=white, line width=0.7pt] (0,0) circle[radius=1.5pt];}

 \usepackage{fontawesome5}

\newcommand{\costTextGpuSec}{3.9}            \newcommand{\costPixelGpuSec}{28.1}          \newcommand{\costRatioEpisode}{7.2}          \newcommand{\costRatioStep}{6.8}             \newcommand{\costReaderCalls}{11}                         \newcommand{\readerTemp}{0}                  \newcommand{\readerRepPenalty}{1.05}         

\newcommand{\budgetShots}{2,000}             \newcommand{\calibEpisodes}{168}             \newcommand{\calibShots}{2,004}              \newcommand{\alignTrainShots}{2,003}         \newcommand{\alignSkippedShots}{1}           \newcommand{\directRoundOneShots}{1,007}     \newcommand{\directRoundOnePassed}{41}       \newcommand{\directRoundTwoShots}{1,000}               

\newcommand{\alignLoraRank}{16}              \newcommand{\alignLR}{10^{-4}}               \newcommand{\alignEpochs}{2}                 \newcommand{\alignGpuMinutes}{93}            

\newcommand{\confTemplates}{29}              \newcommand{\confSeedFirst}{3000}            \newcommand{\confSeedLast}{3011}             \newcommand{\confInstances}{348}             \newcommand{\confEpisodes}{696}              \newcommand{\officialTestTemplates}{256}     \newcommand{\bootReps}{20,000}               \newcommand{\wordThreshold}{5}               

\newcommand{\succRealClean}{43.0}            \newcommand{\succTextClean}{87.5}            \newcommand{\gapClean}{+44.5}                \newcommand{\gapCleanLo}{31.2}               \newcommand{\gapCleanHi}{58.0}               \newcommand{\succRealGeneric}{49.6}          \newcommand{\succTextGeneric}{83.3}          \newcommand{\gapGeneric}{+33.8}              \newcommand{\gapGenericLo}{21.6}             \newcommand{\gapGenericHi}{45.8}             \newcommand{\succRealRate}{52.0}             \newcommand{\succTextRate}{82.6}             \newcommand{\gapRate}{+30.6}                 \newcommand{\gapRateLo}{19.5}                \newcommand{\gapRateHi}{42.2}                \newcommand{\succRealDirect}{55.6}           \newcommand{\succTextDirect}{86.6}           \newcommand{\gapDirect}{+31.0}               \newcommand{\gapDirectLo}{18.4}              \newcommand{\gapDirectHi}{44.0}              \newcommand{\succRealBoth}{55.9}             \newcommand{\succTextBoth}{85.3}             \newcommand{\gapBoth}{+29.5}                 \newcommand{\gapBothLo}{16.8}                \newcommand{\gapBothHi}{42.5}                \newcommand{\gapCriterion}{20}                                                                                                                 

\newcommand{\dDirectClean}{+12.6}            \newcommand{\dDirectCleanLo}{5.7}            \newcommand{\dDirectCleanHi}{20.4}                            \newcommand{\dBothClean}{+12.9}              \newcommand{\dBothCleanLo}{6.2}              \newcommand{\dBothCleanHi}{20.3}                                  \newcommand{\dRateGeneric}{+2.4}             \newcommand{\dRateGenericLo}{-3.6}           \newcommand{\dRateGenericHi}{9.3}                            \newcommand{\dRateDirect}{-3.6}              \newcommand{\dRateDirectLo}{-10.1}           \newcommand{\dRateDirectHi}{3.4}                               

\newcommand{\decompInstances}{93}            \newcommand{\decompSamples}{4}               \newcommand{\decompSuccClean}{88.7}          \newcommand{\decompSuccReaderOrder}{66.9}    \newcommand{\decompSuccReal}{35.5}           \newcommand{\decompSuccShuffle}{23.9}        \newcommand{\effOrderClean}{-21.8}                             \newcommand{\effMisreadLayout}{-47.0}                    \newcommand{\effMisreadReaderOrder}{-30.4}     \newcommand{\effOrderGivenMisread}{-5.1}           \newcommand{\effReaderVsShuffle}{+43.0}                

\newcommand{\persistSmall}{73.2}             \newcommand{\persistSmallIndep}{30.9}        \newcommand{\persistSmallDiff}{+42.3}        \newcommand{\persistSmallLo}{39.5}           \newcommand{\persistSmallHi}{45.8}           \newcommand{\persistLarge}{77.7}             \newcommand{\persistLargeIndep}{24.9}        \newcommand{\persistLargeDiff}{+52.8}        \newcommand{\persistLargeLo}{35.0}           \newcommand{\persistLargeHi}{65.3}           \newcommand{\persistSampled}{61.1}           \newcommand{\persistSampledIndep}{29.8}      \newcommand{\persistSampledDiff}{+31.3}      \newcommand{\persistSampledLo}{25.8}         \newcommand{\persistSampledHi}{37.9}         \newcommand{\persistOther}{89.7}             \newcommand{\persistOtherIndep}{77.6}        \newcommand{\persistOtherDiff}{+12.2}        \newcommand{\persistOtherLo}{10.1}           \newcommand{\persistOtherHi}{14.5}           \newcommand{\persistThreshold}{20}                          \newcommand{\sampledTemp}{0.7}                           \newcommand{\aucBase}{0.827}                 \newcommand{\aucHistory}{0.981}              \newcommand{\tempPersistPct}{84.6}           \newcommand{\rereadIdentical}{10.0}          \newcommand{\rereadBoxOnly}{59.1}            \newcommand{\rereadElemDiff}{30.7}           \newcommand{\rereadOrderOnly}{0.2}           

\newcommand{\fidReal}{0.895}                 \newcommand{\fidRealLo}{0.825}               \newcommand{\fidRealHi}{0.948}                            \newcommand{\fidReaderMax}{0.70}                                             \newcommand{\fidRandom}{0.483}               \newcommand{\fidOrderOnly}{0.388}            \newcommand{\fidClean}{0.249}                                        

\newcommand{\succAlignClean}{80.9}           \newcommand{\dMain}{+25.3}                   \newcommand{\dMainLo}{13.4}                  \newcommand{\dMainHi}{38.1}                                                 \newcommand{\dMainApprox}{25}                \newcommand{\dAlignClean}{+37.9}             \newcommand{\dAlignCleanLo}{25.4}            \newcommand{\dAlignCleanHi}{50.7}                              \newcommand{\gapClosedPct}{85}               \newcommand{\succRealStart}{40.5}            \newcommand{\succAlignStart}{75.9}           \newcommand{\dAlignStart}{+35.3}             \newcommand{\dAlignStartLo}{21.3}            \newcommand{\dAlignStartHi}{50.0}                              \newcommand{\succAlignRate}{76.0}            \newcommand{\dAlignRate}{+24.0}              \newcommand{\dAlignRateLo}{14.1}             \newcommand{\dAlignRateHi}{34.9}                                                               \newcommand{\succReorderClean}{48.3}         \newcommand{\dReorder}{+5.3}                 \newcommand{\dReorderLo}{-2.3}               \newcommand{\dReorderHi}{12.6}                                       \newcommand{\dCleanStartAligned}{+5.0}                 \newcommand{\dCleanRateAligned}{+4.9}                            \newcommand{\dRateStartAligned}{+0.1}        \newcommand{\tmplBetter}{18}                 \newcommand{\tmplSame}{6}                    \newcommand{\tmplWorse}{5}                   

\newcommand{\gateMinReductionPct}{30}        \newcommand{\gateMaxUnparsablePct}{1}        \newcommand{\gateShots}{1,007}               \newcommand{\errGateOrig}{25.62}             \newcommand{\errGateAligned}{1.56}           \newcommand{\errGateReductionPct}{93.9}      \newcommand{\gateUnparsableAligned}{0}       \newcommand{\gateNewShots}{341}              \newcommand{\errGateNewOrig}{27.03}          \newcommand{\errGateNewAligned}{2.92}        \newcommand{\otherTmplCount}{8}              \newcommand{\otherTmplShots}{204}            \newcommand{\errOtherOrig}{22.99}            \newcommand{\errOtherAligned}{1.53}          \newcommand{\otherTmplNewShots}{121}         \newcommand{\errOtherNewOrig}{22.77}         \newcommand{\errOtherNewAligned}{2.44}       \newcommand{\confScreensOrig}{7,266}         \newcommand{\confScreensAligned}{6,041}      \newcommand{\errConfOrig}{26.51}             \newcommand{\errConfAligned}{1.95}           \newcommand{\gateIdenticalPct}{64}           \newcommand{\errIdenticalAligned}{0.85}                \newcommand{\readerSecAligned}{23.4}         \newcommand{\readerSecOrig}{22.6}            \newcommand{\readerLongerPct}{11}            \newcommand{\truncAligned}{0}                \newcommand{\truncOrig}{183}                 

                                        \newcommand{\directSeedSpread}{12}           

\newcommand{\seedCleanA}{45.1}               \newcommand{\seedCleanB}{40.8}               \newcommand{\seedGenericA}{46.3}             \newcommand{\seedGenericB}{52.9}             \newcommand{\seedRateA}{51.7}                \newcommand{\seedRateB}{52.3}                \newcommand{\seedDirectA}{49.7}              \newcommand{\seedDirectB}{61.5}              \newcommand{\seedBothA}{55.5}                \newcommand{\seedBothB}{56.3}                \newcommand{\dRateClean}{+9.1}                                                              \newcommand{\dGenericClean}{+6.6}                                               \newcommand{\dBothDirect}{+0.3}                                                        \newcommand{\dBothRate}{+3.9}                                                                  \newcommand{\origSeedFirst}{1950}            \newcommand{\origSeedLast}{1955}             \newcommand{\origDirectClean}{+7.2}                           \newcommand{\origBothClean}{+8.9}                                 \newcommand{\origRateGeneric}{+2.6}                          \newcommand{\origRateDirect}{-2.9}                             \newcommand{\origGapMin}{29.9}               \newcommand{\origGapMax}{42.0}

\newcommand{\policyThinkTokens}{3,000}       \newcommand{\policyContextTokens}{16,384}    \newcommand{\policyHistActions}{8}           \newcommand{\policyHistNotes}{10}            \newcommand{\policyNoteChars}{100}           \newcommand{\policyTemp}{0.6}                \newcommand{\policyTopP}{0.95}               \newcommand{\policyTopK}{20}                 \newcommand{\policyRepeatStop}{6}            

\newcommand{\genericOmitMax}{20}             \newcommand{\genericCharErrMax}{5}           \newcommand{\genericJitterMax}{15}           \newcommand{\genericJitterScale}{1,000}      \newcommand{\genericFlipMax}{10}             \newcommand{\genericTruncMax}{10}            

\newcommand{\parallelWorkers}{12}

\newcommand{\failOrderFirst}{26}             \newcommand{\failFieldFirst}{21}             

\newcommand{\extraConfAligned}{0.55}         \newcommand{\extraConfOrig}{18.19}           \newcommand{\extraFromTargets}{60}           \newcommand{\extraFromTargetsPerScreens}{100} \newcommand{\extraGateOrig}{17.24}           \newcommand{\extraGateAligned}{0.87}         \newcommand{\extraNewOrig}{10.52}            \newcommand{\extraNewAligned}{1.74}          \newcommand{\extraInTargetsAligned}{29.6}    \newcommand{\extraInTargetsOrig}{9.5}        

\newcommand{\smallAlignShots}{500}           \newcommand{\succSmallAlignClean}{63.4}      \newcommand{\dSmallVsDirect}{+7.8}           \newcommand{\dSmallVsDirectLo}{-3.6}         \newcommand{\dSmallVsDirectHi}{19.8}                     \newcommand{\dFullVsSmall}{+17.5}                                                    \newcommand{\dSmallVsOrig}{+20.4}                                                    \newcommand{\errConfSmall}{7.68}             

\newcommand{\newErrMisreadPct}{93.2}         \newcommand{\misreadReadsPct}{36}            \newcommand{\sameWrongPct}{98.9}             \newcommand{\sameWrongPctRate}{45.6}         \newcommand{\failTotal}{47}                  

\newcommand{\succTextCleanStd}{1.4}          \newcommand{\succRealCleanStd}{3.0}                         \newcommand{\succTextGenericStd}{0.0}        \newcommand{\succRealGenericStd}{4.7}                     \newcommand{\succTextRateStd}{0.2}           \newcommand{\succRealRateStd}{0.4}                           \newcommand{\succTextDirectStd}{2.6}         \newcommand{\succRealDirectStd}{8.3}                       \newcommand{\succTextBothStd}{2.0}           \newcommand{\succRealBothStd}{0.6}                           \newcommand{\succAlignCleanStd}{2.2}         \newcommand{\succAlignRateStd}{3.9}          \newcommand{\succReorderCleanStd}{0.8}       \newcommand{\succSmallAlignCleanStd}{0.6}    \newcommand{\dAlignCleanStd}{5.3}            \newcommand{\dAlignRateStd}{4.3}             

\newcommand{\succLookupClean}{69.4}          \newcommand{\succLookupCleanStd}{1.8}        \newcommand{\dAlignLookup}{+11.5}            \newcommand{\dAlignLookupLo}{1.4}            \newcommand{\dAlignLookupHi}{22.8}                            \newcommand{\lookupGainPct}{70}              \newcommand{\lookupHitPct}{36.9}             \newcommand{\newScreenPixelPct}{63.1}        

\newcommand{\succDirectAligned}{78.2}        \newcommand{\succDirectAlignedStd}{0.4}      \newcommand{\dAlignDirect}{+22.6}                                                   \newcommand{\dDirectCleanAligned}{-2.7}                  \newcommand{\directAlignedShotsApprox}{4,000} \newcommand{\directAlignedDistinctApprox}{3,000} 

\newcommand{\origSuccAlignClean}{79.6}       \newcommand{\origSuccAlignCleanStd}{1.2}     \newcommand{\origSuccRealDirect}{52.0}       \newcommand{\origSuccRealDirectStd}{8.5}     \newcommand{\origSuccRealClean}{44.8}        \newcommand{\origMain}{+27.6}                \newcommand{\origMainLo}{14.7}               \newcommand{\origMainHi}{41.1}                                       \newcommand{\origAlignClean}{+34.8}                            \newcommand{\origTemplates}{29}              \newcommand{\origInstances}{174}             \newcommand{\origEpisodes}{348}              

\newcommand{\stepLimitLevelOne}{15}          \newcommand{\stepLimitLevelTwo}{30}          \newcommand{\stepLimitLevelThree}{45}        \newcommand{\stepLimitLevelFour}{60}         \newcommand{\stepLimitAnswerExtra}{15}       \newcommand{\stepLimitConfMin}{15}           \newcommand{\stepLimitConfMax}{75}           

\newcommand{\fidRate}{0.620}                 \newcommand{\fidReplayOne}{0.683}            \newcommand{\fidReplayTwo}{0.693}            \newcommand{\fidReplayTwoLayout}{0.699}      \newcommand{\fidCleanLo}{0.063}              \newcommand{\fidCleanHi}{0.472}              \newcommand{\fidOrderOnlyLo}{0.081}          \newcommand{\fidOrderOnlyHi}{0.654}          \newcommand{\fidRandomLo}{0.173}             \newcommand{\fidRandomHi}{0.743}             \newcommand{\fidRateLo}{0.357}               \newcommand{\fidRateHi}{0.832}               \newcommand{\fidReplayOneLo}{0.385}          \newcommand{\fidReplayOneHi}{0.905}          \newcommand{\fidReplayTwoLo}{0.412}          \newcommand{\fidReplayTwoHi}{0.904}          \newcommand{\fidReplayTwoLayoutLo}{0.415}    \newcommand{\fidReplayTwoLayoutHi}{0.898}    \newcommand{\fidSuccClean}{88.7}             \newcommand{\fidSuccOrderOnly}{66.9}         \newcommand{\fidSuccRandom}{44.6}            \newcommand{\fidSuccRate}{33.9}              \newcommand{\fidSuccReplayOne}{44.1}         \newcommand{\fidSuccReplayTwo}{36.6}         \newcommand{\fidSuccReplayTwoLayout}{41.7}   \newcommand{\fidPairClean}{+0.646}           \newcommand{\fidPairCleanLo}{0.404}          \newcommand{\fidPairCleanHi}{0.847}          \newcommand{\fidPairOrderOnly}{+0.507}       \newcommand{\fidPairOrderOnlyLo}{0.240}      \newcommand{\fidPairOrderOnlyHi}{0.819}      \newcommand{\fidPairRandom}{+0.412}          \newcommand{\fidPairRandomLo}{0.151}         \newcommand{\fidPairRandomHi}{0.724}         \newcommand{\fidPairRate}{+0.275}            \newcommand{\fidPairRateLo}{0.065}           \newcommand{\fidPairRateHi}{0.527}           \newcommand{\fidPairReplayOne}{+0.213}       \newcommand{\fidPairReplayOneLo}{0.003}      \newcommand{\fidPairReplayOneHi}{0.499}      \newcommand{\fidPairReplayTwo}{+0.202}       \newcommand{\fidPairReplayTwoLo}{0.011}      \newcommand{\fidPairReplayTwoHi}{0.454}      \newcommand{\fidPairReplayTwoLayout}{+0.196} \newcommand{\fidPairReplayTwoLayoutLo}{0.004} \newcommand{\fidPairReplayTwoLayoutHi}{0.461} 

\newcommand{\lookupScreens}{6,304}           

\newcommand{\succSameLabel}{67.2}            \newcommand{\succSameLabelStd}{1.2}          \newcommand{\dAlignSameLabel}{+13.6}         \newcommand{\dAlignSameLabelLo}{3.4}         \newcommand{\dAlignSameLabelHi}{25.1}                   \newcommand{\dSameLabelClean}{+24.3}                                   \newcommand{\dSameLabelDirect}{+11.6}                               \newcommand{\sameLabelDroppedPct}{12}        \newcommand{\sameLabelKeptA}{1,811}          \newcommand{\sameLabelKeptB}{1,834}          \newcommand{\sameLabelDroppedA}{193}         \newcommand{\sameLabelDroppedB}{170}         \newcommand{\sameLabelDroppedPctA}{12.3}     \newcommand{\sameLabelDroppedPctB}{11.5}     \newcommand{\sameLabelRowsA}{5,422}          \newcommand{\sameLabelRowsB}{5,468}          

\newcommand{\dLookupReorder}{+21.1}                                        

\newcommand{\heldGateShots}{284}             \newcommand{\heldErrOrig}{26.27}             \newcommand{\heldErrHeld}{7.59}              \newcommand{\heldErrFull}{1.04}              \newcommand{\heldErrReductionPct}{71.1}      \newcommand{\heldErrLinePct}{30}             \newcommand{\heldTemplates}{9}               \newcommand{\heldSuccHeld}{70.4}             \newcommand{\heldSuccOrig}{73.6}             \newcommand{\heldSuccFull}{94.0}                        \newcommand{\dHeldOrig}{-3.2}                \newcommand{\dHeldOrigLo}{-25.0}             \newcommand{\dHeldOrigHi}{15.3}                                   \newcommand{\dFullHeld}{+23.6}                                            \newcommand{\heldShotsNoWeChat}{1,437}       \newcommand{\heldShotsNoClock}{1,660}        \newcommand{\heldGateCross}{171}             \newcommand{\heldCrossErrOrig}{15.97}        \newcommand{\heldCrossErrHeld}{2.25}         \newcommand{\heldCrossErrFull}{0.24}         \newcommand{\heldCrossTemplates}{3}          

\newcommand{\fidRealTwo}{0.90}                                                                              

                                                                 \newcommand{\figTwoNoiseErr}{24}             \newcommand{\figTwoNoiseOnNever}{19}         

\newcommand{\heldCrossSuccOthers}{0.0}       \newcommand{\heldCrossSuccFullA}{50.0}       \newcommand{\heldCrossSuccFullB}{54.2}       \newcommand{\heldCrossSuccFullC}{0.0}        

\newcommand{\lookupBinLowEpisodes}{41}       \newcommand{\lookupBinMidEpisodes}{366}      \newcommand{\lookupBinHighEpisodes}{289}     \newcommand{\lookupBinLowSucc}{87.8}         \newcommand{\lookupBinMidSucc}{79.2}         \newcommand{\lookupBinHighSucc}{54.3}        \newcommand{\lookupJpegMismatch}{32}         

\newcommand{\succLargeClean}{46.4}           \newcommand{\succLargeCleanStd}{7.9}         \newcommand{\succLargeSeedA}{52.0}           \newcommand{\succLargeSeedB}{40.8}           \newcommand{\dAlignLarge}{+34.5}             \newcommand{\dAlignLargeLo}{21.4}            \newcommand{\dAlignLargeHi}{47.8}                              \newcommand{\dLargeDirect}{-9.2}                                                   \newcommand{\gapLarge}{+41.1}                                                                      \newcommand{\dLargeOrig}{+3.4}                                                             

\newcommand{\newScreenListPct}{48.6}         

\newcommand{\otherTmplTextSucc}{21.4}        \newcommand{\otherTmplThreshold}{50}         \newcommand{\otherTextHolidayLength}{54.2}   \newcommand{\otherTextAlipayBalance}{50.0}   \newcommand{\otherTextReadNote}{45.8}        \newcommand{\otherTextMeetingDurations}{20.8} \newcommand{\otherTextMonthlyIncome}{0.0}    \newcommand{\otherTextAlarmRepeat}{0.0}      \newcommand{\otherTextFastMeeting}{0.0}      \newcommand{\otherTextFriendConfirm}{0.0}    

\newcommand{\errGateLarge}{15.64}            

\newcommand{\curveErrOneHundred}{25.58}      \newcommand{\curveErrTwoFifty}{12.45}        \newcommand{\curveErrFiveHundred}{6.10}      \newcommand{\curveErrOneThousand}{2.87}      \newcommand{\curveNewErrOneHundred}{26.55}   \newcommand{\curveNewErrTwoFifty}{12.37}     \newcommand{\curveNewErrFiveHundred}{6.77}   \newcommand{\curveNewErrOneThousand}{4.39}   \newcommand{\curveNewErrFull}{2.74}          \newcommand{\curveNewFull}{366}              

\newcommand{\curveShotsOneHundred}{100}      \newcommand{\curveShotsTwoFifty}{250}        \newcommand{\curveShotsOneThousand}{1,000}   \newcommand{\curveStepsOneHundred}{13}       \newcommand{\alignStepsFull}{251}            

\newcommand{\errGateAlignSeedB}{1.52}        \newcommand{\errGateAlignSeedC}{1.65}        \newcommand{\newErrAlignSeedB}{2.40}         \newcommand{\newErrAlignSeedC}{2.91}         \newcommand{\typeMissSeedB}{257}             \newcommand{\typeTextSeedB}{300}             \newcommand{\typeStateSeedB}{30}             \newcommand{\extraGateSeedB}{0.47}           \newcommand{\typeMissSeedC}{377}             \newcommand{\typeTextSeedC}{259}             \newcommand{\typeStateSeedC}{1}              \newcommand{\extraGateSeedC}{0.53}           \newcommand{\succAlignSeedB}{76.3}           \newcommand{\succAlignSeedBStd}{3.5}         \newcommand{\dAlignSeedB}{-4.6}                                                        \newcommand{\dSeedBDirect}{+20.7}                                                    

\newcommand{\gateElements}{38,579}           \newcommand{\typeMissOrig}{8,514}            \newcommand{\typeTextOrig}{1,297}            \newcommand{\typeStateOrig}{74}              \newcommand{\typeMissLarge}{4,720}           \newcommand{\typeTextLarge}{1,237}           \newcommand{\typeStateLarge}{76}             \newcommand{\extraGateLarge}{4.33}           \newcommand{\typeMissAligned}{372}           \newcommand{\typeTextAligned}{225}           \newcommand{\typeStateAligned}{4}            

\newcommand{\baseTextSucc}{21.0}             \newcommand{\baseTextThreshold}{20}          \newcommand{\baseOverlongEpisodes}{8}        

\newcommand{\dropPctLow}{10}                 \newcommand{\dropPctHigh}{25}                \newcommand{\dropActualLow}{10.02}           \newcommand{\dropActualHigh}{25.07}          \newcommand{\calibTargetElements}{90,332}    

\newcommand{\directThreeShotsApprox}{3,000}  \newcommand{\directFourShotsApprox}{4,000}

\newcommand{\dSeedBLookup}{+6.9}                                                   \newcommand{\dSeedBSameLabel}{+9.1}                                                  \newcommand{\confApps}{9}                           \newcommand{\matchGrid}{10}                  \newcommand{\persistBoot}{1,000}             \newcommand{\persistApps}{12}                \newcommand{\persistAppsLarge}{11}           \newcommand{\fidBoot}{5,000}                 \newcommand{\fidLineM}{0.76}                 \newcommand{\fidLinePoints}{5}               \newcommand{\fidReplayTwoDiff}{+1.1}         \newcommand{\fidFirstLineCorr}{0.64}         \newcommand{\fidFirstLinePoints}{5}          \newcommand{\fidFirstCorr}{0.584}            \newcommand{\fidFirstDiff}{+4.1}                                   \newcommand{\fidFirstEpisodes}{232}          \newcommand{\fidFirstRealEpisodes}{261}      

\newcommand{\tmplSendMsgText}{0.0}           \newcommand{\tmplDateIntervalText}{8.3}      

\newcommand{\readerMaxTokens}{4,096}         

\newcommand{\textviewRowPx}{6}               \newcommand{\leakTestsOne}{12}               \newcommand{\leakTestsTwo}{20}               \newcommand{\leakTestsThree}{24}             \newcommand{\pixAuditScreens}{240}           \newcommand{\pixAuditElements}{10,571}       \newcommand{\pixAuditUnchanged}{10}          \newcommand{\pixAuditUnchangedPct}{0.095}    \newcommand{\pixAuditFaint}{79}              \newcommand{\pixAuditFaintPct}{0.75}         \newcommand{\pixAuditRecheckScreens}{240}    \newcommand{\pixAuditRecheck}{3}             \newcommand{\pixAuditRecheckPct}{0.027}      \newcommand{\manualCheckScreens}{30}         \newcommand{\manualAuthorScreens}{15}        

\newcommand{\rateMinObs}{20}                 \newcommand{\rateMinObsState}{10}            \newcommand{\rateFitSplit}{80/20}            \newcommand{\rateFitPoints}{5}               

\newcommand{\policyLoraRank}{16}             \newcommand{\policyLoraAlpha}{32}            \newcommand{\policyLoraDropout}{0.05}        \newcommand{\policyLR}{10^{-4}}              \newcommand{\policyWarmupPct}{3}             \newcommand{\policyGradClip}{1.0}            \newcommand{\policyEpochs}{2}                \newcommand{\policyGradAccum}{16}            \newcommand{\policyMaxSampleTokens}{10,240}  \newcommand{\demoEpisodes}{232}              \newcommand{\demoPerTemplate}{8}             \newcommand{\demoPassed}{221}                \newcommand{\demoSamples}{1,806}             \newcommand{\demoTemplates}{28}              \newcommand{\practiceSamplesPerTemplate}{4}  \newcommand{\practiceEpisodes}{116}          \newcommand{\directRoundOneSamples}{325}     \newcommand{\directRoundTwoSamplesPerTemplate}{6} 

\newcommand{\readerPickLossLine}{15}         \newcommand{\readerPickLowLine}{3}           \newcommand{\readerPickSuccLine}{50}         \newcommand{\readerPickCostLine}{0.50}       \newcommand{\readerPickBoot}{4,000}          \newcommand{\readerPickLargeLoss}{13.9}                     \newcommand{\readerPickLargeEpisodes}{72}               \newcommand{\readerPickCheapLoss}{24.0}              \newcommand{\readerPickCheapSucc}{76.0}      \newcommand{\readerPickCheapEpisodes}{50}               \newcommand{\readerPickCostA}{0.491}         \newcommand{\readerPickCostB}{0.483}         \newcommand{\readerPickCostC}{0.499}         \newcommand{\readerPickCostMean}{0.491}            \newcommand{\readerPickInflight}{8}          \newcommand{\readerPickShots}{30}            \newcommand{\readerPickCandidates}{5}        \newcommand{\readerPickSingleEightFull}{24.7}   \newcommand{\readerPickSingleFourHalf}{13.4}   

\newcommand{\policyDroppedTooLong}{1}        \newcommand{\policyLoraTargets}{7}           \newcommand{\pixAuditRecheckFaint}{200}      \newcommand{\pixAuditRecheckFaintPct}{1.81}  \newcommand{\manualCheckRounds}{2}           

\newcommand{\readerPickCheapSameLoss}{30.6}      

\newcommand{\readerOrderDiffPct}{76.6}       \newcommand{\replayFixedProb}{0.72}          

\newcommand{\callsCollection}{3,011}         \newcommand{\callsTraining}{14,826}          \newcommand{\callsDevelopment}{23,026}       \newcommand{\callsEvaluation}{187,887}       \newcommand{\callsAudit}{18,692}             \newcommand{\callsTotal}{247,442}            

\newcommand{\succAlignBase}{18.4}            \newcommand{\succRealBase}{23.0}             \newcommand{\dAlignBase}{-4.6}                                                             

\newcommand{\baseTextTemplates}{14}          \newcommand{\baseSubsetCleanGap}{43.2}       \newcommand{\baseOtherCleanGap}{45.8}        \newcommand{\baseWeightedCleanGap}{34.6}     \newcommand{\baseLost}{32}                   \newcommand{\baseGained}{39}                 \newcommand{\cleanLostA}{150}                \newcommand{\cleanLostB}{172}                \newcommand{\cleanGainedA}{6}                \newcommand{\cleanGainedB}{6}                

\newcommand{\readerPickRescoreDrop}{12.5}    \newcommand{\readerPickLargeLossAsRun}{19.4} \newcommand{\readerPickLargeLossPrev}{26.4}  \newcommand{\readerPickCheapSameLossAsRun}{26.4} \newcommand{\readerPickCheapSameLossPrev}{43.1} 

\newcommand{\succLargeFullClean}{48.9}       \newcommand{\succLargeFullCleanStd}{0.8}     \newcommand{\dAlignLargeFull}{+32.0}         \newcommand{\dAlignLargeFullLo}{18.0}        \newcommand{\dAlignLargeFullHi}{46.8}                  \newcommand{\gapLargeFull}{+38.6}            \newcommand{\gapLargeFullLo}{24.9}           \newcommand{\gapLargeFullHi}{53.3}                           

\newcommand{\costRatioReadLargeFull}{0.52}   \newcommand{\costRatioEpisodeLargeFull}{0.48}  

\newcommand{\readerPickFirstEpisodes}{52}    

\newcommand{\errGateDropLow}{4.34}           \newcommand{\errGateDropHigh}{12.31}         

\newcommand{\succLargeFullSeedA}{49.4}       \newcommand{\succLargeFullSeedB}{48.3}       

\newcommand{\errGateLargeFull}{17.45}        \newcommand{\typeMissLargeFull}{5,413}       \newcommand{\typeTextLargeFull}{1,228}       \newcommand{\typeStateLargeFull}{90}         \newcommand{\extraGateLargeFull}{14.92}      

\newcommand{\dLargeFullOrig}{+5.9}                                         \newcommand{\dLargeFullDirect}{-6.8}                               \newcommand{\dLargeFullHalf}{+2.4}                                          

\newcommand{\costReadInflight}{12}           \newcommand{\costReadAlignedSec}{1.58}       \newcommand{\costReadLargeFullSec}{3.05}     \newcommand{\costPolicyCallSec}{0.25}        \newcommand{\costStepsAligned}{8.68}         \newcommand{\costStepsLargeFull}{10.12}      

\newcommand{\typeMissDropLow}{1,391}         \newcommand{\typeTextDropLow}{284}           \newcommand{\typeStateDropLow}{0}            \newcommand{\extraGateDropLow}{0.50}         \newcommand{\typeMissDropHigh}{4,461}        \newcommand{\typeTextDropHigh}{287}          \newcommand{\typeStateDropHigh}{0}           \newcommand{\extraGateDropHigh}{0.41}        

\newcommand{\succDirectThree}{59.2}          \newcommand{\succDirectThreeStd}{9.3}        \newcommand{\succTextDirectThree}{86.9}      \newcommand{\dAlignDirectThree}{+21.7}                          

\newcommand{\succDirectFour}{63.6}           \newcommand{\succDirectFourStd}{4.3}         \newcommand{\succTextDirectFour}{85.3}       \newcommand{\dDirectFourVsTwo}{+8.0}                                 \newcommand{\dAlignDirectFour}{+17.2}                                

\newcommand{\freshSuccAlignClean}{81.5}      \newcommand{\freshSuccAlignCleanSeedA}{79.3} \newcommand{\freshSuccAlignCleanSeedB}{83.6} \newcommand{\freshSuccAlignCleanStd}{3.0}    \newcommand{\freshSuccRealDirect}{54.9}      \newcommand{\freshSuccRealDirectSeedA}{48.3} \newcommand{\freshSuccRealDirectSeedB}{61.5} \newcommand{\freshSuccRealDirectStd}{9.3}    \newcommand{\freshMain}{+26.6}               \newcommand{\freshMainLo}{13.1}              \newcommand{\freshMainHi}{40.4}                                     \newcommand{\freshTemplates}{29}             \newcommand{\freshInstances}{348}            \newcommand{\freshReads}{13,600}             

\newcommand{\tdEpisodes}{6,960}              \newcommand{\tdErrors}{28}                   \newcommand{\tdScreens}{31,652}              \newcommand{\tdOrderSameScreenPct}{45.8}     \newcommand{\tdOrderSimilarPct}{44.7}        \newcommand{\tdOrderTopDownPct}{9.4}         \newcommand{\tdOrderMin}{11.4}               \newcommand{\tdOrderMax}{20.1}               \newcommand{\tdMisreadMin}{26.4}             \newcommand{\tdMisreadMax}{32.9}             \newcommand{\tdRestMin}{-14.9}               \newcommand{\tdRestMax}{-8.2}                                            \newcommand{\tdOrderDirectClean}{-0.6}                 

\newcommand{\succDropHigh}{22.1}             \newcommand{\succDropHighSeedA}{21.8}        \newcommand{\succDropHighSeedB}{22.4}        \newcommand{\readsDropHigh}{7,482}           \newcommand{\dDropHighOrig}{-20.8}                                           \newcommand{\dFullDropHigh}{+58.8}                                             \newcommand{\mechTaskElements}{137}          \newcommand{\mechTaskOccurrences}{734}       \newcommand{\mechOrigAllMissed}{22.1}        \newcommand{\mechOrigTaskMissed}{6.4}        \newcommand{\mechHighAllMissed}{11.6}        \newcommand{\mechHighTaskMissed}{22.5}       \newcommand{\mechLowAllMissed}{3.6}          \newcommand{\mechLowTaskMissed}{7.8}         \newcommand{\mechFullAllMissed}{1.0}         \newcommand{\mechFullTaskMissed}{0.7}        \newcommand{\mechPersistSeen}{95}            \newcommand{\mechHighAlways}{21}             \newcommand{\mechOrigAlways}{5}              \newcommand{\mechFullAlways}{1}              \newcommand{\mechHighStuck}{98}              \newcommand{\mechOrigStuck}{87}              \newcommand{\mechHighIdentical}{6}           \newcommand{\mechHighNeededMissed}{6}        \newcommand{\mechExampleStuck}{5}            

\newcommand{\viewportW}{360}                 \newcommand{\viewportH}{800}                 \newcommand{\visOpacityMin}{0.05}            \newcommand{\visMinSide}{3}                  

\newcommand{\visDotsMin}{4}                  \newcommand{\visContrastMin}{1.08}           

\newcommand{\awShardsUsed}{4}                                                                                                                             \newcommand{\awAlignShots}{2,001}                       \newcommand{\awAlignApps}{140}                             \newcommand{\awHeldShots}{1,000}                         \newcommand{\awHeldApps}{100}                                \newcommand{\awHeldNewShots}{967}            \newcommand{\awAppsBoth}{87}                 \newcommand{\awCallsTotal}{4,021}            \newcommand{\awCallsSmoke}{20}               \newcommand{\awCallsOriginal}{3,001}         \newcommand{\awCallsAligned}{1,000}          \newcommand{\awLineRecurPoints}{20}          \newcommand{\awLineDropPct}{30}              \newcommand{\awLineUnparsablePct}{1}         \newcommand{\awRecurReader}{58.3}            \newcommand{\awRecurIndep}{34.8}             \newcommand{\awRecurDiff}{+23.5}             \newcommand{\awRecurLo}{18.0}                \newcommand{\awRecurHi}{29.8}                                                                                                                        \newcommand{\awErrOrig}{36.68}               \newcommand{\awErrAligned}{27.45}            \newcommand{\awErrDropPct}{25.2}                         \newcommand{\awUnparsableAligned}{0}                                                     \newcommand{\awErrNewOrig}{36.77}            \newcommand{\awErrNewAligned}{27.50}                   \newcommand{\awTrainMinutes}{77}                                     \newcommand{\awTrainLoss}{0.252}             \newcommand{\awPaperTrainLoss}{0.018}        \newcommand{\awUnseenApps}{13}                             \newcommand{\awUnseenOrig}{26.5}             \newcommand{\awUnseenAligned}{27.2}                                    \newcommand{\awManyApps}{14}                 \newcommand{\awManyOrig}{42.3}               \newcommand{\awManyAligned}{26.8}                                 \newcommand{\awSeenElDropPct}{44.6}          \newcommand{\awUnseenElDropPct}{19.9}                                                                                                                                            \newcommand{\awSpotShots}{30}                \newcommand{\awSpotElements}{87}                            \newcommand{\awSpotMissedPct}{52}                                                    \newcommand{\awBoxShareOrigPct}{63}                                      \newcommand{\awBoxPointsOrig}{10.2}          \newcommand{\awBoxPointsAligned}{7.6}        \newcommand{\awBoxDropPct}{25.3}                                \newcommand{\awBoxRecurDiff}{+20.3}                                    

\newcommand{\slPixCheckShots}{6,304}         \newcommand{\slPixCheckMissed}{32}                                    \newcommand{\slPixNewPct}{52.1}              \newcommand{\slPixLowEpisodes}{155}                              \newcommand{\slPixLowAlign}{94.2}            \newcommand{\slPixLowDirect}{87.7}           \newcommand{\slPixLowCleanOrig}{63.9}        \newcommand{\slPixLowLookup}{89.7}           \newcommand{\slPixLowGap}{+6.5}                                                       \newcommand{\slPixMidEpisodes}{347}                             \newcommand{\slPixMidAlign}{77.8}            \newcommand{\slPixMidDirect}{50.1}           \newcommand{\slPixMidCleanOrig}{40.9}        \newcommand{\slPixMidLookup}{59.7}           \newcommand{\slPixMidGap}{+27.7}                                                       \newcommand{\slPixHighEpisodes}{194}                          \newcommand{\slPixHighAlign}{75.8}           \newcommand{\slPixHighDirect}{39.7}          \newcommand{\slPixHighCleanOrig}{29.9}       \newcommand{\slPixHighLookup}{70.6}          \newcommand{\slPixHighGap}{+36.1}                                                   \newcommand{\slFailCount}{133}                             \newcommand{\slFailOrder}{26}                              \newcommand{\slFailPolicy}{66}                                 \newcommand{\slFailReaderSide}{67}                           \newcommand{\slFailActed}{40}                \newcommand{\slFailNeeded}{1}                                                                                                                                      \newcommand{\slDedupRows}{2,004}             \newcommand{\slDedupRemoved}{1,685}          \newcommand{\slDedupRemovedPct}{84.1}        \newcommand{\slDedupKept}{319}                                          \newcommand{\slDedupSucc}{56.2}                                        \newcommand{\slDedupSuccStd}{1.8}                           \newcommand{\slDedupReads}{7,338}            \newcommand{\slDedupCardHours}{6.5}          \newcommand{\slDedupGate}{12.18}             \newcommand{\slDedupGateNew}{12.08}                 \newcommand{\slDedupGateReads}{1,007}        \newcommand{\slDedupTrainUsed}{318}                           \newcommand{\slDedupResidShots}{7,338}                  \newcommand{\slDedupResidKeptPct}{0.14}               \newcommand{\slDedupResidAllPct}{40.3}       \newcommand{\slDedupVsDirect}{+0.6}          \newcommand{\slDedupVsDirectLo}{-12.6}       \newcommand{\slDedupVsDirectHi}{13.6}                 \newcommand{\slDedupVsLookup}{-13.2}                                   \newcommand{\slDedupVsAlign}{-24.7}                                    \newcommand{\slLargeSucc}{76.0}                                        \newcommand{\slLargeSuccStd}{2.6}                           \newcommand{\slLargeReads}{6,045}            \newcommand{\slLargeCardHours}{13.2}         \newcommand{\slLargeGate}{1.46}                                \newcommand{\slLargeGateReads}{1,007}        \newcommand{\slLargeTrainUsed}{2,003}                        \newcommand{\slLargeVsAlign}{-4.9}           \newcommand{\slLargeVsAlignLo}{-12.1}        \newcommand{\slLargeVsAlignHi}{0.4}                      \newcommand{\slLargeVsDirect}{+20.4}         \newcommand{\slLargeVsDirectLo}{8.9}         \newcommand{\slLargeVsDirectHi}{32.9}                   \newcommand{\slSeedThreeSucc}{73.9}                            \newcommand{\slSeedThreeSuccStd}{2.0}                   \newcommand{\slSeedThreeReads}{6,164}        \newcommand{\slSeedThreeCardHours}{4.4}      \newcommand{\slSeedMeanThree}{77.0}          \newcommand{\slReadsTotal}{21,561}           \newcommand{\slSeedThreeVsDirect}{+18.2}                 \newcommand{\slSeedThreeVsAlign}{-7.0}

\definecolor{cvprblue}{rgb}{0.21,0.49,0.74}
\usepackage[breaklinks,colorlinks,allcolors=cvprblue]{hyperref}

\def\confName{}
\def\confYear{}

\newcommand{\arxivauthors}{Zhongling Xu$^{*}$ \qquad Hairong Wang\\[0.45em]
  University of Texas at Austin, Austin, TX, USA}
\newcommand{\arxivcorresponding}{xuzhl@utexas.edu}
\newcommand{\arxivpdfauthor}{Zhongling Xu, Hairong Wang}
\newcommand{\codeurl}{https://github.com/Carlxuzhl/practice-in-text}
\newcommand{\modelurl}{https://huggingface.co/xuzhl/practice-in-text-aligned-reader}
 
\definecolor{hfyellow}{HTML}{FFD21E}
\definecolor{hforange}{HTML}{FF9D00}
\newcommand{\hficon}{\raisebox{-0.17em}{\tikz[x=1em, y=1em, line cap=round]{\fill[hfyellow] (0,0) circle (0.46);\fill[black!85] (-0.15,0.08) circle (0.055) (0.15,0.08) circle (0.055);\draw[black!85, line width=0.35pt] (-0.17,-0.09) .. controls (-0.08,-0.22) and (0.08,-0.22) .. (0.17,-0.09);\fill[hforange] (-0.35,-0.3) ellipse [x radius=0.15, y radius=0.11] (0.35,-0.3) ellipse [x radius=0.15, y radius=0.11];}}}

\hypersetup{pdftitle={Practice in Text, Act in Pixels: Correcting Where Perception Fails}, pdfauthor={\arxivpdfauthor}}

\title{Practice in Text, Act in Pixels: Correcting Where Perception Fails}

\makeatletter
\patchcmd{\@maketitle}{\vskip .375in}{\vskip .1in}{}{\PackageWarning{arxiv}{title spacing patch failed}}
\makeatother

\author{\arxivauthors\\[0.7em]{\small\faGithub~\textbf{Code}: \url{\codeurl}}\\[0.3em]{\small\hficon~\textbf{Aligned reader}: \url{\modelurl}}}

\begin{document}
\twocolumn[{\renewcommand\twocolumn[1][]{#1}\maketitle
  \newcommand{\teaserhead}[1]{{\sffamily\bfseries\fontsize{8}{9.5}\selectfont #1}\par\vspace{3pt}}
\begin{center}
  \captionsetup{type=figure}
\begin{minipage}[t]{0.665\textwidth}\vspace{0pt}
    \vspace*{-3.3pt}\includegraphics[width=\linewidth]{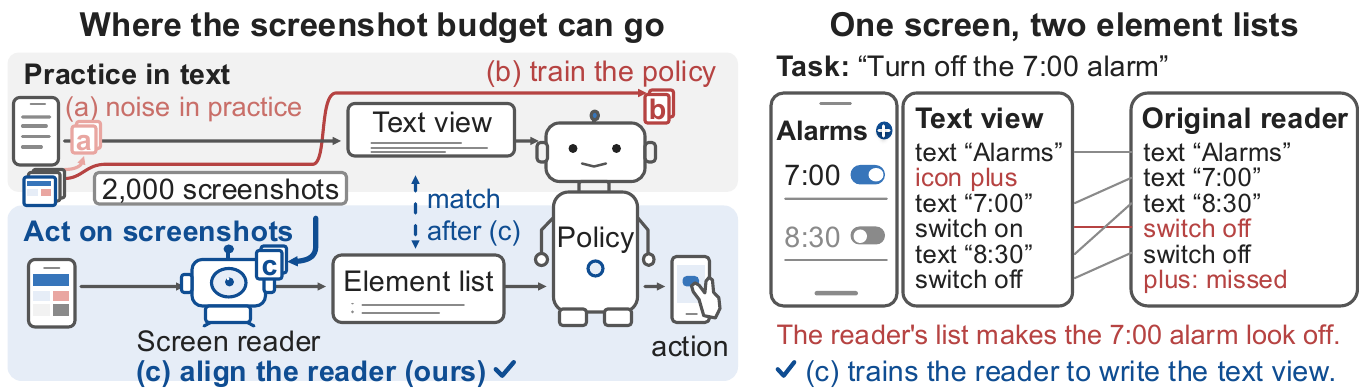}
  \end{minipage}\hfill
\begin{minipage}[t]{0.305\textwidth}\vspace{0pt}\centering
    \teaserhead{Same budget, different uses}
    \begin{tikzpicture}[x=0.042cm, y=1cm, plotfont]
  \def\hb{0.066}\colorlet{noshotink}{black!58}\definecolor{noiseink}{HTML}{C96F69}
  \newcommand{\plotbar}[6]{\fill[#4] (0,#1-\hb) rectangle (#2,#1+\hb);
    \draw[black!80, line width=0.45pt] (#2-#3,#1) -- (#2+#3,#1)
      (#2-#3,#1-0.045) -- (#2-#3,#1+0.045) (#2+#3,#1-0.045) -- (#2+#3,#1+0.045);
    \node[anchor=west, inner sep=0.8pt, fill=white, #6] at (#2+#3+1.2,#1) {#2};
    \node[anchor=south west, inner sep=0pt] at (0,#1+\hb+0.012) {#5};}
\newcommand{\plottoken}[3]{\pic[scale=0.8] at ($(0,#1+\hb+0.117)+(0.11cm,0)$) {budget={#2}{#3}};}\def\tokpad{\hspace*{0.27cm}}
\draw[refgray, dashed, line width=0.5pt] (\succTextClean,0.62) -- (\succTextClean,2.48); \node[anchor=west, inner sep=1.5pt, text=refgray, align=left] at (\succTextClean,2.30) {in text\\\succTextClean};
  \plotbar{2.22}{\succRealClean}{\succRealCleanStd}{black!30}{\textcolor{noshotink}{No screenshots}}{}
  \plotbar{1.78}{\succRealRate}{\succRealRateStd}{lightred}{\tokpad\textcolor{noiseink}{Noise in practice}}{}
  \plottoken{1.78}{lightred}{a}
  \plotbar{1.34}{\succRealDirect}{\succRealDirectStd}{basered}{\tokpad\textcolor{basered}{Train the policy}}{}
  \plottoken{1.34}{basered}{b}
  \plotbar{0.90}{\succSameLabel}{\succSameLabelStd}{darkred}{\textcolor{darkred}{Adapted policy, (c)'s labels}}{}
  \plotbar{0.46}{\succAlignClean}{\succAlignCleanStd}{ourblue}{\tokpad\textbf{\textcolor{ourblue}{Align the reader (ours)}}}{font=\sffamily\bfseries\fontsize{7}{8.2}\selectfont}
  \plottoken{0.46}{ourblue}{c}
\draw[ourblue, line width=0.6pt] (101,1.34) -- (103.5,1.34) -- (103.5,0.46) -- (101,0.46);
  \node[anchor=west, inner sep=1pt, align=left, text=ourblue, font=\sffamily\bfseries\fontsize{7}{8.2}\selectfont] at (103.8,0.90) {\dMain\\[-1.5pt]{\fontsize{6}{7}\selectfont\mdseries points}};
\draw[black!75, line width=0.45pt] (0,0.20) -- (100,0.20);
  \foreach \t in {0,25,50,75,100} {\draw[black!75, line width=0.45pt] (\t,0.20) -- (\t,0.15);
    \node[anchor=north, inner sep=1pt, plottick] at (\t,0.15) {\t};}
  \node[anchor=north] at (50,-0.10) {Full-task success on screenshots (\%)};
\end{tikzpicture}   \end{minipage}
\caption{\textbf{Spending the same budget of about \budgetShots{} screenshots on aligning the reader gives clearly higher success than spending it on the policy.} Left: the budget can go to practice (a; \cref{sec:simulating}), the policy (b; \cref{sec:main}) or the reader (c; \cref{sec:method}). Middle: an illustration of a missed element, a misread state and the reader's own order (\cref{fig:errors} shows real misreads). Right: full-task success on new instances (mean\,$\pm$\,std, two training seeds); dark red: the policy adapted with (c)'s labels instead; dashed: the no-screenshot policy in text.}
  \label{fig:teaser}
\end{center}
 }]
\begin{abstract}
GUI agents can learn cheaply from a text view of an interface and act on screenshots through a vision-language screen reader. Yet on a simulated mobile platform with a low-cost reader, policies that complete most tasks in text fail on many of the same tasks on screenshots. We ask where a small budget of real screenshots is best spent to close this gap. We find that the reader's misreads are systematic and recur on the same interface elements. Noise matched to the reader's error rates, a natural sim-to-real remedy, misses this structure, and in a pre-specified comparison it did not beat training the policy directly on screenshots. We therefore correct the misreads at their source with \textbf{reader alignment}, which fine-tunes the reader on screenshots paired with their text views. The reader learns to output each screen's text view, as in text practice, and the policy stays unchanged. Spent on the reader, the screenshot budget gives \succAlignClean\% full-task success on new task instances, against \succRealDirect\% when spent on training the policy. Given the same labeled screenshots, adapting the reader also beats adapting the policy, and a single aligned reader improves every text-trained policy we tried without retraining any of them. \end{abstract}
 {\let\thefootnote\relax\footnotetext{$^{*}$Corresponding: \texttt{\arxivcorresponding}}}\section{Introduction}
\label{sec:intro}

Agents that operate graphical interfaces can learn from screenshots or from a text view of the interface, a list of its visible elements.
Verifiable simulated GUI platforms that run many episodes in parallel~\cite{wu2026mobilegym} and LLM-based simulation of UI states~\cite{wang2025uisimulator} make large-scale practice cheap, and far cheaper in text than in pixels. On MobileGym, the simulated mobile platform we use, the text view is generated from each rendered screen, and an episode costs about \costRatioEpisode$\times$ more GPU time on screenshots than in text (supplementary \cref{supp:accounting}). At test time, however, an agent often has only the screen, because structured text such as HTML can be lengthy or unavailable~\cite{cheng2024seeclick}.
A policy that learned in text then needs a vision-language \emph{screen reader} that turns each screenshot into the same kind of element list.
We call this recipe \textbf{practice in text, act in pixels}: the policy does all its learning in the text view, from imitating scripted demonstrations to retraining on its own successful episodes, and only the reader sees pixels (\cref{fig:teaser}, left). 

Yet with a low-cost reader, Qwen3-VL-4B on half-size screenshots, a policy trained in text completes \succTextClean\% of new instances of its practiced MobileGym tasks in text but only \succRealClean\% on screenshots of the same instances. This \emph{text-to-pixels gap} is at least \nosign{\gapBoth} points for each of five training methods, including the three that already spend screenshots on practice or on the policy. The question is where a fixed budget of real screenshots is best spent to close this gap: on the practice environment, on the policy or on the reader (uses (a)--(c) in \cref{fig:teaser}).

\begin{figure}[t]
  \centering
  \includegraphics[width=\linewidth]{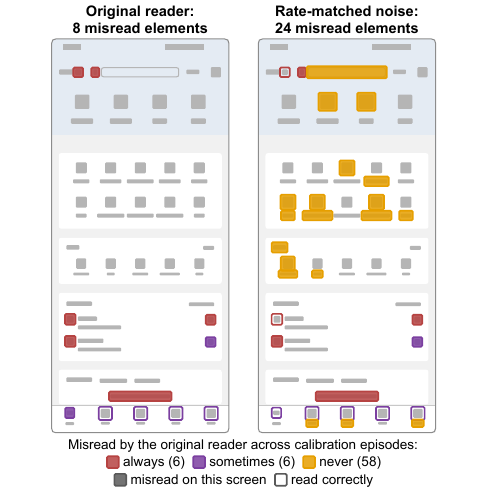}
\caption{\textbf{The reader misreads largely the same elements across episodes; rate-matched noise mostly corrupts other elements.} A schematic of one calibration screen (\cref{sec:setup}), drawn from its text view (bars: text; squares: icons; screenshot in supplementary \cref{fig:supp_textview}) and chosen by a rule fixed in advance (supplementary \cref{supp:reader}). Color: how often the original reader misreads each element in calibration episodes; never-misread elements are gray unless misread here. Colored fill: misread on this read or in this noise draw. The noise draw puts \figTwoNoiseOnNever{} of its \figTwoNoiseErr{} misreads on never-misread elements; the noise matches the reader's error rates over the calibration screens, not per screen.} \label{fig:errors}
\end{figure}

The reader's output differs from the text view in two ways (\cref{fig:teaser}, middle): the reader misreads some of the content, leaving elements out or getting their text or state wrong, and it lists the elements in its own order instead of the top-to-bottom \emph{layout order} of the text view.
A controlled decomposition indicates that each difference lowers success and that together they cost less than the sum of their separate costs (\cref{sec:decomp}). The key observation is that the misreads are systematic (\cref{fig:errors}).
An element that the reader misread before is misread again in other episodes far more often than independent errors at the same rate would predict (\persistSmall\% vs.\ \persistSmallIndep\%; \cref{sec:systematic}); the same holds with sampled decoding and for a larger reader of the same family. 

A natural way to spend the budget, in the spirit of domain randomization~\cite{tobin2017domain}, is to measure the reader's errors and inject such errors into the text view, so that practice looks like the reader.
Noise matched to the reader's measured error rates (\emph{rate-matched noise}) draws its errors independently in each episode, so the errors do not stay on the elements that the reader misreads (\cref{fig:errors}). Practice environments built this way approximately match the reader's overall effect on success, but they predict which tasks fail on screenshots less accurately than a second run on screenshots does (correlation across tasks of at most \fidReaderMax, against \fidRealTwo). In a pre-specified comparison with equal screenshot budgets, practice with rate-matched noise showed no measurable difference from training the policy directly on screenshots. 

Instead of making practice look like the reader, we make the reader look like practice.
\textbf{Reader alignment} fine-tunes the screen reader on about \budgetShots{} screenshots paired with their text views. The reader learns to output each screen's text view, with the content and order the policy practiced on, and the policy stays unchanged. Each screenshot thus teaches the reader a full element list, whereas direct training on screenshots learns only from episodes that succeed. Because the misreads recur on largely the same elements, we expect a small budget of labeled screenshots to correct them at their source. Reader alignment is supervised fine-tuning, the GUI counterpart of mapping real inputs to a canonical simulated view in robotic grasping~\cite{james2019sim}; we add evidence on where the screenshots should go. 

With the same budget of about \budgetShots{} screenshots, aligning the reader gives \succAlignClean\% full-task success on new instances of the practiced tasks, against \succRealDirect\% for training the policy directly on screenshots (\cref{fig:teaser}, right), and the comparison replicates on two further instance sets. Given the same labeled screenshots, adapting the reader also beats adapting the policy. The aligned 4B reader clearly exceeds an unaligned 8B reader, and it raises all three text-trained policies we tried by at least \nosign{\dAlignRate} points without retraining any of them.

Because MobileGym renders repeated screens identically, a lookup baseline that returns the stored text view for exact repeats of alignment screenshots, and the sorted reader output otherwise, recovers \lookupGainPct\% of alignment's gain over the original reader. Alignment remains clearly higher, and the aligned reader also misreads fewer elements on screens absent from those screenshots. On apps held out of alignment, misreads still fall, but task success shows no measurable difference. 

Our contributions are as follows: \begin{enumerate}[nosep,leftmargin=*]
  \item Reader alignment, a simple way to spend a small screenshot budget that leaves the policy unchanged; alignment gives \dMainApprox{} points higher full-task success than spending the same budget on training the policy. \item A controlled measurement of the text-to-pixels gap and its decomposition into misread content and the reader's order.
  \item Evidence that a vision-language screen reader's errors are systematic and tied to specific elements, which noise matched to its error rates does not reproduce. \item A task-level test of whether a text practice environment predicts which tasks fail on screenshots, adapted from Sim2Real predictivity~\cite{kadian2020sim2real}.
\end{enumerate}

Every headline comparison was decided by a rule fixed before its result was seen; we proposed reader alignment after the first comparison of five training methods (\cref{sec:method}).
 \section{Related work}
\label{sec:related}

\paragraph{Learning GUI agents in text and in pixels.}
Web and mobile agents are often trained or evaluated on text renderings of the interface, such as the DOM or the accessibility tree~\cite{shi2017wob,liu2018workflow,zhou2024webarena,deng2023mind2web,rawles2025androidworld}, while others act directly on screenshots~\cite{hong2024cogagent,cheng2024seeclick,qin2025uitars,lin2025showui,xu2025aguvis}.
ALFWorld pairs a text world with an embodied one~\cite{shridhar2021alfworld}, and in EMMA an expert in a parallel text world teaches a visual agent~\cite{yang2024emma}.
Our text view shows only what is visible on the screen, so that a screen reader can produce it; we study where the step from this text to pixels breaks down and repair it at the reader, keeping the policy in text.

\paragraph{Screen parsing and grounding.}
Screen-parsing models read interface elements from screenshots~\cite{lu2024omniparser,lee2023pix2struct,you2024ferretui}.
Grounding is a major source of failure for agents built on large multimodal models~\cite{zheng2024seeact,xie2024osworld}, and grounding models are now trained on millions of synthesized GUI elements~\cite{gou2025uground,wu2025osatlas}. These models aim at one reader for all interfaces. We start from an existing reader and ask whether a fixed budget of about \budgetShots{} screenshots from the target platform is better spent on the reader or on the policy. 

\paragraph{Sim-to-real transfer.}
Domain randomization makes practice harder and broader so that the real world looks like one more variation~\cite{tobin2017domain,peng2018sim}, and the randomization can be adapted with real-world experience~\cite{chebotar2019closing}.
Domain-randomized training has also been studied for tool-use agents~\cite{zhou2026simulation}, and perception error models inject detector-like errors into simulation for autonomous driving~\cite{piazzoni2020modeling}; our rate-matched noise follows this line for a screen reader.
Closest to our approach are methods that bridge the gap at the input, by adapting simulated images toward real ones~\cite{bousmalis2018using} or by mapping real inputs to a canonical simulated view~\cite{james2019sim,zhang2019vrgoggles}, which we carry over to GUI screen readers.

\paragraph{Predicting real performance.}
Sim2Real predictivity~\cite{kadian2020sim2real} validates a simulator by how well simulated performance predicts real performance.
We apply the same idea to text practice environments, measure predictivity per task template, and take the agreement between two runs on screenshots as the ceiling.
 \section{Setup}
\label{sec:setup}

\paragraph{Platform.}
We use MobileGym~\cite{wu2026mobilegym}, which runs simulated Chinese-language mobile apps, such as WeChat, Alipay and Clock, in a browser (code under Apache-2.0, data under CC BY-NC 4.0, used for academic research only). Each task, such as deleting an alarm or checking a balance, comes from a \emph{template}, and an \emph{instance seed} fills in its details, so one template yields many instances. An \emph{episode} is one attempt at an instance; when it ends, a program called the \emph{judge} checks its outcome, and the episode is a \emph{full-task success} only if the judge accepts it.

\paragraph{Text view and practice.}
When practicing in text, the policy sees each screen as a \emph{text view}: one line per visible element, with its role (text, icon, input field, switch, check mark or image), its text or icon name, its state, its color and its box on the screen (example in supplementary \cref{fig:supp_textview}). The lines run from top to bottom and, within a row, from left to right; we call this the \emph{layout order}.
The text view is generated from the rendered page and holds only what the screen shows, with no HTML attributes, routes, hidden state or list of legal actions (leakage checks in supplementary \cref{supp:textview}). A \emph{practice environment} runs the apps and shows the policy their text view, either clean or corrupted to imitate reading errors.

\paragraph{Acting on screenshots.}
At test time the agent has only screenshots.
A vision-language \emph{screen reader}, Qwen3-VL-4B-Instruct~\cite{qwen3vl2025}, reads half-size screenshots with greedy decoding and turns each into a list in the format of the text view, but in its own order, \emph{the reader's order}. An element is \emph{misread} when the reader misses it or gives it the wrong text or state, and the share of elements that are misread is the \emph{element error rate}.
The policy chooses an action from the reader's output, an executor taps the box that the reader gave for the chosen element, and the judge checks the outcome; we call this chain the \emph{real pipeline}. We fixed this low-cost reader configuration with a rule, before comparing training methods. The rule's main requirements were that a read cost at most half as much as with Qwen3-VL-8B on full-size screenshots and that the reader lower a scripted policy's success by at least \readerPickLossLine{} points, so that a gap remained to study.
The thresholds were set in advance. After seeing results, we changed how reads were scored (twice) and timed (one read at a time, then \readerPickInflight{} in parallel), and the chosen configuration depends on the second scoring change and on the timing change (full rule and history in supplementary \cref{supp:reader}). 

\paragraph{Policies.}
The policy is the language model Qwen3-8B~\cite{qwen3_2025} with LoRA~\cite{hu2022lora}.
At each step it reads the task, its recent actions with the notes it wrote, and the current element list, and writes the next action (settings in supplementary \cref{supp:practice}). It first learns by imitating scripted demonstrations recorded in the text view; we call the result the \emph{start policy}.

\paragraph{Calibration set and training groups.}
The \emph{calibration set} provides the screenshots that the methods spend: we run the start policy in the real pipeline on training instances and store each screenshot with its true text view (\calibEpisodes{} episodes, \calibShots{} screenshots).
The first round of our study compares five groups: \begin{itemize}[nosep,leftmargin=*]
  \item \emph{clean-text practice}, on the true text view;
  \item \emph{generic-noise practice}, on a text view corrupted by noise with preset parameters (supplementary \cref{supp:practice});
  \item \emph{rate-matched-noise practice}, on a text view with misreads injected at the reader's element error rates on the calibration screenshots;
  \item \emph{direct screenshot training} (\emph{direct training} for short), which practices in the real pipeline, on the reader's output;
  \item \emph{combined training}, which uses the first \directRoundOneShots{} calibration screenshots to fit rate-matched noise and also to train the policy directly.
\end{itemize}
Each group then runs two practice rounds: the policy attempts tasks in its own practice environment, keeps the episodes that the judge accepts, and a new LoRA is trained on all episodes kept so far and the demonstrations. All groups share the initialization, the learning algorithm and its hyperparameters.

\paragraph{Screenshot budget.}
All methods that use screenshots spend the same budget of about \budgetShots{} screenshots; reader alignment and rate-matched noise use the same \calibShots{} calibration screenshots, while direct training uses the first \directRoundOneShots{} of them plus about \directRoundTwoShots{} it collects itself. In their first practice round, direct screenshot training and combined training learn from the judge-accepted episodes in their part of the calibration set, on the reader's output; in their second, they collect the rest of their budget (supplementary \cref{supp:practice}). 

\paragraph{Evaluation.}
We evaluate on the \emph{confirmation set}: \confInstances{} new instances (instance seeds \confSeedFirst--\confSeedLast) of the \confTemplates{} templates of the calibration set, which cover \confApps{} apps. Each group is evaluated with its two training seeds (\confEpisodes{} episodes), and the start policy, trained once, with \confInstances{} episodes.
The \emph{first-round evaluation set} is an earlier, disjoint set of instances of the same templates (supplementary \cref{supp:first_round}).
We use none of MobileGym's \officialTestTemplates{} official test templates, because our policies learn only the practiced templates and mostly fail on new ones even in text (supplementary \cref{supp:limitations}). The main metric is full-task success in the real pipeline; success in text is only a diagnostic.

\paragraph{Statistics.}
Differences are in points of full-task success, computed from unrounded rates.
We assess each difference with a bootstrap over task templates and describe it with one of four fixed phrases: if the bootstrap confirms the difference, \emph{clearly higher/lower} (at least \wordThreshold{} points) or \emph{slightly higher/lower} (less); if not, \emph{points toward} (at least \wordThreshold{} points) or \emph{no measurable difference} (less). Decision rules for the headline comparisons were fixed before their results were seen; the comparisons added after the main comparison (\cref{sec:results}: same labels, lookup, replication, held-out apps, full-size 8B reader) each had their rule fixed before their result, and the rest are descriptive. Supplementary \cref{supp:stats} gives the procedure, the rules and all intervals.
 \section{Diagnosing the text-to-pixels gap}
\label{sec:findings}

\subsection{Success drops from text to screenshots}
\label{sec:overstate}
Every first-round group loses at least \nosign{\gapBoth} points of success from text to screenshots (\cref{tab:text_vs_real}), so every group meets the criterion fixed in advance: a gap of at least \gapCriterion{} points that the bootstrap confirms (supplementary \cref{tab:supp_intervals}). The same policies complete at least \succTextRate\% of these instances from the true text view, so we look for the loss in what the reader gives them.

\begin{table}[t]
  \centering
  \footnotesize
  \setlength{\tabcolsep}{4.5pt}
  \caption{\textbf{Every first-round group succeeds far less often on screenshots than in text.} Full-task success (\%) on the confirmation set from the true text view and on screenshots with the original reader (mean\,$\pm$\,std over two training seeds). Subscripts: pixels minus text, in points; in all tables, a subscript is green or red when the bootstrap confirms a rise or a drop and gray when it does not. Budget: screenshots used, rounded to the nearest hundred.} \label{tab:text_vs_real}
\begin{tabularx}{\linewidth}{@{}Xrll@{}}
    \toprule
    Training group              & Budget                & Text                                        & Pixels \\
    \midrule
    Clean-text practice         & 0                     & \msd{\succTextClean}{\succTextCleanStd}     & \msd{\succRealClean}{\succRealCleanStd}\ddown{\gapClean} \\
    Generic-noise practice      & 0                     & \msd{\succTextGeneric}{\succTextGenericStd} & \msd{\succRealGeneric}{\succRealGenericStd}\ddown{\gapGeneric} \\
    Rate-matched-noise practice & \budgetShots          & \msd{\succTextRate}{\succTextRateStd}       & \msd{\succRealRate}{\succRealRateStd}\ddown{\gapRate} \\
    Direct screenshot training  & \budgetShots          & \msd{\succTextDirect}{\succTextDirectStd}   & \msd{\succRealDirect}{\succRealDirectStd}\ddown{\gapDirect} \\
    Combined training           & \budgetShots          & \msd{\succTextBoth}{\succTextBothStd}       & \msd{\succRealBoth}{\succRealBothStd}\ddown{\gapBoth} \\
    \bottomrule
  \end{tabularx}
\end{table}

\subsection{The gap has two overlapping sources}
\label{sec:decomp}

\begin{figure}[t]
  \centering
  \colorlet{cellgray}{black!68}
\begin{tikzpicture}
\begin{axis}[paperplot, width=5.9cm, height=2.6cm,xmin=-0.6, xmax=2.4, ymin=0, ymax=100,
  xtick={0,1,2}, xticklabels={Layout order, Reader's order, Random shuffle},
  ytick={0,25,50,75,100}, ylabel={Full-task success (\%)},
  legend style={draw=none, fill=none, font=\sffamily\fontsize{7}{8.2}\selectfont, legend columns=-1,
    at={(0.5,1.07)}, anchor=south, /tikz/every even column/.append style={column sep=9pt}},
  legend cell align=left]
  \addplot[refgray, dashed, line width=0.5pt, forget plot] coordinates {(-0.6,\decompSuccReal) (2.4,\decompSuccReal)};
  \node[anchor=north west, inner sep=1.5pt, text=refgray] at (axis cs:-0.6,\decompSuccReal) {real pipeline \decompSuccReal};
  \addplot[cellgray, line width=0.9pt, mark=*, mark size=1.7pt] coordinates {(0,\decompSuccClean) (1,\decompSuccReaderOrder)};
  \addlegendentry{correct elements}
  \addplot[basered, line width=0.9pt, mark=square*, mark size=1.5pt] coordinates {(0,\fidSuccReplayTwoLayout) (1,\fidSuccReplayTwo)};
  \addlegendentry{misreads injected}
  \addplot[cellgray, densely dotted, line width=0.8pt, forget plot] coordinates {(1,\decompSuccReaderOrder) (2,\decompSuccShuffle)};
  \addplot[cellgray, only marks, mark=*, mark size=1.7pt, mark options={fill=white, line width=0.7pt}, forget plot] coordinates {(2,\decompSuccShuffle)};
\node[anchor=south, inner sep=2.5pt] at (axis cs:0,\decompSuccClean) {\decompSuccClean};
  \node[anchor=south west, inner sep=2pt] at (axis cs:1,\decompSuccReaderOrder) {\decompSuccReaderOrder};
  \node[anchor=west, inner sep=3pt] at (axis cs:2,\decompSuccShuffle) {\decompSuccShuffle};
  \node[anchor=south east, inner sep=2pt] at (axis cs:0,\fidSuccReplayTwoLayout) {\fidSuccReplayTwoLayout}; \node[anchor=south west, inner sep=2pt] at (axis cs:1,\fidSuccReplayTwo) {\fidSuccReplayTwo}; \node[anchor=south west, inner sep=1pt, text=cellgray] at (axis cs:0.5,{(\decompSuccClean+\decompSuccReaderOrder)/2}) {\fsign{\effOrderClean}};
  \node[anchor=south, inner sep=2.5pt, text=basered] at (axis cs:0.5,{(\fidSuccReplayTwoLayout+\fidSuccReplayTwo)/2}) {\fsign{\effOrderGivenMisread}};
\end{axis}
\end{tikzpicture} \caption{\textbf{Misreads and the reader's order both lower success, and their losses overlap.} Full-task success of the start policy on \decompInstances{} unseen instances, \decompSamples{} samples each, with correct elements (gray) or with misreads injected by the element-replay simulator with resampled reads (red; \cref{sec:simulating}), in the layout order, the reader's order or a random shuffle (open marker). Dashed: real pipeline. Labels: cost of the reader's order in points; with misreads, the difference only points toward a cost (supplementary \cref{tab:supp_decomp}).}
  \label{fig:decomp}
\end{figure}
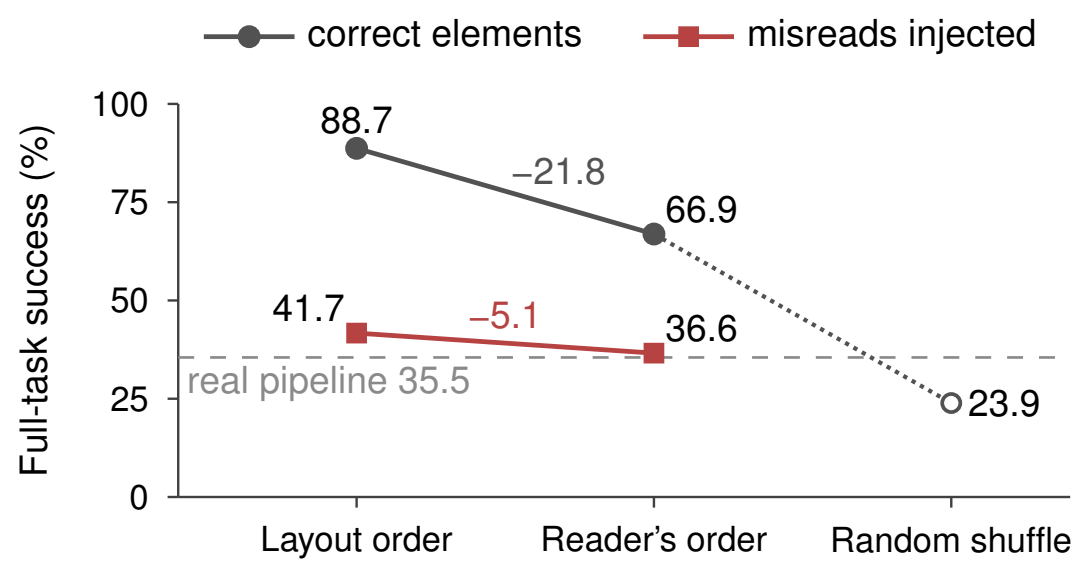

The reader's output differs from the text view in content and in order, and each difference lowers success (\cref{fig:decomp}).
To separate the two, we evaluate the start policy, the common starting point of all groups, on instances it never saw (\cref{fig:decomp}). Misreads are injected by an element-replay simulator, which replays the reader's recorded reads of each element and, for a screen in the calibration set, draws one of its recorded reads anew in every episode (\cref{sec:simulating}; supplementary \cref{supp:fidelity}). The reader's order is taken from its output on similar screens.
Listing the correct elements in the reader's order lowers success by \magn{\effOrderClean} points.
Misreads cost \magn{\effMisreadLayout} points in the layout order and \magn{\effMisreadReaderOrder} in the reader's order, so the two losses overlap: together they cost less than the sum of their separate costs. These effects are approximate (supplementary \cref{tab:supp_decomp}), because the simulator matches the real pipeline in aggregate success but not per template (\cref{sec:simulating}). The reader's order still beats a random shuffle (\cref{fig:decomp}), yet both give lower success than the layout order. A remedy must restore both the content and the order of the text view. 

\subsection{Misreads are systematic}
\label{sec:systematic}

The reader's misreads recur on the same elements across episodes (\cref{fig:errors}), and most of them are missed elements (\cref{tab:reader}, bottom). For every element that the reader misreads in one episode, we ask whether it is misread again in another episode (matching rule in supplementary \cref{supp:reader}), and compare with a control in which errors occur independently at the same rate.
For Qwen3-VL readers of two sizes, and with sampling at temperature \sampledTemp, misreads recur on the same elements far more often than independent errors would (\cref{tab:persist}). For a reader from a second model family, InternVL3-8B~\cite{zhu2025internvl3}, misreads also recur clearly more often than independent errors, but with a much smaller margin. Because this reader misses most elements under our prompt, even independent errors at that rate would recur often, and the margin misses our pre-specified threshold of \persistThreshold{} points.

\begin{table}[t]
  \centering
  \footnotesize
  \setlength{\tabcolsep}{4.5pt}
  \caption{\textbf{Misreads recur on the same elements more often than independent errors would, far more so for the Qwen3-VL readers.} Share (\%) of misread elements that are misread again in another episode, for the reader and for independent errors at the same rate. Subscripts: observed minus independent, in points. Last column: the difference reaches the pre-specified threshold of \persistThreshold{} points. $T$: sampling temperature; the Qwen3-VL-8B row uses full-size screenshots (supplementary \cref{supp:reader}).} \label{tab:persist}
\begin{tabularx}{\linewidth}{@{}Xcllc@{}}
    \toprule
                 &              & \multicolumn{2}{c}{Misread again (\%)}                  & \\
    \cmidrule(lr){3-4}
    Reader       & $T$          & Observed                                 & Independent  & $\geq$\,\persistThreshold \\
    \midrule
    Qwen3-VL-4B  & \readerTemp  & \persistSmall\dup{\persistSmallDiff}     & \persistSmallIndep   & \cmark \\
    Qwen3-VL-8B  & \readerTemp  & \persistLarge\dup{\persistLargeDiff}     & \persistLargeIndep   & \cmark \\
    Qwen3-VL-4B  & \sampledTemp & \persistSampled\dup{\persistSampledDiff} & \persistSampledIndep & \cmark \\
    InternVL3-8B & \readerTemp  & \persistOther\dup{\persistOtherDiff}     & \persistOtherIndep   & \xmark \\
    \bottomrule
  \end{tabularx}
\end{table}

On new instances, \newErrMisreadPct\% of the 4B reader's errors fall on elements it already misread in the calibration screenshots, although these elements account for only \misreadReadsPct\% of its reads.
The errors also repeat in content. When the 4B reader gets the same piece of text wrong in two episodes, it writes the same wrong text in \sameWrongPct\% of cases, against \sameWrongPctRate\% under the control. Adding an element's own calibration history to a predictor based on the error rates of similar elements raises the AUC for its next error from \aucBase{} to \aucHistory{} (supplementary \cref{supp:reader}). Error rates alone do not say where the errors fall; \cref{sec:simulating} tests environments that inject errors by rate and by element.

\subsection{Simulating the reader falls short}
\label{sec:simulating}

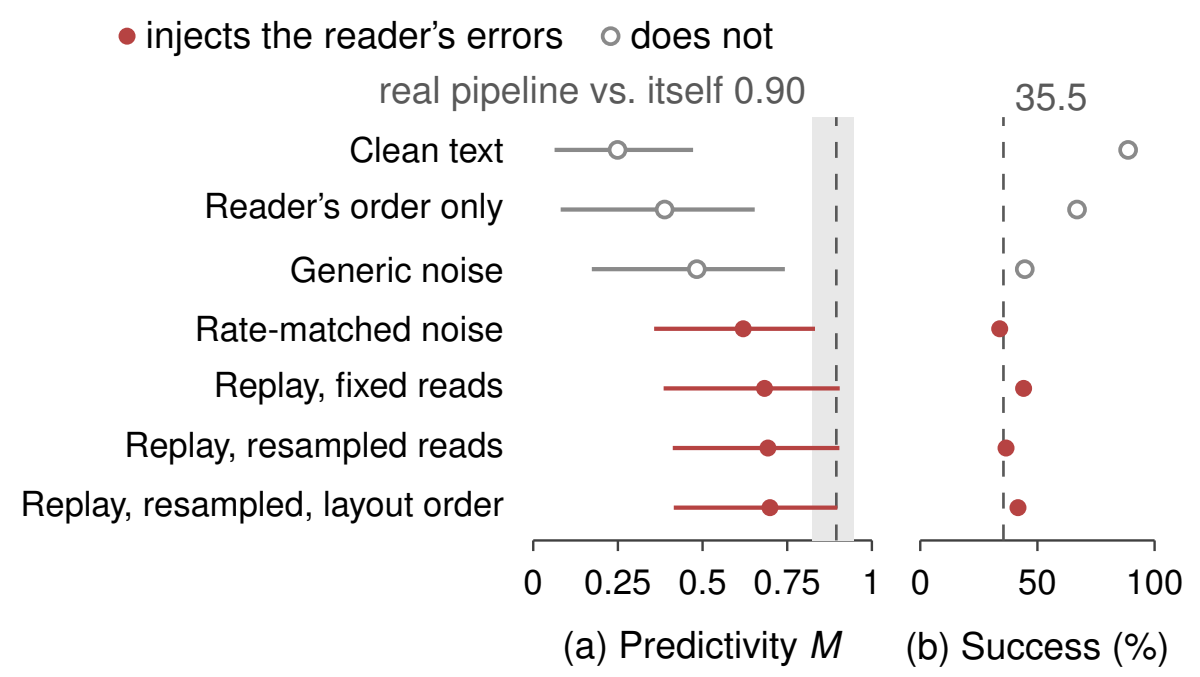
\begin{figure}[t]
  \centering
  \newcommand{\plotenvci}[5]{\ifnum#5=1
    \draw[basered, line width=0.8pt] (axis cs:#3,#1) -- (axis cs:#4,#1);
    \fill[basered] (axis cs:#2,#1) circle[radius=1.6pt];
  \else
    \draw[refgray, line width=0.8pt] (axis cs:#3,#1) -- (axis cs:#4,#1);
    \filldraw[draw=refgray, fill=white, line width=0.7pt] (axis cs:#2,#1) circle[radius=1.5pt];
  \fi}
\newcommand{\plotenvpt}[3]{\ifnum#3=1 \fill[basered] (axis cs:#2,#1) circle[radius=1.6pt];
  \else \filldraw[draw=refgray, fill=white, line width=0.7pt] (axis cs:#2,#1) circle[radius=1.5pt]; \fi}
\begin{tikzpicture}
\begin{axis}[paperplot, name=pa, width=2.24cm, height=2.8cm,xmin=0, xmax=1, ymin=0.45, ymax=7.55,
  xtick={0,0.25,0.5,0.75,1}, xticklabels={0,0.25,0.5,0.75,1},
  ytick={1,...,7}, yticklabels={{Replay, resampled, layout order},{Replay, resampled reads},{Replay, fixed reads},{Rate-matched noise},{Generic noise},{Reader's order only},{Clean text}},
  y axis line style={draw=none}, ytick style={draw=none},
  xlabel={(a) Predictivity \textit{M}}]
  \fill[black!9] (axis cs:\fidRealLo,0.45) rectangle (axis cs:\fidRealHi,7.55);
  \draw[black!65, dashed, line width=0.5pt] (axis cs:\fidReal,0.45) -- (axis cs:\fidReal,7.55);
  \node[anchor=south east, inner sep=1pt, text=black!65] at (axis cs:\fidRealLo,7.55) {real pipeline vs.\ itself \fidRealTwo}; \plotenvci{7}{\fidClean}{\fidCleanLo}{\fidCleanHi}{0}
  \plotenvci{6}{\fidOrderOnly}{\fidOrderOnlyLo}{\fidOrderOnlyHi}{0}
  \plotenvci{5}{\fidRandom}{\fidRandomLo}{\fidRandomHi}{0}
  \plotenvci{4}{\fidRate}{\fidRateLo}{\fidRateHi}{1}
  \plotenvci{3}{\fidReplayOne}{\fidReplayOneLo}{\fidReplayOneHi}{1}
  \plotenvci{2}{\fidReplayTwo}{\fidReplayTwoLo}{\fidReplayTwoHi}{1}
  \plotenvci{1}{\fidReplayTwoLayout}{\fidReplayTwoLayoutLo}{\fidReplayTwoLayoutHi}{1}
\end{axis}
\begin{axis}[paperplot, at={($(pa.east)+(0.32cm,0)$)}, anchor=west, width=1.55cm, height=2.8cm,
  xmin=0, xmax=100, ymin=0.45, ymax=7.55, xtick={0,50,100}, ytick=\empty,
  y axis line style={draw=none}, xlabel={(b) Success (\%)}]
  \draw[black!65, dashed, line width=0.5pt] (axis cs:\decompSuccReal,0.45) -- (axis cs:\decompSuccReal,7.55);
  \node[anchor=south west, inner sep=1pt, xshift=1pt, text=black!65] at (axis cs:\decompSuccReal,7.55) {\decompSuccReal}; \plotenvpt{7}{\fidSuccClean}{0}
  \plotenvpt{6}{\fidSuccOrderOnly}{0}
  \plotenvpt{5}{\fidSuccRandom}{0}
  \plotenvpt{4}{\fidSuccRate}{1}
  \plotenvpt{3}{\fidSuccReplayOne}{1}
  \plotenvpt{2}{\fidSuccReplayTwo}{1}
  \plotenvpt{1}{\fidSuccReplayTwoLayout}{1}
\end{axis}
\node[anchor=south west, inner sep=0pt, plotfont] at ($(pa.north west)+(-2.75cm,0.40cm)$)
  {\tikz[baseline=-0.6ex]{\fill[basered] (0,0) circle[radius=1.6pt];}~injects the reader's errors\quad
   \tikz[baseline=-0.6ex]{\filldraw[draw=refgray, fill=white, line width=0.7pt] (0,0) circle[radius=1.5pt];}~does not};
\end{tikzpicture} \caption{\textbf{Environments that inject the reader's errors roughly match the real pipeline's success but predict per-template success less well than a repeated real run.} (a)~Predictivity $M$ with bootstrap error bars; dashed line and gray band: the real pipeline predicting itself, which every environment falls measurably short of (supplementary \cref{tab:supp_fidelity}). (b)~Success of the start policy inside each environment, on the instances of \cref{fig:decomp}; dashed: real pipeline. Filled: injects the reader's errors. Replay: element-replay simulator.}
  \label{fig:fidelity}
\end{figure}

Environments that inject the reader's errors predict per-template success less well than a repeated real run, and practicing in one did not beat direct screenshot training.

\paragraph{Predictivity.}
If a practice environment imitates the real pipeline well, success in it should predict success on screenshots. We measure this per task template with the \emph{predictivity} $M$ of an environment: the Pearson correlation across templates between its success on one set of instances and the real pipeline's success on other instances of the same templates.
The real pipeline predicts itself with $M=\fidRealTwo$, which serves as the ceiling. Environments that inject the reader's errors (rate-matched noise and two versions of the element-replay simulator, with fixed or resampled reads) reach the highest $M$ (\cref{fig:fidelity}; supplementary \cref{tab:supp_fidelity}). Every environment stays below the ceiling, as the bootstrap confirms, although only barely for the element-replay environments. Both versions of the element-replay simulator failed our pre-specified predictivity test (supplementary \cref{supp:fidelity}), so no policy was trained with them.

\paragraph{Training outcomes.}
In the four pre-specified comparisons between first-round groups on the confirmation set (\cref{tab:text_vs_real}; supplementary \cref{tab:supp_intervals}), direct screenshot training and combined training are each clearly higher than clean-text practice, which uses no screenshots. Rate-matched-noise practice, which spends its budget on calibrating practice, shows no measurable difference from generic-noise practice or from direct screenshot training. All five pre-specified first-round comparisons (these four and the text-to-pixels gap) keep their sign on the first-round evaluation set (supplementary \cref{supp:first_round}). Training on screenshots also leaves the cost of the reader's order in place: after training, listing the true elements in the reader's order lowers every group's success by at least \nosign{\tdOrderMin} points, with no measurable difference in this cost between direct screenshot training and clean-text practice (supplementary \cref{supp:decomp}). Since the reader's errors are tied to elements (\cref{sec:systematic}), we spend the screenshots on the reader instead (\cref{sec:method}).

\begin{table}[t]
  \centering
  \footnotesize
  \setlength{\tabcolsep}{3pt}
\caption{\textbf{Aligning the reader gives clearly higher success than every alternative in this table.} Full-task success (\%), mean\,$\pm$\,std over two training seeds, on the confirmation set unless noted. Each row pairs a policy with a reader; blue: the clean-text policy with the aligned reader. Subscripts: difference from the blue row of the same block, in points. 8B: unaligned Qwen3-VL-8B. Budget: screenshots used, rounded to the nearest hundred. $^\dagger$Subscript not pre-specified.} \label{tab:main}
\begin{tabularx}{\linewidth}{Xlrl}\toprule
    Policy                & Reader                          & Budget                & Success (\%)\,$\uparrow$ \\
    \midrule
    \rowcolor{rowblue}Clean-text & \textbf{Aligned (ours)} & \budgetShots & \textbf{\succAlignClean}\pmstd{\succAlignCleanStd} \\
    Direct-training       & Original                        & \budgetShots          & \msd{\succRealDirect}{\succRealDirectStd}\ddown{\dMain} \\
    Adapted, same labels  & Original                        & \budgetShots          & \msd{\succSameLabel}{\succSameLabelStd}\ddown{\dAlignSameLabel} \\
    Clean-text            & Lookup                          & \budgetShots          & \msd{\succLookupClean}{\succLookupCleanStd}\ddown{\dAlignLookup} \\
    Clean-text            & 8B, half size                   & 0                     & \msd{\succLargeClean}{\succLargeCleanStd}\ddown{\dAlignLarge} \\
Clean-text            & 8B, full size                   & 0                     & \msd{\succLargeFullClean}{\succLargeFullCleanStd}\ddown{\dAlignLargeFull} \\
    Clean-text            & Aligned with \smallAlignShots$^\dagger$ & \smallAlignShots      & \msd{\succSmallAlignClean}{\succSmallAlignCleanStd}\ddown{\dFullVsSmall} \\
    \midrule
    \multicolumn{4}{l}{\textit{First-round evaluation set}} \\
    \rowcolor{rowblue}Clean-text & \textbf{Aligned (ours)} & \budgetShots & \msd{\origSuccAlignClean}{\origSuccAlignCleanStd} \\
    Direct-training       & Original                        & \budgetShots          & \msd{\origSuccRealDirect}{\origSuccRealDirectStd}\ddown{\origMain} \\
\midrule
    \multicolumn{4}{l}{\textit{Fresh instance set}} \\
    \rowcolor{rowblue}Clean-text & \textbf{Aligned (ours)} & \budgetShots & \msd{\freshSuccAlignClean}{\freshSuccAlignCleanStd} \\
    Direct-training       & Original                        & \budgetShots          & \msd{\freshSuccRealDirect}{\freshSuccRealDirectStd}\ddown{\freshMain} \\
    \bottomrule
  \end{tabularx}
\end{table}

\section{Reader alignment}
\label{sec:method}

Reader alignment trains the screen reader so that its parsed output for a screenshot is that screen's text view, with the same content and order as in practice. The policy is left unchanged. 

\paragraph{Data.}
We train the reader on the calibration screenshots. Each screenshot's \emph{training label} is its text view in the layout order, written in the reader's output format so that the same parser can read it (example in supplementary \cref{fig:supp_align_example}).
Alignment thus needs the text view of each calibration screenshot, which a simulated platform provides directly (supplementary \cref{supp:limitations} discusses real devices). 

\paragraph{Training.}
We train LoRA adapters of rank \alignLoraRank{} on all linear layers of the reader's language model, with the vision encoder frozen, at learning rate $\alignLR$ for \alignEpochs{} epochs, and keep the last checkpoint; training takes \alignGpuMinutes{} minutes on one A6000 GPU. The aligned reader keeps the original prompt, decoding settings, downscaling and parser, so only its weights change (supplementary \cref{supp:reader}). A reader call takes about as long after alignment as before (supplementary \cref{supp:accounting}).

\paragraph{Development history.}
Reader alignment was proposed after the first-round comparison and the gap decomposition.
Its decision rules were frozen before any result with the aligned reader was seen; at that point, only the direct-training group's results on the confirmation set had been seen.
There is a single version, and no hyperparameter was tuned.  \section{Results}
\label{sec:results}

\subsection{Reader alignment beats direct training}
\label{sec:main}
With the same budget of about \budgetShots{} screenshots, the clean-text policy with the aligned reader reaches \succAlignClean\% full-task success, clearly higher than the \succRealDirect\% of direct screenshot training with the original reader ($\dMain$ points; \cref{tab:main}). The result replicates on a second instance set ($\origMain$ points), the first-round evaluation set whose results motivated the method. The aligned reader had never been run on these instances, and the decision rule was fixed before the run (supplementary \cref{supp:first_round}). The result also replicates on a fresh instance set that no group had been run on ($\freshMain$ points; \cref{tab:main} and supplementary \cref{supp:first_round}). 

On the confirmation set, the clean-text policy with the aligned reader is higher than direct screenshot training on \tmplBetter{} templates, equal on \tmplSame{} and lower on \tmplWorse{} (supplementary \cref{supp:per_template}). Every screenshot used for alignment gives the reader a full element list to learn from, whereas direct screenshot training, in its first practice round, keeps only the \directRoundOnePassed{} judge-accepted episodes (\directRoundOneSamples{} policy samples) in its part of the calibration set. Given the same labeled screenshots, adapting the reader also beats adapting the policy (\cref{sec:scope}). A reader aligned with a second alignment seed gives \succAlignSeedB\%, with no measurable difference from the first, and is still clearly higher than direct screenshot training; it points toward higher success than the lookup baseline (\cref{sec:reader_side}) and the adapted policy (supplementary \cref{supp:reader}). 

Alignment alone closes \gapClosedPct\% of the clean-text policy's text-to-pixels gap. Once the reader's output matches the text view, the policy carries over most of what it learned in text.
A few templates account for most of the remaining text-to-pixels gap; \texttt{Calendar\allowbreak Earliest\allowbreak ToAlarm}, for example, succeeds in every text episode but in none on screenshots, with either reader and under direct screenshot training (supplementary \cref{supp:per_template}). In every failed episode of the four templates with most of the remaining gap, a pre-specified procedure finds that the first discrepancy between the aligned reader's output and the true text view is an ordering difference or a misread input field; none is traced to the executor or to the policy alone, although the first discrepancy need not be the only cause of a failure. 

\subsection{One reader, many policies}
\label{sec:reuse}
The aligned reader, trained once, raises the success of the three text-trained policies that we evaluated with it (start, clean-text and rate-matched-noise) by at least \nosign{\dAlignRate} points without retraining any of them, and also raises that of the direct-training policy (\cref{tab:reuse}). With it, the three text-trained policies end close together: the clean-text policy points toward higher success than the start policy, and the other two pairs show no measurable difference (supplementary \cref{tab:supp_intervals}). Practicing against rate-matched noise thus shows no measurable benefit once the reader is aligned, and what the policies learned in text carries over whether it came from imitation alone or from imitation and practice. 

The exception is a policy that never learned in text. The untrained base policy completes only \baseTextSucc\% of the instances even in text, and with the aligned reader its success shows no measurable difference from that with the original reader (\cref{tab:reuse}); it has no text-to-pixels gap for alignment to close. 

\subsection{Beyond reordering and exact repeats}
\label{sec:order_only}

\paragraph{Reordering alone.}
Reordering the original reader's output recovers little of alignment's \nosign{\dAlignClean}-point gain. Sorting the output by box position into the layout order needs no training and no screenshots.
For the clean-text policy, the sorted output points toward higher success than the unsorted one ($\dReorder$ points; \cref{tab:reuse}). This is close to the ordering cost of \magn{\effOrderGivenMisread} points that the decomposition finds once misreads are present (\cref{fig:decomp}), although policies and instances differ. 

\begin{table}[t]
  \centering
  \footnotesize
  \setlength{\tabcolsep}{4pt}
\caption{\textbf{One aligned reader raises all three text-trained policies we evaluated, without retraining.} Full-task success (\%) on the confirmation set, mean\,$\pm$\,std over two training seeds (one run for the start and base policies). Reordered: the original reader's output sorted into the layout order. Subscripts: difference from the original reader, in points. $^\dagger$Subscript not pre-specified.} \label{tab:reuse}
  \begin{tabularx}{\linewidth}{@{}Xlll@{}}
    \toprule
                                    & \multicolumn{3}{c}{Reader: success (\%)\,$\uparrow$} \\
    \cmidrule(l){2-4}
    Policy                          & Original                                    & Reordered                                                      & Aligned \\
    \midrule
    \multicolumn{4}{@{}l}{\textit{Learned only in text}} \\ Clean-text                      & \msd{\succRealClean}{\succRealCleanStd}     & \msd{\succReorderClean}{\succReorderCleanStd}\dgray{\dReorder} & \textbf{\succAlignClean}\pmstd{\succAlignCleanStd}\dup{\dAlignClean} \\
    Start                           & \succRealStart\nostd                        & --                                                             & \succAlignStart\nostd\dup{\dAlignStart} \\
    Rate-matched noise              & \msd{\succRealRate}{\succRealRateStd}       & --                                                             & \msd{\succAlignRate}{\succAlignRateStd}\dup{\dAlignRate} \\
    \midrule
    \multicolumn{4}{@{}l}{\textit{Other policies}} \\
    Direct-training$^\dagger$       & \msd{\succRealDirect}{\succRealDirectStd}   & --                                                             & \msd{\succDirectAligned}{\succDirectAlignedStd}\dup{\dAlignDirect} \\
Untrained base$^\dagger$        & \succRealBase\nostd                         & --                                                             & \succAlignBase\nostd\dgray{\dAlignBase} \\
    \bottomrule
  \end{tabularx}
\end{table}

\begin{table}[t]
  \centering
  \footnotesize
  \setlength{\tabcolsep}{5pt}
  \caption{\textbf{Alignment lowers the element error rate on every set and cuts every type of error.} Top: element error rate (\%) of the original and the aligned reader. Screens: number of screenshots (confirmation set: screens encountered with the original\,/\,aligned reader). New screens: text view in no training label. Held-out apps: their held-out screenshots, each read by a reader aligned without that app (\cref{sec:scope}). Bottom: errors by type on the held-out screenshots (counts) and extra elements, which match nothing on the screen (\% of the reader's output).}
  \label{tab:reader}
\begin{tabularx}{\linewidth}{@{}Xrrr@{}}
    \toprule
                                     &                    & \multicolumn{2}{c}{Reader} \\
    \cmidrule(l){3-4}
    Evaluation set                   & Screens            & Original         & Aligned \\
    \midrule
    \multicolumn{4}{@{}l}{\textit{Element error rate (\%)\,$\downarrow$}} \\
    Held-out screenshots             & \gateShots         & \errGateOrig     & \textcolor{ourblue}{\textbf{\errGateAligned}} \\
    Held-out, new screens only       & \gateNewShots      & \errGateNewOrig  & \textcolor{ourblue}{\textbf{\errGateNewAligned}} \\
    \otherTmplCount{} templates outside calibration & \otherTmplShots & \errOtherOrig & \textcolor{ourblue}{\textbf{\errOtherAligned}} \\
    \otherTmplCount{} templates, new screens only & \otherTmplNewShots & \errOtherNewOrig & \textcolor{ourblue}{\textbf{\errOtherNewAligned}} \\
    Confirmation set                 & \confScreensOrig{} / \confScreensAligned & \errConfOrig & \textcolor{ourblue}{\textbf{\errConfAligned}} \\
    \midrule Held-out apps                    & \heldGateShots     & \heldErrOrig     & \textcolor{ourblue}{\heldErrHeld} \\
    \midrule
    \multicolumn{4}{@{}l}{\textit{Held-out screenshots, errors by type\,$\downarrow$}} \\
    Missed elements (count)          &                    & \typeMissOrig    & \textcolor{ourblue}{\textbf{\typeMissAligned}} \\
    Wrong text (count)               &                    & \typeTextOrig    & \textcolor{ourblue}{\textbf{\typeTextAligned}} \\
    Wrong state (count)              &                    & \typeStateOrig   & \textcolor{ourblue}{\textbf{\typeStateAligned}} \\
    Extra elements (\%)              &                    & \extraGateOrig   & \textcolor{ourblue}{\textbf{\extraGateAligned}} \\
    \bottomrule
  \end{tabularx}
\end{table}

\paragraph{Element errors.}
\label{sec:reader_side}
Alignment lowers the element error rate on every set we measured, including screens that the calibration set does not contain (\cref{tab:reader}).
On \gateShots{} held-out screenshots, from episodes recorded after the calibration set and used in no training, the element error rate falls from \errGateOrig\% to \errGateAligned\%; with two further alignment seeds, the rate is nearly the same (supplementary \cref{tab:supp_errtypes}). The rate also falls on new screens, whose text view matches no training label, and on templates outside the calibration set, from apps that the set does contain (task success could not be compared there; supplementary \cref{supp:limitations}). On the held-out screenshots, alignment cuts every type of error, and the aligned reader outputs far fewer extra elements, which match nothing on the screen (\cref{tab:reader}, bottom). A reader that memorized its training labels could output text from them that is not on the current screen; on the confirmation set, the aligned reader does so about once per \extraFromTargetsPerScreens{} screens. 

\paragraph{Exact repeats.}
Recalling exactly repeated screens recovers much of the gain, but not all of it. Because MobileGym renders a screen identically across instances, \lookupHitPct\% of the screens encountered on the confirmation set exactly repeat a calibration screenshot. A lookup baseline returns the stored text view for these exact repeats and the reordered output of the original reader otherwise. It recovers \lookupGainPct\% of the gain of alignment, reordering included, and alignment remains clearly higher ($\dAlignLookup$ points; \cref{tab:main}). The lookup baseline's success falls as fewer of an episode's screens are exact repeats (supplementary \cref{tab:supp_newscreens}). 

\subsection{Alternatives and scope}
\label{sec:scope}

\paragraph{Adapting the policy with the same labels.}
The same labeled screenshots can instead adapt the policy to the original reader: the adapted policy learns to reproduce, from the original reader's output, what the clean-text policy does on the true text view (supplementary \cref{supp:practice}).
It gains \nosign{\dSameLabelClean} points over the clean-text policy with the original reader, and aligning the reader instead gains \nosign{\dAlignSameLabel} more, so given the same labeled screenshots, adapting the reader beats adapting the policy (\cref{tab:main}). One possible reason, which we did not test, is that the policy cannot click an element that the original reader misses; such actions, about \sameLabelDroppedPct\% of element actions, had to be dropped from its training data. 

\paragraph{A larger reader.}
A larger reader does not close the text-to-pixels gap.
An unaligned reader twice the size, Qwen3-VL-8B, reaches \succLargeClean\% on the same half-size screenshots and \succLargeFullClean\% on full-size ones. The text-to-pixels gap persists with this stronger configuration (\nosign{\gapLargeFull} points). The aligned 4B reader clearly exceeds both 8B configurations (\cref{tab:main}), at about \costRatioReadLargeFull{} of the 8B reader's cost per read on full-size screenshots, as timed with each reader alone on one GPU. Episodes with the aligned reader also take fewer steps, and an episode costs about \costRatioEpisodeLargeFull{} times as much as one with the 8B reader on full-size screenshots (supplementary \cref{supp:reader}). On half-size screenshots, the 8B reader misreads fewer elements of the held-out screenshots than the original 4B reader (\errGateLarge\% against \errGateOrig\%), yet its task success shows no measurable difference from that of the original 4B reader (supplementary \cref{tab:supp_intervals}). 

\begin{figure}[t]
  \centering
  \begin{tikzpicture}
\begin{axis}[paperplot, name=ea, width=2.4cm, height=3.2cm, xmode=log,xmin=80, xmax=2600, ymin=0, ymax=30, ytick={0,10,20,30},
  xtick={100,250,500,1000,2003}, xticklabels={100,250,500,1k,2k},
  xlabel={Alignment screenshots (log)}, ylabel={Element error rate (\%)},
  title={(a) Held-out screenshots}, title style={font=\sffamily\bfseries\fontsize{7}{8.2}\selectfont, yshift=-3pt}]
  \addplot[refgray, dashed, line width=0.5pt] coordinates {(80,\errGateOrig) (2600,\errGateOrig)};
  \node[anchor=north east, inner sep=1.5pt, text=refgray, align=right] at (axis cs:2600,\errGateOrig) {original\\reader \errGateOrig};
  \addplot[ourblue, line width=0.9pt, mark=*, mark size=1.5pt] coordinates {
    (\curveShotsOneHundred,\curveErrOneHundred) (\curveShotsTwoFifty,\curveErrTwoFifty) (\smallAlignShots,\curveErrFiveHundred)
    ({\numnc{\curveShotsOneThousand}},\curveErrOneThousand) ({\numnc{\alignTrainShots}},\errGateAligned)};
  \addplot[ourblue, densely dashed, line width=0.8pt, mark=*, mark size=1.5pt, mark options={solid, fill=white}] coordinates {
    (\curveShotsOneHundred,\curveNewErrOneHundred) (\curveShotsTwoFifty,\curveNewErrTwoFifty) (\smallAlignShots,\curveNewErrFiveHundred)
    ({\numnc{\curveShotsOneThousand}},\curveNewErrOneThousand) ({\numnc{\alignTrainShots}},\curveNewErrFull)};
\node[anchor=north east, inner sep=1pt, align=left, font=\sffamily\fontsize{7}{8.6}\selectfont] at (axis cs:2600,18.2) {\tikz[baseline=-0.6ex, x=1cm, y=1cm]{\draw[ourblue, line width=0.9pt] (0,0) -- (0.36,0); \fill[ourblue] (0.18,0) circle[radius=1.5pt];}~all\\
    \tikz[baseline=-0.6ex, x=1cm, y=1cm]{\draw[ourblue, densely dashed, line width=0.8pt] (0,0) -- (0.36,0); \filldraw[draw=ourblue, fill=white, line width=0.6pt] (0.18,0) circle[radius=1.5pt];}~pixel-new};
\end{axis}
\begin{axis}[paperplot, at={($(ea.east)+(1.2cm,0)$)}, anchor=west, width=2.85cm, height=3.2cm,
  xmin=0, xmax=4300, ymin=0, ymax=100, xtick={0,1000,2000,3000,4000}, xticklabels={0,1k,2k,3k,4k},
  ytick={0,25,50,75,100}, xlabel={Screenshots}, ylabel={Full-task success (\%)},
  title={(b) Confirmation set}, title style={font=\sffamily\bfseries\fontsize{7}{8.2}\selectfont, yshift=-3pt}]
  \addplot[refgray, densely dotted, line width=0.6pt] coordinates {(0,\succRealClean) (4300,\succRealClean)};
  \node[anchor=north west, inner sep=1.5pt, text=refgray] at (axis cs:0,\succRealClean) {no screenshots \succRealClean};
  \addplot[ourblue, line width=0.9pt, mark=*, mark size=1.6pt, error bars/.cd, y dir=both, y explicit, error bar style={solid}]
    coordinates {(\smallAlignShots,\succSmallAlignClean) +- (0,\succSmallAlignCleanStd)
                 ({\numnc{\alignTrainShots}},\succAlignClean) +- (0,\succAlignCleanStd)};
  \addplot[basered, densely dashed, line width=0.8pt, mark=square*, mark size=1.4pt, mark options={solid},
    error bars/.cd, y dir=both, y explicit, error bar style={solid}]
    coordinates {({\numnc{\budgetShots}},\succRealDirect) +- (0,\succRealDirectStd)
                 ({\numnc{\directThreeShotsApprox}},\succDirectThree) +- (0,\succDirectThreeStd)
                 ({\numnc{\directFourShotsApprox}},\succDirectFour) +- (0,\succDirectFourStd)};
  \node[anchor=north, inner sep=3pt, text=ourblue] at (axis cs:\smallAlignShots,\succSmallAlignClean) {\succSmallAlignClean};
  \node[anchor=south, inner sep=3pt, text=ourblue, font=\sffamily\bfseries\fontsize{7}{8.2}\selectfont]
    at (axis cs:{\numnc{\alignTrainShots}},\succAlignClean+\succAlignCleanStd) {\succAlignClean};
  \node[anchor=south, inner sep=1.5pt, text=basered] at (axis cs:{\numnc{\budgetShots}},\succRealDirect+\succRealDirectStd) {\succRealDirect};
  \node[anchor=south, inner sep=1.5pt, text=basered] at (axis cs:{\numnc{\directThreeShotsApprox}},\succDirectThree+\succDirectThreeStd) {\succDirectThree};
  \node[anchor=south, inner sep=1.5pt, text=basered] at (axis cs:{\numnc{\directFourShotsApprox}},\succDirectFour+\succDirectFourStd) {\succDirectFour};
  \node[anchor=south east, inner sep=2pt, align=left, font=\sffamily\fontsize{7}{8.6}\selectfont] at (axis cs:4300,2) {\tikz[baseline=-0.6ex, x=1cm, y=1cm]{\draw[ourblue, line width=0.9pt] (0,0) -- (0.42,0); \fill[ourblue] (0.21,0) circle[radius=1.6pt];}~reader alignment\\
    \tikz[baseline=-0.6ex, x=1cm, y=1cm]{\draw[basered, densely dashed, line width=0.8pt] (0,0) -- (0.42,0); \fill[basered] (0.18,-0.05) rectangle (0.24,0.05);}~direct training};
\end{axis}
\end{tikzpicture} \caption{\textbf{Alignment already helps with \smallAlignShots{} screenshots and keeps improving with more.} (a)~Element error rate on the \gateShots{} held-out screenshots against the calibration screenshots used for alignment, on all held-out screenshots and on the pixel-new ones (pixel-identical to no alignment screenshot); gray dashed: original reader. (b)~Full-task success on the confirmation set against the screenshots spent (mean\,$\pm$\,std over two training seeds), for alignment with \smallAlignShots{} and all calibration screenshots and for direct screenshot training after its second, third and fourth practice rounds; dotted: no screenshots (clean-text policy, original reader).} \label{fig:data_efficiency}
\end{figure}
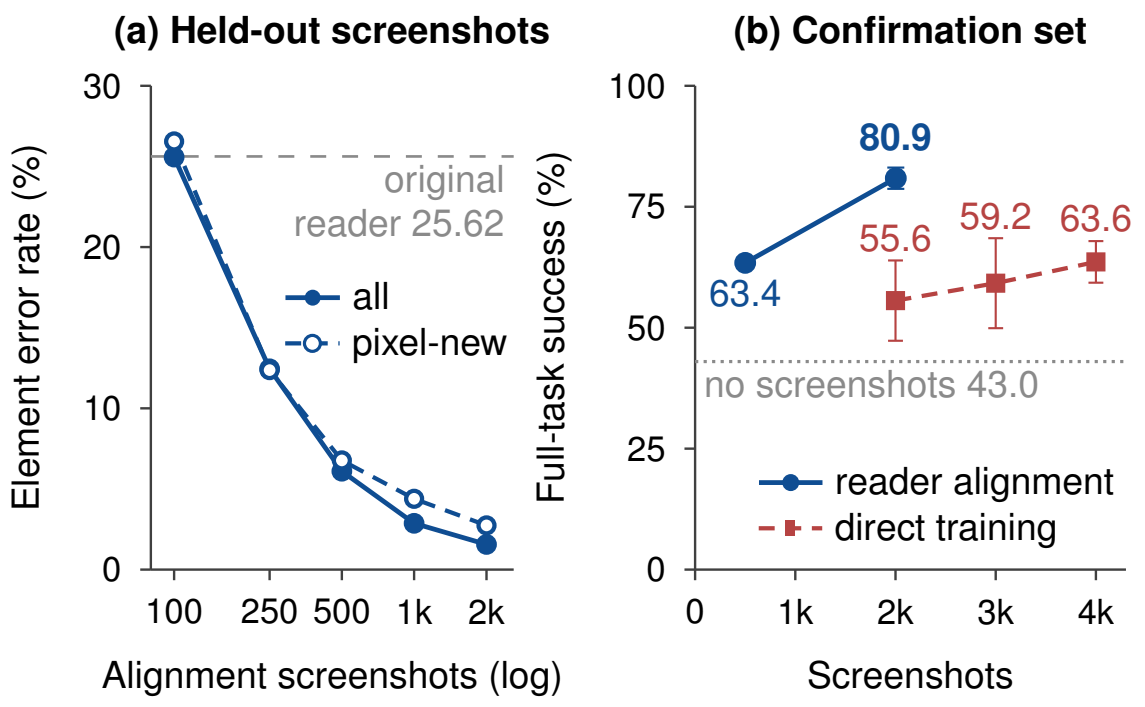

\paragraph{Data efficiency.}
With only \smallAlignShots{} screenshots, alignment reaches \succSmallAlignClean\%, \emph{comparable} to training the policy directly on about \budgetShots{} (\succRealDirect\%; the word \emph{comparable} was fixed in advance, supplementary \cref{supp:stats}). This success is already clearly higher than with the original reader, and with all calibration screenshots it rises to \succAlignClean\%, clearly higher than with \smallAlignShots. Even when direct training spends about \directFourShotsApprox{} screenshots, twice the alignment budget, alignment stays clearly higher ($\dAlignDirectFour$ points; \cref{fig:data_efficiency}b). On the confirmation set, the element error rate falls from \errConfOrig\% to \errConfSmall\% with \smallAlignShots{} screenshots and to \errConfAligned\% with all of them. On the held-out screenshots, the element error rate halves with alignment on \curveShotsTwoFifty{} screenshots (\cref{fig:data_efficiency}a). 

\paragraph{Held-out apps.}
On apps held out of alignment, the element error rate still falls, but task success shows no measurable difference.
For each of two apps chosen by a fixed rule (WeChat and Clock), we align a reader, the \emph{held-out-app reader}, without that app's calibration screenshots and test it on the app. On the held-out screenshots of these apps, the element error rate falls from \heldErrOrig\% to \heldErrHeld\%.
On their \heldTemplates{} single-app templates, task success is \heldSuccHeld\% with the held-out-app reader and \heldSuccOrig\% with the original reader, compared with \heldSuccFull\% with full alignment. The reader should therefore be aligned on screenshots of the apps the agent will use. 

\paragraph{Direct-training policy with the aligned reader.}
With the aligned reader and no retraining, the direct-training policy reaches \succDirectAligned\% (\cref{tab:reuse}). Policy and reader each used a budget of about \budgetShots{} screenshots, yet the pair shows no measurable difference from the clean-text policy with the aligned reader (supplementary \cref{tab:supp_intervals}). Once the reader is aligned, spending a second screenshot budget on the policy brings no measurable gain.  \section{Conclusion}
\label{sec:conclusion}
Policies succeed far less often on screenshots than in text because the screen reader misreads content and reorders elements.
The misreads are systematic, and on MobileGym about \budgetShots{} labeled screenshots suffice to align the reader to the text view of practice. Spent on the reader, a budget of this size gives clearly higher success than when spent on training the policy (\succAlignClean\% vs.\ \succRealDirect\%), and the aligned reader raises every text-trained policy we tried without retraining.
Aligning the reader is cheap and effective on this platform, whereas noise injected into practice would have to reproduce which elements the reader misreads. Supplementary \cref{supp:limitations} discusses the limitations.  
\ifdefined\arxivack
\section*{Acknowledgments}
\arxivack
\fi

{
    \small
    \bibliographystyle{ieeenat_fullname}

}

\clearpage
\makeatletter
\edef\arxiv@nextpage{\number\value{page}}
\let\arxiv@maketitlesupplementary\maketitlesupplementary
\def\maketitlesupplementary{\setcounter{page}{\arxiv@nextpage}\renewcommand{\thepage}{\arabic{page}}\arxiv@maketitlesupplementary}
\makeatother

\newcommand{\androidcontrolsource}{~\cite{li2024androidcontrol}}

\clearpage
\setcounter{page}{1}
\renewcommand{\thepage}{S\arabic{page}}
\maketitlesupplementary
\raggedbottom \renewcommand{\floatpagefraction}{0.8}

\setcounter{section}{0}
\renewcommand{\thesection}{\Alph{section}}
\renewcommand{\theHsection}{supp.\Alph{section}}
\setcounter{table}{0}
\renewcommand{\thetable}{S\arabic{table}}
\renewcommand{\theHtable}{supp.\arabic{table}}
\setcounter{figure}{0}
\renewcommand{\thefigure}{S\arabic{figure}}
\renewcommand{\theHfigure}{supp.\arabic{figure}}

This supplement gives the details behind the main paper: statistics (\cref{supp:stats}), the first-round results in full (\cref{supp:first_round}), per-template results (\cref{supp:per_template}), the predictivity test (\cref{supp:fidelity}), the gap decomposition (\cref{supp:decomp}), the screen reader (\cref{supp:reader}), the text view (\cref{supp:textview}), the practice environments (\cref{supp:practice}), cost and screen-reader accounting (\cref{supp:accounting}), the limitations (\cref{supp:limitations}), a second platform (\cref{supp:second_platform}) and further tests of reader alignment (\cref{supp:further}).

\section{Statistical details}
\label{supp:stats}

Unless stated otherwise, every comparison is a difference $\Delta = X - Y$, in percentage points, of full-task success on the confirmation set. All intervals are 95\% percentile bootstrap intervals. For each instance we first average success over the training seeds of each group; the bootstrap then resamples the \confTemplates{} templates with replacement \bootReps{} times, with both groups of a comparison sharing each draw. A second bootstrap also redraws the two training seeds with replacement in each draw; conditions that use the same policies with different readers share this draw, so they stay paired by seed. The bootstrap \emph{confirms} a difference when the intervals from both bootstraps exclude zero, and \cref{tab:supp_wording} gives the wording. For the recurrence of misreads (\cref{tab:persist}), the bootstrap instead resamples app groups \persistBoot{} times, where each app and each set of cross-app templates is one group (\persistApps{} groups; \persistAppsLarge{} for the 8B reader); for predictivity $M$, its gaps to the ceiling and the effects in \cref{tab:supp_decomp}, it resamples templates \fidBoot{} times. These have no second bootstrap, so their one interval decides. 

For the data-efficiency comparison (\cref{sec:scope}), the pre-specified rule uses the word \emph{comparable} when alignment with fewer screenshots shows no measurable difference from direct training, points toward alignment, or is higher. The comparisons with a decision rule fixed before the results were seen, called pre-specified and marked $^\ast$ in \cref{tab:supp_intervals}, are the five first-round comparisons (the text-to-pixels gap, judged in every group, and the four comparisons between groups of \cref{sec:simulating}), the comparisons in \cref{tab:main,tab:reuse} not marked $\dagger$, the recurrence comparisons of \cref{tab:persist}, the held-out-app and data-efficiency comparisons and the text-to-pixels gap of the full-size 8B reader (\cref{sec:scope}), and the replication on the fresh instance set (\cref{supp:first_round}). The two comparisons of \cref{supp:second_platform} and those marked $^\ast$ in \cref{tab:supp_more_readers} (\cref{supp:further}) are pre-specified as well. All other comparisons are descriptive, and no correction for multiple comparisons is applied. For the text-to-pixels gap, the pre-specified criterion was a gap of at least \gapCriterion{} points with a lower bound above zero in every group, and all five groups meet it. 

The tables give success as mean\,$\pm$\,standard deviation over the two training seeds of a group (sample standard deviation, with $n-1$ in the denominator). For the pre-specified comparisons between the aligned and the original reader, the standard deviation of the per-seed differences is \dAlignCleanStd{} points for the clean-text policy and \dAlignRateStd{} for the rate-matched-noise policy; there is one start policy, so it has none.
Intervals are given only for the pre-specified comparisons, in \cref{tab:supp_intervals,tab:supp_androidcontrol,tab:supp_more_readers}, and for the predictivity gaps, in \cref{tab:supp_fidelity}; all other differences are given without intervals. \cref{tab:supp_intervals,tab:supp_more_readers} show the first bootstrap's interval, because the second never changes the wording of a pre-specified comparison. Differences are computed from unrounded success rates, so they can differ slightly from the difference of the rounded rates shown.

\begin{table}[!tp]
  \centering
  \footnotesize
  \setlength{\tabcolsep}{2.5pt}
  \caption{\textbf{Comparisons in \cref{sec:findings,sec:results}.} Differences in points, colored as in \cref{tab:supp_wording}; intervals (templates resampled, app groups for recurrence) only for the pre-specified comparisons, marked $^\ast$. All other rows are descriptive.}
  \label{tab:supp_intervals}
\renewcommand{\arraystretch}{0.88}\begin{tabularx}{\linewidth}{@{}Lrr@{}}\toprule
    Comparison ($X - Y$) & $\Delta$ & Interval \\
    \midrule
    \tabblock{3}{Text $-$ pixels, original reader (\cref{tab:text_vs_real})} \\
    Clean-text practice$^\ast$ & \vup{\gapClean} & \ci{\gapCleanLo}{\gapCleanHi} \\
    Generic-noise practice$^\ast$ & \vup{\gapGeneric} & \ci{\gapGenericLo}{\gapGenericHi} \\
    Rate-matched-noise practice$^\ast$ & \vup{\gapRate} & \ci{\gapRateLo}{\gapRateHi} \\
    Direct screenshot training$^\ast$ & \vup{\gapDirect} & \ci{\gapDirectLo}{\gapDirectHi} \\
    Combined training$^\ast$ & \vup{\gapBoth} & \ci{\gapBothLo}{\gapBothHi} \\
    \addlinespace[2pt]
    \tabblock{3}{Between first-round groups (\cref{sec:simulating})} \\
    Direct training $-$ clean text$^\ast$ & \vup{\dDirectClean} & \ci{\dDirectCleanLo}{\dDirectCleanHi} \\
    Combined $-$ clean text$^\ast$ & \vup{\dBothClean} & \ci{\dBothCleanLo}{\dBothCleanHi} \\
    Rate-matched $-$ generic noise$^\ast$ & \vgray{\dRateGeneric} & \ci{\dRateGenericLo}{\dRateGenericHi} \\
    Rate-matched $-$ direct$^\ast$ & \vgray{\dRateDirect} & \ci{\dRateDirectLo}{\dRateDirectHi} \\
    Rate-matched $-$ clean text & \vgray{\dRateClean} & \\
    Generic noise $-$ clean text & \vgray{\dGenericClean} & \\
    Combined $-$ direct training & \vgray{\dBothDirect} & \\
    Combined $-$ rate-matched & \vgray{\dBothRate} & \\
    \addlinespace[2pt]
    \tabblock{3}{Recurrence of misreads (\cref{tab:persist}; $T$: temperature)} \\
    Qwen3-VL-4B$^\ast$, $T{=}\readerTemp$ & \vup{\persistSmallDiff} & \ci{\persistSmallLo}{\persistSmallHi} \\
    Qwen3-VL-8B$^\ast$, $T{=}\readerTemp$ & \vup{\persistLargeDiff} & \ci{\persistLargeLo}{\persistLargeHi} \\
    Qwen3-VL-4B$^\ast$, $T{=}\sampledTemp$ & \vup{\persistSampledDiff} & \ci{\persistSampledLo}{\persistSampledHi} \\
    InternVL3-8B$^\ast$, $T{=}\readerTemp$ & \vup{\persistOtherDiff} & \ci{\persistOtherLo}{\persistOtherHi} \\
    \addlinespace[2pt]
    \tabblock{3}{Main comparison (\cref{sec:main})} \\
    Aligned $-$ direct training$^\ast$ & \vup{\dMain} & \ci{\dMainLo}{\dMainHi} \\
    \quad on the first-round set$^\ast$ & \vup{\origMain} & \ci{\origMainLo}{\origMainHi} \\
    \quad on the fresh set$^\ast$ & \vup{\freshMain} & \ci{\freshMainLo}{\freshMainHi} \\
    Aligned, seed 2 $-$ seed 1 & \vgray{\dAlignSeedB} & \\
    Aligned, seed 2 $-$ direct & \vup{\dSeedBDirect} & \\
    \addlinespace[2pt]
    \tabblock{3}{Aligned $-$ original reader (\cref{sec:reuse})} \\
    Clean-text policy$^\ast$ & \vup{\dAlignClean} & \ci{\dAlignCleanLo}{\dAlignCleanHi} \\
    \quad on the first-round set & \vup{\origAlignClean} & \\
    Start policy$^\ast$ & \vup{\dAlignStart} & \ci{\dAlignStartLo}{\dAlignStartHi} \\
    Rate-matched-noise policy$^\ast$ & \vup{\dAlignRate} & \ci{\dAlignRateLo}{\dAlignRateHi} \\
    Untrained base policy & \vgray{\dAlignBase} & \\
    Direct-training policy & \vup{\dAlignDirect} & \\
    \addlinespace[2pt]
    \tabblock{3}{Between policies, aligned reader} \\
    Clean text $-$ start & \vgray{\dCleanStartAligned} & \\
    Clean text $-$ rate-matched & \vgray{\dCleanRateAligned} & \\
    Rate-matched $-$ start & \vgray{\dRateStartAligned} & \\
    Direct training $-$ clean text & \vgray{\dDirectCleanAligned} & \\
    \addlinespace[2pt]
    \tabblock{3}{Controls and alternatives (\cref{sec:results})} \\
    Reordered $-$ original output$^\ast$ & \vgray{\dReorder} & \ci{\dReorderLo}{\dReorderHi} \\
    Aligned $-$ lookup$^\ast$ & \vup{\dAlignLookup} & \ci{\dAlignLookupLo}{\dAlignLookupHi} \\
    Aligned, seed 2 $-$ lookup & \vgray{\dSeedBLookup} & \\
    Lookup $-$ reordered output & \vup{\dLookupReorder} & \\
    Aligned $-$ adapted policy$^\ast$ & \vup{\dAlignSameLabel} & \ci{\dAlignSameLabelLo}{\dAlignSameLabelHi} \\
    Aligned, seed 2 $-$ adapted & \vgray{\dSeedBSameLabel} & \\
    Adapted policy $-$ original & \vup{\dSameLabelClean} & \\
    Adapted policy $-$ direct & \vgray{\dSameLabelDirect} & \\
    Held-out-app reader $-$ original$^\ast$ & \vgray{\dHeldOrig} & \ci{\dHeldOrigLo}{\dHeldOrigHi} \\
    Aligned $-$ held-out-app reader & \vup{\dFullHeld} & \\
    \addlinespace[2pt]
    \tabblock{3}{Data efficiency, reader size (\cref{sec:scope})} \\
    \smallAlignShots{} screenshots $-$ direct$^\ast$ & \vgray{\dSmallVsDirect} & \ci{\dSmallVsDirectLo}{\dSmallVsDirectHi} \\
    Aligned $-$ \smallAlignShots{} screenshots & \vup{\dFullVsSmall} & \\
    \smallAlignShots{} screenshots $-$ original & \vup{\dSmallVsOrig} & \\
    Aligned $-$ direct after round 3 & \vup{\dAlignDirectThree} & \\ Aligned $-$ direct after round 4 & \vup{\dAlignDirectFour} & \\
    Aligned 4B $-$ 8B, half-size$^\ast$ & \vup{\dAlignLarge} & \ci{\dAlignLargeLo}{\dAlignLargeHi} \\
    Aligned 4B $-$ 8B, full-size$^\ast$ & \vup{\dAlignLargeFull} & \ci{\dAlignLargeFullLo}{\dAlignLargeFullHi} \\
    8B, half-size $-$ original 4B & \vgray{\dLargeOrig} & \\
    8B, full-size $-$ original 4B & \vgray{\dLargeFullOrig} & \\
    8B, half-size $-$ direct & \vgray{\dLargeDirect} & \\
    8B, full-size $-$ direct & \vgray{\dLargeFullDirect} & \\
    8B, full-size $-$ 8B, half-size & \vgray{\dLargeFullHalf} & \\
    Text $-$ pixels, 8B, half-size & \vup{\gapLarge} & \\
    Text $-$ pixels, 8B, full-size$^\ast$ & \vup{\gapLargeFull} & \ci{\gapLargeFullLo}{\gapLargeFullHi} \\
    \bottomrule
  \end{tabularx}
\end{table}

\begin{table}[!tp]
  \centering
  \footnotesize
  \setlength{\tabcolsep}{4pt}
  \caption{\textbf{Fixed wording for a difference $\Delta = X - Y$ (points), and its color in the tables.}}\label{tab:supp_wording}
\begin{tabularx}{\linewidth}{@{}llXl@{}}
    \toprule
    Bootstrap & $|\Delta|$ & Wording & Color \\
    \midrule
    confirms a rise  & $\geq\wordThreshold$ & $X$ is clearly higher             & \textcolor{upgreen}{\textbf{green}} \\
    confirms a rise  & $<\wordThreshold$    & $X$ is slightly higher            & \textcolor{upgreen}{\textbf{green}} \\
    does not confirm & $<\wordThreshold$    & no measurable difference          & \textcolor{black!50}{gray} \\
    does not confirm & $\geq\wordThreshold$ & points toward $X$ ($Y$ if $\Delta<0$) & \textcolor{black!50}{gray} \\confirms a drop  & --                   & as for a rise, with $X$ and $Y$ swapped & \textcolor{basered}{\textbf{red}} \\
    \bottomrule
  \end{tabularx}
\end{table}

\section{First-round results and replications}
\label{supp:first_round}

On screenshots, the two training seeds of direct screenshot training differ by about \directSeedSpread{} points, more than those of any other first-round group (\cref{tab:supp_seeds}; confirmation set, original reader). Of the first-round comparisons that were not pre-specified (\cref{tab:supp_intervals}), rate-matched-noise and generic-noise practice point toward higher success than clean-text practice, and combined training shows no measurable difference from direct screenshot training or from rate-matched-noise practice. On the first-round evaluation set, whose instances are disjoint from those of the confirmation set, all five pre-specified comparisons have the same sign as on the confirmation set (\cref{tab:supp_second_set}). This set crosses \origTemplates{} templates with instance seeds \origSeedFirst--\origSeedLast, giving \origInstances{} instances, and each group has two training seeds (\origEpisodes{} episodes). 

Reader alignment was later tested on this set (\cref{sec:main}). With the aligned reader, the clean-text policy reaches \origSuccAlignClean\%, clearly higher than direct screenshot training (\origSuccRealDirect\%; $\origMain$ points); it is also \nosign{\origAlignClean} points above the same policy with the original reader (\origSuccRealClean\%), a descriptive comparison.
None of the calibration screenshots or of the held-out screenshots of \cref{sec:reader_side} come from these instances. The numbers for direct screenshot training and for the original reader come from the first-round runs on these instances and were not rerun; the aligned-reader run differs from the clean-text policy's first-round run only in the reader, and its decision rule was fixed before it started. 

The fresh instance set (\cref{sec:main}) tests the main comparison on \freshInstances{} new instances: the \freshTemplates{} templates of the confirmation set with instance seeds that no earlier run had used. Both groups use the policies, readers and run settings of the main comparison, and each training seed is run once on every instance. With the aligned reader, the clean-text policy reaches \freshSuccAlignClean\%, clearly higher than direct screenshot training (\freshSuccRealDirect\%; $\freshMain$ points; \cref{tab:supp_intervals}). Per training seed, the aligned reader gives \freshSuccAlignCleanSeedA\% and \freshSuccAlignCleanSeedB\%, and direct screenshot training \freshSuccRealDirectSeedA\% and \freshSuccRealDirectSeedB\%. The instances, the statistics and the decision rule were fixed before the runs started; the runs made \freshReads{} reader calls, counted as evaluation in \cref{tab:supp_accounting}. 

\begin{table}[!tp]
  \centering
  \footnotesize
  \setlength{\tabcolsep}{4pt}
  \caption{\textbf{The two training seeds differ most for direct screenshot training.} First-round groups on the confirmation set: full-task success (\%) on screenshots with the original reader, per training seed and on average, and in text (mean of the two seeds).}
  \label{tab:supp_seeds}
\begin{tabularx}{\linewidth}{@{}Xcccc@{}}
    \toprule
                                & \multicolumn{3}{c}{Screenshots (\%)\,$\uparrow$}       & Text (\%)\,$\uparrow$ \\
    \cmidrule(lr){2-4}\cmidrule(l){5-5}
    Training group              & Seed 1                & Seed 2                & Mean             & Mean \\
    \midrule
    Clean-text practice         & \seedCleanA           & \seedCleanB           & \succRealClean   & \succTextClean \\
    Generic-noise practice      & \seedGenericA         & \seedGenericB         & \succRealGeneric & \succTextGeneric \\
    Rate-matched-noise practice & \seedRateA            & \seedRateB            & \succRealRate    & \succTextRate \\
    Direct screenshot training  & \textbf{\seedDirectA} & \textbf{\seedDirectB} & \succRealDirect  & \succTextDirect \\
    Combined training               & \seedBothA            & \seedBothB            & \succRealBoth    & \succTextBoth \\
    \bottomrule
  \end{tabularx}
\end{table}

\begin{table}[!tp]
  \centering
  \footnotesize
  \setlength{\tabcolsep}{4pt}
  \caption{\textbf{All five pre-specified first-round comparisons have the same sign on both instance sets.} Differences $X - Y$ in points, colored as in \cref{tab:supp_wording}. Last row: range of the five text-to-pixels gaps.}
  \label{tab:supp_second_set}
\begin{tabularx}{\linewidth}{@{}Lrr@{}}\toprule
    Comparison ($X - Y$)         & First-round set          & Confirmation set \\
    \midrule
    Direct $-$ clean text        & \vup{\origDirectClean}   & \vup{\dDirectClean} \\
    Combined $-$ clean text      & \vup{\origBothClean}     & \vup{\dBothClean} \\
    Rate-matched $-$ generic     & \vgray{\origRateGeneric} & \vgray{\dRateGeneric} \\
    Rate-matched $-$ direct      & \vgray{\origRateDirect}  & \vgray{\dRateDirect} \\
    Text $-$ pixels, range       & \origGapMin--\origGapMax & \nosign{\gapBoth}--\nosign{\gapClean} \\
    \bottomrule
  \end{tabularx}
\end{table}

\section{Per-template results}
\label{supp:per_template}

\cref{tab:supp_per_template} lists success per template for the main comparison. Even in text, the clean-text policy completes \tmplSendMsgText\% of \texttt{Send\allowbreak Message\allowbreak ToContact}, for which no scripted demonstration succeeded (\cref{supp:practice}), and \tmplDateIntervalText\% of \texttt{Calculate\allowbreak Date\allowbreak Interval}. \begin{table*}[!tp]
  \centering
  \footnotesize
  \caption{\textbf{Per template, the aligned reader is higher than direct screenshot training on \tmplBetter{} templates, equal on \tmplSame{} and lower on \tmplWorse.} Full-task success (\%) on the confirmation set, over both training seeds: the clean-text policy in text, with the original reader and with the aligned reader (blue), and direct screenshot training with the original reader. Every template has the same number of instances, so the last row, the mean over templates, equals the overall success.}
  \label{tab:supp_per_template}
\begin{tabular}{@{}llrr>{\color{ourblue}}rr@{}}
    \toprule
        &          & \multicolumn{3}{c}{Clean-text policy}                         & Direct training \\
    \cmidrule(lr){3-5}\cmidrule(l){6-6}
    App & Template & Text & Original & \textbf{Aligned} & Original \\
    \midrule
Alipay & \texttt{CheckBalance} & 100.0 & 95.8 & 100.0 & 100.0 \\
 & \texttt{ConfigureLanguageAndFastPay} & 100.0 & 70.8 & 100.0 & 87.5 \\
 & \texttt{SendMessageToContact} & 0.0 & 8.3 & 0.0 & 8.3 \\
Calendar & \texttt{CalculateDateInterval} & 8.3 & 12.5 & 8.3 & 16.7 \\
 & \texttt{CompareScheduleDensity} & 54.2 & 25.0 & 62.5 & 58.3 \\
Clock & \texttt{AddWorldCity} & 100.0 & 95.8 & 79.2 & 100.0 \\
 & \texttt{CityLocalTimeDiff} & 83.3 & 83.3 & 87.5 & 95.8 \\
 & \texttt{DeleteAlarm} & 79.2 & 79.2 & 79.2 & 87.5 \\
 & \texttt{LatestTimezoneCity} & 100.0 & 100.0 & 100.0 & 100.0 \\
Cross-app & \texttt{AlipayMonthlySpendToWechat} & 50.0 & 0.0 & 50.0 & 0.0 \\
 & \texttt{CalendarEventToWechat} & 62.5 & 0.0 & 54.2 & 0.0 \\
 & \texttt{CalendarEarliestToAlarm} & 100.0 & 0.0 & 0.0 & 0.0 \\
Notes & \texttt{PinNote} & 100.0 & 87.5 & 100.0 & 87.5 \\
 & \texttt{ReadNotesCount} & 100.0 & 20.8 & 100.0 & 100.0 \\
 & \texttt{TodoBatchWorkflow} & 100.0 & 12.5 & 100.0 & 41.7 \\
Railway 12306 & \texttt{CheckDefaultPassengerName} & 100.0 & 12.5 & 100.0 & 12.5 \\
SMS & \texttt{DeleteConversation} & 100.0 & 91.7 & 100.0 & 95.8 \\
 & \texttt{ReadUnreadConversationCount} & 100.0 & 33.3 & 100.0 & 37.5 \\
Tencent Meeting & \texttt{ChatInMeeting} & 100.0 & 12.5 & 79.2 & 50.0 \\
 & \texttt{ChatWithSpecificUser} & 100.0 & 0.0 & 45.8 & 41.7 \\
 & \texttt{CheckPersonalRoomId} & 100.0 & 33.3 & 100.0 & 62.5 \\
WeChat & \texttt{EnableDarkMode} & 100.0 & 70.8 & 100.0 & 100.0 \\
 & \texttt{OpenBlacklist} & 100.0 & 100.0 & 100.0 & 91.7 \\
 & \texttt{OpenNewFriends} & 100.0 & 83.3 & 100.0 & 100.0 \\
 & \texttt{SetAddMeSearch} & 100.0 & 8.3 & 100.0 & 50.0 \\
 & \texttt{SetSignature} & 100.0 & 41.7 & 100.0 & 54.2 \\
WeChat Reading & \texttt{AddBookToShelf} & 100.0 & 29.2 & 100.0 & 25.0 \\
 & \texttt{CheckBookRating} & 100.0 & 12.5 & 100.0 & 4.2 \\
 & \texttt{CheckCalendarMonthReading} & 100.0 & 25.0 & 100.0 & 4.2 \\
\midrule
\multicolumn{2}{@{}l}{Mean} & \succTextClean & \succRealClean & \succAlignClean & \succRealDirect \\
    \bottomrule
  \end{tabular}
\end{table*}

\paragraph{Remaining failures.}
For the failed episodes of the four templates that account for most of the remaining text-to-pixels gap (\texttt{Calendar\allowbreak Earliest\allowbreak ToAlarm}, \texttt{Chat\allowbreak With\allowbreak Specific\allowbreak User}, \texttt{Chat\allowbreak In\allowbreak Meeting} and \texttt{Add\allowbreak World\allowbreak City}), a procedure fixed in advance finds the first discrepancy between the aligned reader's output and the true text view (\cref{tab:supp_failures}); this discrepancy need not be the only cause of a failure. The procedure walks through each failed episode step by step and stops at the first of three discrepancies: (i)~the policy acts on an element that the reader misread or invented; (ii)~an element that the policy used in the successful text episode of the same instance is misread or missed; or (iii)~such an element is read correctly, but its place among the elements that both lists contain differs from the layout order. A line that the parser could not read counts as a format difference. If none occurs, the failure is attributed to the policy when the text episode failed too, to the executor when a tap on a correctly read element that the task needs left the screen unchanged, and otherwise to the platform. All \failTotal{} failed episodes succeeded in text, and none is attributed to format, executor, policy or platform. The \failFieldFirst{} misreads are all of kind (i) and all concern input fields; in each, the element that the policy acts on matches no element of the true text view.

\begin{table}[!tp]
  \centering
  \footnotesize
  \caption{\textbf{In every failed episode of the four templates with most of the remaining gap, the first discrepancy is an ordering difference or a misread input field.} First discrepancy between the aligned reader's output and the true text view (clean-text policy, aligned reader).}
  \label{tab:supp_failures}
\begin{tabularx}{\linewidth}{@{}Xr@{}}
    \toprule
    First discrepancy        & Failed episodes \\
    \midrule
    Ordering difference      & \failOrderFirst \\
    Misread input field      & \failFieldFirst \\
    \midrule
    Total                    & \failTotal \\
    \bottomrule
  \end{tabularx}
\end{table}

\section{Predictivity test}
\label{supp:fidelity}
\begin{table*}[!tp]
  \centering
  \footnotesize
  \setlength{\tabcolsep}{6pt}
  \caption{\textbf{Every practice environment stays below the ceiling set by the real pipeline.} Full results of the predictivity test (\cref{fig:fidelity}): predictivity $M$, its gap to the real pipeline's own $M$ with the gap's interval, and the start policy's aggregate full-task success (\%) on the instances of \cref{sec:decomp}. The bootstrap confirms every gap. Errors: \injyes{} the environment injects the reader's errors, \injno{} it does not.}
  \label{tab:supp_fidelity}
\begin{tabular}{lccccc}
    \toprule
                                    &        &                     & \multicolumn{2}{c}{Gap to the ceiling\,$\downarrow$}  & \\
    \cmidrule(lr){4-5}
    Environment                     & Errors & $M$\,$\uparrow$     & $M_{\text{real}} - M$    & Interval                                                  & Success (\%) \\
    \midrule
    \rowcolor{refgray!18}Real pipeline (ceiling) & -- & \fidReal & --                       & --                                                        & \decompSuccReal \\
    Clean text                      & \injno  & \fidClean           & \fidPairClean            & \ci{\fidPairCleanLo}{\fidPairCleanHi}                     & \fidSuccClean \\
    Reader's order only             & \injno  & \fidOrderOnly       & \fidPairOrderOnly        & \ci{\fidPairOrderOnlyLo}{\fidPairOrderOnlyHi}             & \fidSuccOrderOnly \\
    Generic noise                   & \injno  & \fidRandom          & \fidPairRandom           & \ci{\fidPairRandomLo}{\fidPairRandomHi}                   & \fidSuccRandom \\
    Rate-matched noise              & \injyes & \fidRate            & \fidPairRate             & \ci{\fidPairRateLo}{\fidPairRateHi}                       & \fidSuccRate \\
    Replay, fixed reads             & \injyes & \fidReplayOne       & \fidPairReplayOne        & \ci{\fidPairReplayOneLo}{\fidPairReplayOneHi}             & \fidSuccReplayOne \\
    Replay, resampled reads         & \injyes & \fidReplayTwo       & \fidPairReplayTwo        & \ci{\fidPairReplayTwoLo}{\fidPairReplayTwoHi}             & \fidSuccReplayTwo \\
    Replay, resampled, layout order & \injyes & \fidReplayTwoLayout & \fidPairReplayTwoLayout  & \ci{\fidPairReplayTwoLayoutLo}{\fidPairReplayTwoLayoutHi} & \fidSuccReplayTwoLayout \\
    \bottomrule
  \end{tabular}
\end{table*}

The two versions of the element-replay simulator, with fixed and with resampled reads, each failed a pre-specified predictivity test before any training (\cref{sec:simulating}); the two tests differ. In \cref{tab:supp_fidelity}, which uses the environment names of \cref{fig:fidelity}, \emph{Replay, resampled, layout order} lists the elements of the resampled-read version in the layout order, and \emph{Reader's order only} lists the true elements in the reader's order, without misreads (\cref{supp:decomp}).
For each environment, $M$ is the Pearson correlation, across the \confTemplates{} templates, between two per-template success rates, the start policy's in the environment on the \decompInstances{} instances of \cref{sec:decomp} (\decompSamples{} samples each) and the real pipeline's on the \calibEpisodes{} calibration episodes, other instances of the same templates. For the ceiling, the first rate is the real pipeline's own on the \decompInstances{} instances. The test of the resampled-read version, fixed before that version was evaluated, required $M \geq \fidLineM$ and aggregate success within \fidLinePoints{} points of the real pipeline's \decompSuccReal\%. That version reached $M=\fidReplayTwo$ ($\fidReplayTwoDiff$ points in aggregate) and failed. The fixed-read version had failed an earlier test with other thresholds and data. It compared \fidFirstEpisodes{} episodes of the start policy in this version, on training instances, with all \fidFirstRealEpisodes{} real-pipeline episodes of the start policy (the \calibEpisodes{} calibration episodes and the \decompInstances{} on the unseen instances of \cref{sec:decomp}). The per-template Pearson correlation was \fidFirstCorr{}, below the threshold of \fidFirstLineCorr{}, while aggregate success was within \fidFirstLinePoints{} points ($\fidFirstDiff$). The fixed-read version's $M$ in \cref{tab:supp_fidelity} was computed afterwards, as in the test of the resampled-read version. 

Both versions replay the original reader's reads of the \calibShots{} calibration screenshots. The fixed-read version writes a screen whose true text view occurs among them as the reader wrote it there. On other screens, each element seen in the calibration screenshots (same app, role, text, icon name and state, box coordinates rounded to multiples of \matchGrid) is written as the reader wrote it in one of its calibration reads, drawn in proportion to how often each occurred: left out, written as a mark or an icon, merged with its right neighbor, or kept with the reader's role, text, icon name, state and box offset, together with any extra elements of that read. An element never seen gets the draws of rate-matched noise for its element category (\cref{supp:practice}). Each draw is determined by the element alone with probability \replayFixedProb{} and otherwise by the element and the screen, never by the episode, so a screen is written the same way in every episode. The resampled-read version changes two things. A screen whose true text view occurs among the calibration screenshots gets one of its calibration reads, drawn anew in every episode but fixed within it. On other screens, the list takes the order of the calibration read from the same app that shares the most elements with it if that read shares at least half of them, and the layout order otherwise.

\section{Gap decomposition}
\label{supp:decomp}

The effects of \cref{sec:decomp} are in \cref{tab:supp_decomp}. 

\begin{table}[!htb]
  \centering
  \footnotesize
  \setlength{\tabcolsep}{4pt}
  \caption{\textbf{Misreads and the reader's order each lower success, and their costs overlap.} Effects in the gap decomposition (\cref{sec:decomp}): $\Delta = X - Y$ in points of full-task success of the start policy on \decompInstances{} unseen instances with \decompSamples{} samples each, colored as in \cref{tab:supp_wording}.}
  \label{tab:supp_decomp}
\begin{tabularx}{\linewidth}{@{}Xr@{}}\toprule
    Comparison ($X - Y$)                      & $\Delta$ \\
    \midrule
    Reader's vs.\ layout order, no misreads   & \vdown{\effOrderClean} \\
    Reader's vs.\ layout order, with misreads & \vgray{\effOrderGivenMisread} \\
    Misreads vs.\ none, layout order          & \vdown{\effMisreadLayout} \\
    Misreads vs.\ none, reader's order        & \vdown{\effMisreadReaderOrder} \\
    Reader's vs.\ random order, no misreads   & \vup{\effReaderVsShuffle} \\
    \bottomrule
  \end{tabularx}
\end{table}

\paragraph{Construction.}
The decomposition uses the resampled-read version of the element-replay simulator (\cref{supp:fidelity}). Misreads without the reader's order are its reads sorted into the layout order; the reader's order without misreads lists the true elements in the simulator's order; and the random order shuffles the true elements, with a shuffle drawn from the episode and the screen, so that an unchanged screen keeps its order within an episode. 

\paragraph{After training.}
The decomposition was repeated for the ten trained first-round policies (five groups, two training seeds each) on the confirmation set, in two more text environments: the true elements in the reader's order, and the element-replay simulator with resampled reads, which also injects the reader's misreads. Both take the order from the original reader's calibration reads, as the resampled-read version does (\cref{supp:fidelity}): the read of the same screen (\tdOrderSameScreenPct\% of the \tdScreens{} screens of these episodes), else that of the calibration screen of the same app that shares the most elements with it (\tdOrderSimilarPct\%), else the layout order (\tdOrderTopDownPct\%).
The clean text view and the real pipeline reuse the confirmation runs, so these runs made no new reader calls. Of their \tdEpisodes{} episodes, \tdErrors{} ended with a prompt longer than the policy's context and count as failures; all of them ran on the true elements in the reader's order, in three calendar templates.
In every group, the reader's order costs \tdOrderMin--\tdOrderMax{} points on the true elements, and the simulator's misreads cost another \tdMisreadMin--\tdMisreadMax{} points.
Between direct screenshot training and clean-text practice, the cost of the order shows no measurable difference ($\tdOrderDirectClean$ points).
Success with the simulator is \magn{\tdRestMax}--\magn{\tdRestMin} points below the real pipeline's, clearly lower in four groups; for rate-matched-noise practice, the comparison points toward the real pipeline. The two costs together therefore exceed the whole text-to-pixels gap, and the simulator also failed the predictivity test (\cref{supp:fidelity}).
All comparisons in this paragraph are descriptive.

\section{Screen reader}
\label{supp:reader}

\paragraph{Choice of the configuration.}
The reader configuration was chosen with a scripted policy, a fixed program that decides each step from the true page but may act only on the elements that the reader lists. With the reader, an episode succeeds only if a scoring rule finds that every element the program needs was read correctly. A configuration's loss is the scripted policy's success in text minus its success with the reader, in points, with a lower bound from \readerPickBoot{} template resamples. The rule kept the original plan, Qwen3-VL-8B on full-size screenshots, if its loss was at least \readerPickLossLine{} points with a lower bound of at least \readerPickLowLine.
Otherwise it switched to the 4B reader on half-size screenshots, the cheapest of the \readerPickCandidates{} configurations we timed, if its loss was at least \readerPickLossLine{} points, its success with the reader at least \readerPickSuccLine\%, and its cost per read at most \readerPickCostLine{} times that of the 8B reader on full-size screenshots, both on the mean of three paired timings and in at least two of them. The switch was then to be confirmed with the same loss thresholds on the episodes of the 8B measurement. If neither reader qualified, the study would have moved to another platform.
The loss threshold of \readerPickLossLine{} points was set before the scripted policy was first run with the 8B reader, on \readerPickFirstEpisodes{} episodes; the thresholds for the lower bound, the success with the reader and the cost were set after that run, but before the loss measurements and timings to which the rule was applied. With the scripted policy, reading cost \readerPickLargeLoss{} points with the 8B reader on full-size screenshots (\readerPickLargeEpisodes{} episodes), below the threshold of \readerPickLossLine. With the 4B reader on half-size screenshots it cost \readerPickCheapLoss{} points on a separate set of \readerPickCheapEpisodes{} episodes, with \readerPickCheapSucc\% success with the reader; its cost per read is given below with the timings. In text, the scripted policy succeeded on all of these episodes. On the \readerPickLargeEpisodes{} episodes of the 8B measurement, the 4B configuration lost \readerPickCheapSameLoss{} points; these episodes were only partly fresh, because the 8B results on them had been inspected and the scoring rule was then revised, so this number is not an estimate on unseen episodes. 

The scoring of reads was revised twice after the 8B results on these episodes had been inspected. The first revision raised both readers' losses (the 8B reader's from \readerPickLargeLossAsRun{} to \readerPickLargeLossPrev{} points), partly by counting as misses reads whose content was present, and under it, as under the original scoring, the rule would have kept the 8B reader. The second revision rewrote that rule for matching reads to true elements: every change it made turned a miss into a correct read, and it lowered both readers' losses on these episodes by the same \readerPickRescoreDrop{} points, the 8B reader's to \readerPickLargeLoss{}, below the threshold of \readerPickLossLine{}. The move to the cheaper configuration therefore depends on the second revision. On the same episodes the 4B configuration lost \readerPickCheapSameLossAsRun{}, \readerPickCheapSameLossPrev{} and \readerPickCheapSameLoss{} points under the three scorings. 

The cost threshold was also applied to a revised measurement. Timed one read at a time, as first implemented, the 4B configuration took \readerPickSingleFourHalf{} GPU-seconds per read against \readerPickSingleEightFull{} for the 8B reader, more than half, and no cheaper configuration qualified. After these results had been inspected, we timed reads with \readerPickInflight{} in parallel. An earlier draft of the rule, written before the measurements but not the version that was run, had specified this, and all task runs issue reads concurrently. Timed that way, the ratio was \readerPickCostA, \readerPickCostB{} and \readerPickCostC{} in three paired measurements (mean \readerPickCostMean), just below the threshold of \readerPickCostLine. The move to the cheaper configuration therefore depends on this revision as well. 

\paragraph{Prompt, output format and parser.}
The original and the aligned reader use the same prompt, temperature (\readerTemp), repetition penalty (\readerRepPenalty), half-size input and parser (\cref{sec:method}). Each screenshot is halved in width and height with a Lanczos filter and read once per step, without retries, with at most \readerMaxTokens{} output tokens. The prompt is in Chinese, like the apps; this is our English translation: \par\smallskip
{\footnotesize\fontencoding{T1}\selectfont\raggedright\leftskip=1em\parindent=0pt
This is a screenshot of a phone screen. List everything visible on the screen, one line per element, from top to bottom and from left to right.\par\smallskip
One element per line, in this fixed format:\par
\begin{minipage}{\dimexpr\linewidth-1em\relax}\scriptsize\ttfamily
text "text on the screen" color (x1,y1,x2,y2)\\
icon icon-name color (x1,y1,x2,y2)\\
image (x1,y1,x2,y2)\\
field "text shown in the box" empty (x1,y1,x2,y2)\\
switch on color (x1,y1,x2,y2)\\
switch off color (x1,y1,x2,y2)\\
mark filled color (x1,y1,x2,y2)\\
mark empty color (x1,y1,x2,y2)\end{minipage}\par\smallskip
Rules:\par
\hangindent=0.8em\hangafter=1 -- The coordinates are the element's bounding box, normalized to \texttt{0-1000} by the screen's width and height; x is horizontal, y is vertical.\par
\hangindent=0.8em\hangafter=1 -- The text of a text element must match the screen exactly: do not translate, complete or rewrite it; keep the ellipsis of truncated text.\par
\hangindent=0.8em\hangafter=1 -- When the background is neither white nor black, add \texttt{on-<background color>} after the color, e.g.\ \texttt{white on-blue}; add \texttt{bold} for bold text and \texttt{dim} for clearly faded elements.\par
\hangindent=0.8em\hangafter=1 -- Use only these colors: black gray lightgray white red orange yellow green cyan blue purple pink.\par
\hangindent=0.8em\hangafter=1 -- \texttt{icon} is a small icon without text; give its name as a short English word, e.g.\ \texttt{search}, \texttt{plus}, \texttt{chevron-right}, \texttt{settings}; leave the name out if you cannot tell.\par
\hangindent=0.8em\hangafter=1 -- \texttt{image} is a photo or a large picture. \texttt{field} is an input box; add \texttt{empty} when the box shows only a placeholder. \texttt{switch} is a toggle. \texttt{mark} is a checkbox, a radio circle or a small dot.\par
\hangindent=0.8em\hangafter=1 -- Write only what is visible on the screen: no explanations, no numbering, no Markdown.\par}
\smallskip
Although the prompt asks for the elements from top to bottom and from left to right, the original reader's order differs from the layout order on \readerOrderDiffPct\% of the calibration and held-out screenshots (its cost: \cref{sec:decomp}). The parser reads the output line by line, strips whitespace and backquotes and skips empty lines. A line must start with a role word, possibly after an index such as \texttt{[n]}, and contain a box \texttt{(x1,y1,x2,y2)}; the last box on the line is used, and the attributes after it are still read. Coordinates are clipped to 0--\genericJitterScale. Lines without this form and lines whose box has a width or height of zero or less are unparsable; they are counted and dropped. A role outside the six of the prompt (for example \texttt{button}) becomes \texttt{text} if the line has quoted text and \texttt{icon} otherwise. The first quoted string, with JSON escapes undone, is the element's text; a \texttt{text} or \texttt{field} element without one has empty text. Of the other words, a color name gives the color, \texttt{on-}color the background, and \texttt{bold}, \texttt{dim}, \texttt{empty}, \texttt{on}/\texttt{off} and \texttt{filled}/\texttt{empty} the flags and states of their roles; the first other word of an \texttt{icon} is its name. A \texttt{switch} without a state counts as off and a \texttt{mark} as empty. Elements are numbered in the order in which the reader wrote them, and the policy sees this order.
In task runs the policy sees the parsable lines, or an empty list when no line parses, and an output cut at the token limit is used up to the cut (how often this happens: \cref{supp:accounting}). For the acceptance test below, a screenshot from which no element can be parsed counts as unparsable, and all its elements count as missed. 

\begin{figure*}[t]
  \centering
  \includegraphics[width=\linewidth]{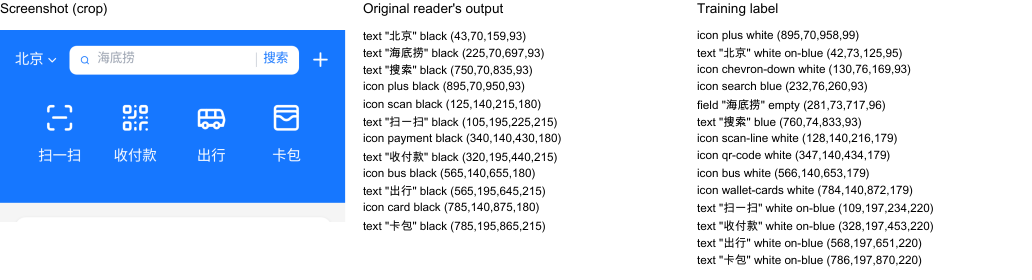}
  \caption{\textbf{The original reader's output differs from the training label in content and in order.} Left to right, for the part of the screen of \cref{fig:errors} just below the status bar: the screenshot, the original reader's output and the training label (the text view of this part in the reader's output format and the layout order). The original reader writes every color as black and leaves out the blue background, misses the drop-down arrow and the search icon, writes the search field as text, uses other icon names (\texttt{scan}, \texttt{payment} and \texttt{card} for \texttt{scan-line}, \texttt{qr-code} and \texttt{wallet-cards}) and lists the icons and their labels column by column; the training label lists the row of icons before the row of labels, and the plus sign comes first because its top edge falls in a higher row band (\cref{supp:textview}). The interface text is Chinese and is shown as it appears on the screen.}
  \label{fig:supp_align_example}
\end{figure*}

\paragraph{Training labels and settings.}
Each training label is the text view of its screenshot, written in the reader's output format and in the layout order (\cref{sec:method}).
\cref{fig:supp_align_example} compares the original reader's output with the training label on the screen of \cref{fig:errors}, for the elements whose box centers lie in the quarter of the screen height just below the status bar; this part of the screen was fixed before the figure was drawn. Apart from the base model, reader alignment uses the LoRA and optimizer settings of policy training (\cref{supp:practice}). Alignment trains on \alignTrainShots{} of the \calibShots{} calibration screenshots, skipping \alignSkippedShots{} screenshot whose training sample exceeds the length limit (\cref{supp:practice}). 

\paragraph{Acceptance test.} Before any policy was evaluated with it, the aligned reader had to pass a pre-specified acceptance test on a separate set of \gateShots{} held-out screenshots, from start-policy episodes collected after the calibration set was complete; no group used them.
The test required the element error rate to fall by at least \gateMinReductionPct\% relative to the original reader, with at most \gateMaxUnparsablePct\% of screenshots unparsable.
The aligned reader passed it on the first attempt, with an element error rate of \errGateAligned\% against \errGateOrig\% for the original reader and no unparsable screenshot (\cref{sec:reader_side}). 

\paragraph{Data efficiency.}
The element error rate falls with the number of calibration screenshots used for alignment (\cref{fig:data_efficiency}a and \cref{sec:scope}). The readers of the curve are aligned on the first \curveShotsOneHundred, \curveShotsTwoFifty, \smallAlignShots{} and \curveShotsOneThousand{} calibration screenshots, in the order of the episodes, with the seed and hyperparameters of full alignment (on all calibration screenshots, which gives the main aligned reader), and each reads the \gateShots{} held-out screenshots. With \curveShotsOneHundred{} screenshots the error rate does not move (\curveErrOneHundred\%, against \errGateOrig\% before alignment); one possible reason, which we did not test, is that this reader is trained for only \curveStepsOneHundred{} steps (\alignStepsFull{} with all \alignTrainShots). It is \curveErrTwoFifty\% with \curveShotsTwoFifty{} screenshots, \curveErrFiveHundred\% with \smallAlignShots, \curveErrOneThousand\% with \curveShotsOneThousand{} and \errGateAligned\% with all \alignTrainShots. In \cref{fig:data_efficiency}a, a held-out screenshot is pixel-new if it is not pixel-identical to any of the screenshots used for that point's alignment, so each point has its own pixel-new screens (\curveNewFull{} with all \alignTrainShots); on them the error rate falls from \curveNewErrOneHundred\% with \curveShotsOneHundred{} screenshots to \curveNewErrFull\% with all \alignTrainShots. The new screens of \cref{tab:reader} are stricter, because their text view also matches no training label; there are \gateNewShots{} of them, with an error rate of \errGateNewAligned\% after full alignment.
On the confirmation set, the element error rate falls from \errConfOrig\% with the original reader to \errConfSmall\% with \smallAlignShots{} screenshots and \errConfAligned\% with all \alignTrainShots. \cref{fig:data_efficiency}b also shows direct screenshot training continued for a third and a fourth practice round, each run like its second round with about \directRoundTwoShots{} more screenshots, to about \directThreeShotsApprox{} and \directFourShotsApprox{} screenshots. Its success on screenshots is \succDirectThree\% after the third practice round and \succDirectFour\% after the fourth, against \succRealDirect\% with about \budgetShots{}; in text it is \succTextDirectThree\% and \succTextDirectFour\%. The three comparisons with these practice rounds are descriptive. After the fourth practice round, direct screenshot training is clearly higher than after two practice rounds ($\dDirectFourVsTwo$ points).
The aligned reader, with about \budgetShots{} screenshots, is clearly higher than direct training after the third practice round ($\dAlignDirectThree$ points) and after the fourth ($\dAlignDirectFour$ points; \cref{sec:scope}).

\paragraph{Extra elements.}
The aligned reader rarely outputs elements that match nothing in the true text view, \extraConfAligned\% of its elements on the confirmation set against \extraConfOrig\% for the original reader. \cref{tab:supp_extra} gives this share on all screenshot sets and, for the confirmation set, how often extra text that is not on the screen appears in some training label. The original reader never saw these labels, so its share is a control for coincidental overlap. \cref{tab:supp_errtypes} gives all four error types on the held-out screenshots. On the confirmation set, the aligned reader outputs text that appears in a training label but not on the current screen \extraFromTargets{} times over the \confScreensAligned{} screens it read. 

\paragraph{Alignment seeds.}
Two more readers are aligned on all calibration screenshots with the same hyperparameters, changing only the alignment seed. On the held-out screenshots the three aligned readers err on \errGateAligned\%, \errGateAlignSeedB\% and \errGateAlignSeedC\% of elements (\cref{tab:supp_errtypes}), and on the pixel-new screens of \cref{fig:data_efficiency}a on \curveNewErrFull\%, \newErrAlignSeedB\% and \newErrAlignSeedC\%.
With the second reader, the clean-text policy reaches \succAlignSeedB\,$\pm$\,\succAlignSeedBStd\%, with no measurable difference from the main aligned reader (\succAlignClean\%; $\dAlignSeedB$ points), and it is still clearly higher than direct screenshot training (\succRealDirect\%; $\dSeedBDirect$ points). Its comparisons with the lookup baseline ($\dSeedBLookup$ points) and with adapting the policy with the same labels ($\dSeedBSameLabel$ points) point toward the second reader. All four comparisons are descriptive (\cref{tab:supp_intervals}); the last two were added after the results were seen, and the pre-specified comparisons use the first seed. \cref{supp:further} gives the third reader's task success. 

\begin{table}[!tp]
  \centering
  \footnotesize
  \caption{\textbf{The aligned reader outputs far fewer elements that match nothing on the screen.} Extra elements: output elements matching nothing in the true text view, in \% of the output. New screens as in \cref{tab:reader}. Last row, confirmation set: of the extra elements whose text is not on the screen, \% found in a training label.}
  \label{tab:supp_extra}
\begin{tabularx}{\linewidth}{@{}Xrr@{}}
    \toprule
                                                  & \multicolumn{2}{c}{Reader} \\
    \cmidrule(l){2-3}
    Evaluation set                                & Original            & Aligned \\
    \midrule
    \multicolumn{3}{@{}l}{\textit{Extra elements (\%)\,$\downarrow$}} \\
    Held-out screenshots                          & \extraGateOrig      & \textcolor{ourblue}{\textbf{\extraGateAligned}} \\
    Held-out, new screens only                    & \extraNewOrig       & \textcolor{ourblue}{\textbf{\extraNewAligned}} \\
    Confirmation set, screens encountered         & \extraConfOrig      & \textcolor{ourblue}{\textbf{\extraConfAligned}} \\
    \midrule
    Extra elements found in a training label (\%) & \extraInTargetsOrig & \extraInTargetsAligned \\
    \bottomrule
  \end{tabularx}
\end{table}

\begin{table}[!tp]
  \centering
  \footnotesize
  \setlength{\tabcolsep}{3pt}
  \caption{\textbf{Alignment cuts every type of error; with label elements deleted at random, more errors remain, mostly missed elements.} Errors by type on the \gateShots{} held-out screenshots (\gateElements{} true elements), half-size except in the full-size row. Aligned 4B: aligned on all calibration screenshots with alignment seed 1 (blue, the main aligned reader), 2 or 3; deleted: each label element deleted independently with this probability. Error rate: missed elements, wrong text and wrong state over all true elements; extra: as in \cref{tab:supp_extra}.}
  \label{tab:supp_errtypes}
\begin{tabularx}{\linewidth}{Xrrrrr}
    \toprule
                                    & Error rate              & \multicolumn{3}{c}{Errors (count)\,$\downarrow$}             & Extra \\
    \cmidrule(lr){3-5}
    Reader                          & (\%)\,$\downarrow$      & Missed             & Text               & State               & (\%)\,$\downarrow$ \\
    \midrule
    Original 4B                     & \errGateOrig            & \typeMissOrig      & \typeTextOrig      & \typeStateOrig      & \extraGateOrig \\
    Unaligned 8B, half-size         & \errGateLarge           & \typeMissLarge     & \typeTextLarge     & \typeStateLarge     & \extraGateLarge \\
    Unaligned 8B, full-size         & \errGateLargeFull       & \typeMissLargeFull & \typeTextLargeFull & \typeStateLargeFull & \extraGateLargeFull \\
    \midrule
    \rowcolor{rowblue}Aligned 4B, seed 1 & \textbf{\errGateAligned} & \typeMissAligned & \typeTextAligned & \typeStateAligned & \extraGateAligned \\
    Aligned 4B, seed 2              & \errGateAlignSeedB      & \typeMissSeedB     & \typeTextSeedB     & \typeStateSeedB     & \extraGateSeedB \\
    Aligned 4B, seed 3              & \errGateAlignSeedC      & \typeMissSeedC     & \typeTextSeedC     & \typeStateSeedC     & \extraGateSeedC \\
    \midrule
    Aligned 4B, \dropPctLow\% deleted  & \errGateDropLow      & \typeMissDropLow   & \typeTextDropLow   & \typeStateDropLow   & \extraGateDropLow \\
    Aligned 4B, \dropPctHigh\% deleted & \errGateDropHigh     & \typeMissDropHigh  & \typeTextDropHigh  & \typeStateDropHigh  & \extraGateDropHigh \\
    \bottomrule
  \end{tabularx}
\end{table}

\paragraph{Incomplete training labels.}
To test alignment with incomplete training labels, each element of every training label is deleted independently with probability \dropPctLow\% or \dropPctHigh\% (\dropActualLow\% and \dropActualHigh\% of the \calibTargetElements{} label elements are deleted), and one reader is aligned on each set with the seed and hyperparameters of the main aligned reader. Both readers read the held-out screenshots and are scored against the true text views.
With \dropPctLow\% of the label elements missing at random, the aligned reader's element error rate on the held-out screenshots is \errGateDropLow\% (\errGateAligned\% with complete labels, \errGateOrig\% before alignment). With \dropPctHigh\% missing, it is \errGateDropHigh\%. Compared with complete labels, almost all of the additional errors are missed elements (\cref{tab:supp_errtypes}). With the clean-text policy on the confirmation set, the reader aligned with \dropPctHigh\% of the label elements deleted (the \dropPctHigh\% reader) reaches \succDropHigh\% full-task success (\succDropHighSeedA\% and \succDropHighSeedB\% for the policy's two training seeds). It is clearly lower than the original reader (\succRealClean\%; $\dDropHighOrig$ points), and the main aligned reader, aligned on complete labels (\succAlignClean\%), is clearly higher than it ($\dFullDropHigh$ points). Both comparisons are descriptive; these runs made \readsDropHigh{} evaluation calls of the reader. 

\paragraph{Where the \dropPctHigh\% reader's misses fall.}
The \dropPctHigh\% reader misreads fewer elements of the held-out screenshots than the original reader (\errGateDropHigh\% against \errGateOrig\%), yet the policy completes fewer tasks with it (\succDropHigh\% against \succRealClean\%). We added a post-hoc count of where its misses fall after part of the confirmation runs had been seen and wrote down its procedure before the runs finished; it tests no cause. Task-relevant elements are the elements that the clean-text policy clicked or typed into in the successful text episodes of its first practice round (\cref{sec:setup}), matched by the matching rule of \emph{Recurrence of misreads} below; the definition involves no reader. They occur \mechTaskOccurrences{} times among the \gateElements{} elements of the held-out screenshots (\mechTaskElements{} distinct elements).
The original reader misses \mechOrigAllMissed\% of all elements but only \mechOrigTaskMissed\% of the task-relevant ones. The \dropPctHigh\% reader misses \mechHighAllMissed\% of all elements and \mechHighTaskMissed\% of the task-relevant ones; the reader aligned with \dropPctLow\% deleted misses \mechLowAllMissed\% and \mechLowTaskMissed\%, and the main aligned reader \mechFullAllMissed\% and \mechFullTaskMissed\%.
Of the \mechPersistSeen{} task-relevant elements that appear on held-out screenshots of at least two episodes, the \dropPctHigh\% reader misses \mechHighAlways{} every time, against \mechOrigAlways{} for the original reader and \mechFullAlways{} for the main aligned reader.
On the confirmation set, the policy got stuck, repeating one action on an unchanged screen, in \mechHighStuck{} of \confEpisodes{} episodes with the \dropPctHigh\% reader and in \mechOrigStuck{} with the original reader.
Only \mechHighIdentical{} of the \dropPctHigh\% reader's stuck episodes reach a screen, identified by its true text view, on which the policy clicked an element in a successful episode of the same instance with the main aligned reader; in all \mechHighNeededMissed, the \dropPctHigh\% reader had missed that element. The element it missed most often in its stuck episodes is the text button that adds the book to the shelf in \texttt{Add\allowbreak Book\allowbreak ToShelf} of WeChat Reading (\mechExampleStuck{} episodes). 

\paragraph{New screens.}
A screen is pixel-new if it is identical to no calibration screenshot; in the lookup baseline's run, \newScreenPixelPct\% of the \lookupScreens{} screens encountered on the confirmation set are pixel-new. A screen is new, as in \cref{tab:reader}, if its true text view appears in no training label; in the aligned reader's run, \newScreenListPct\% are new. Screens with the same text view can still differ in pixels, so fewer screens are new than pixel-new; the two numbers also come from different runs.
On the held-out screenshots, \gateIdenticalPct\% exactly repeat a calibration screenshot; on those the aligned reader errs on \errIdenticalAligned\% of elements. The lookup baseline's success falls as the share of pixel-new screens in an episode grows (\cref{tab:supp_newscreens}).
Because the aligned reader's run stored only JPEG screenshots, a rule fixed in advance allowed JPEG-based matching on that run only if it first reproduced every exact match recorded during the lookup run. JPEG-based matching disagreed on \lookupJpegMismatch{} of the \lookupScreens{} screens, all in one direction (an exact match judged pixel-new from the JPEG), so \cref{tab:supp_newscreens} gives no such breakdown for the aligned reader's run; \cref{supp:further} gives one under a later rule that accepts this one-sided error. 

\begin{table}[!tp]
  \centering
  \footnotesize
  \caption{\textbf{The lookup baseline succeeds less often when more of an episode's screens are pixel-new.} Episodes on the confirmation set binned by their share of pixel-new screens (identical to no calibration screenshot), and the lookup baseline's full-task success in each bin (descriptive; the template mix differs across bins).}
  \label{tab:supp_newscreens}
\begin{tabularx}{\linewidth}{@{}Xrr@{}}
    \toprule
    Share of pixel-new screens         & Episodes               & Success (\%)\,$\uparrow$ \\
    \midrule
    Below one third              & \lookupBinLowEpisodes  & \lookupBinLowSucc \\
    One third to two thirds      & \lookupBinMidEpisodes  & \lookupBinMidSucc \\
    Above two thirds             & \lookupBinHighEpisodes & \lookupBinHighSucc \\
    \bottomrule
  \end{tabularx}
\end{table}

\paragraph{Screen of \cref{fig:errors}.}
\cref{fig:errors} shows the first screen, in the order of the calibration set, that contains elements of all three classes (always, sometimes and never misread); the rule was fixed before the figure was drawn. 

\paragraph{Recurrence of misreads.}
\cref{tab:persist} gives, for all four reader settings, the share of misread elements that are misread again in another episode (\cref{sec:systematic}).
Under the matching rule of \cref{sec:systematic}, two reads count as the same element when they come from the same app and agree in role, text, icon name, state and position (each box coordinate rounded to a multiple of \matchGrid{} on the 0--\genericJitterScale{} scale); the status bar is excluded, and each element is counted once per episode, at its first read. The Qwen3-VL-8B row of \cref{tab:persist} was measured on full-size screenshots, as in one of the two configurations of the 8B reader's task runs (\cref{sec:scope}); the other uses half-size screenshots, like both 4B rows. Exact outputs vary between reads of the same screen. When the same screen is read twice, the two outputs are identical in only \rereadIdentical\% of cases, but most differences are box coordinates a few pixels apart (\rereadBoxOnly\%); the elements themselves differ in \rereadElemDiff\% of cases and only their order in \rereadOrderOnly\%. When the 4B reader misreads the same piece of text in two episodes, the two misreadings are identical in \sameWrongPct\% of cases, against \sameWrongPctRate\% under rate-matched noise, whose errors are drawn anew in each episode (\cref{supp:practice}). Elements misread at temperature \readerTemp{} are still misread at temperature \sampledTemp{} in \tempPersistPct\% of cases. Adding an element's own calibration history (its misreads in the calibration screenshots) to a predictor that uses the error rate of similar elements raises the AUC for predicting its next error from \aucBase{} to \aucHistory.

\paragraph{Lookup baseline.}
A fallback fixed in advance, a lookup-based correction for the case that the reader could not be fine-tuned in time, was not needed; a lookup of the calibration screenshots later served as a control (\cref{sec:reader_side}). The lookup matches a screen only when it is pixel-identical to a calibration screenshot, as fixed before the run; a lookup with approximate matching might recover more of the gain. Read as a chain of successive controls, alignment's \nosign{\dAlignClean}-point gain splits into \nosign{\dReorder} points from reordering, \nosign{\dLookupReorder} from recalling exactly repeated screens and \nosign{\dAlignLookup} from reading the other screens correctly (descriptive). 

\paragraph{Unaligned 8B reader.}
The larger reader is Qwen3-VL-8B-Instruct without alignment. Everything else matches the 4B runs, including the half-size screenshots, the prompt, the repetition penalty, the parser and the clean-text policy's two seeds. With it, the clean-text policy reaches \succLargeSeedA\% and \succLargeSeedB\% for its two training seeds, against \seedCleanA\% and \seedCleanB\% with the original 4B reader.
Its text-to-pixels gap is \nosign{\gapLarge} points, and the comparison with direct screenshot training points toward direct training ($\dLargeDirect$ points; both descriptive, \cref{tab:supp_intervals}).
For its element error rate it reads the \gateShots{} held-out screenshots with the settings of its task runs (\gateShots{} reader calls, counted under Audit in \cref{tab:supp_accounting}). It misses fewer elements than the original 4B reader and outputs fewer extra ones, while wrong text and wrong state are about as frequent (\cref{tab:supp_errtypes}).
Its task success shows no measurable difference from that with the original 4B reader (descriptive), perhaps because the elements it additionally reads correctly are not the ones the tasks need; we did not test this. \cref{supp:further} reports it after alignment. 

\paragraph{Unaligned 8B reader on full-size screenshots.}
In a second configuration the unaligned 8B reader reads full-size screenshots; everything else matches its half-size runs. With it, the clean-text policy reaches \succLargeFullSeedA\% and \succLargeFullSeedB\% for its two training seeds.
The comparison with the original 4B reader points toward this configuration ($\dLargeFullOrig$ points), and the comparison with direct screenshot training toward direct training ($\dLargeFullDirect$ points).
Against the 8B reader on half-size screenshots it shows no measurable difference ($\dLargeFullHalf$ points; all three comparisons descriptive, \cref{tab:supp_intervals}). On the \gateShots{} held-out screenshots, read with the same settings (\gateShots{} reader calls, counted under Audit in \cref{tab:supp_accounting}), the 8B reader's element error rate is \errGateLargeFull\% at full size, against \errGateLarge\% at half size; almost all of the rise is in missed elements (\typeMissLargeFull{} against \typeMissLarge; \cref{tab:supp_errtypes}). 

\paragraph{Cost against the 8B reader on full-size screenshots.}
In its task runs the 8B reader on full-size screenshots has a GPU of its own next to the policy's, whereas the aligned 4B reader shares one GPU with the policy, so we compare costs by timing each reader and the policy alone rather than from the GPU time of the runs. Each reader, alone on one GPU, read the held-out screenshots with \costReadInflight{} reads in parallel; a read took \costReadAlignedSec{} GPU-seconds with the aligned 4B reader on half-size screenshots and \costReadLargeFullSec{} with the 8B reader on full-size screenshots, a ratio of \costRatioReadLargeFull. The cost threshold of the reader choice (\readerPickCostLine) was applied only to the timings of the reader choice, which used the original 4B reader, \readerPickShots{} other screenshots and \readerPickInflight{} reads in parallel. Each step of an episode takes one read and one policy call; a policy call took \costPolicyCallSec{} GPU-seconds, timed the same way on the clean-text policy's runs in text.
On the confirmation set, an episode took \costStepsAligned{} steps on average with the aligned reader and \costStepsLargeFull{} with the 8B reader on full-size screenshots, so it cost \costRatioEpisodeLargeFull{} times as much with the aligned reader. 

\paragraph{Templates outside the calibration set.}
The \otherTmplCount{} templates of \cref{tab:reader} lie outside the \confTemplates{} templates used for training and for the main evaluation. We did not compare task success on these \otherTmplCount{} templates, because the clean-text policy, which never practiced them, completes \otherTmplTextSucc\% of them even in text, below the pre-specified threshold of \otherTmplThreshold\%. \cref{tab:supp_unpracticed} gives the text success per template.

\begin{table}[!tp]
  \centering
  \footnotesize
  \caption{\textbf{Even in text, the clean-text policy completes few tasks of the templates it never practiced.} Text success (\%) on the \otherTmplCount{} unpracticed templates (two training seeds). The real pipeline would have been run only if the average over all \otherTmplCount{} reached \otherTmplThreshold\%.}
  \label{tab:supp_unpracticed}
\begin{tabularx}{\linewidth}{@{}Xr@{}}
    \toprule
    Template                               & Text success (\%)\,$\uparrow$ \\
    \midrule
    \texttt{QueryHolidayLength}            & \otherTextHolidayLength \\
    \texttt{AlipayBalanceToWechat}         & \otherTextAlipayBalance \\
    \texttt{ReadNoteContent}               & \otherTextReadNote \\
    \texttt{CompareParticipationDurations} & \otherTextMeetingDurations \\
    \texttt{MonthlyIncomeByCounterparty}   & \otherTextMonthlyIncome \\
    \texttt{SetAlarmRepeat}                & \otherTextAlarmRepeat \\
    \texttt{StartFastMeeting}              & \otherTextFastMeeting \\
    \texttt{ToggleFriendConfirmation}      & \otherTextFriendConfirm \\
    \midrule
    All                                    & \otherTmplTextSucc \\
    \bottomrule
  \end{tabularx}
\end{table}

\paragraph{Held-out apps.}
Without WeChat, \heldShotsNoWeChat{} calibration screenshots remain, and without Clock, \heldShotsNoClock; each held-out-app reader is trained with the settings of full alignment. The pre-specified test on element errors uses the \heldGateShots{} held-out screenshots of single-app episodes of the held-out apps; the relative reduction in element errors, \heldErrReductionPct\%, passes the pre-specified threshold of \heldErrLinePct\%.
On these held-out screenshots, full alignment reaches an element error rate of \heldErrFull\%. On the \heldGateCross{} held-out screenshots of cross-app episodes, reported separately, the element error rate is \heldCrossErrOrig\% with the original reader, \heldCrossErrHeld\% with the held-out-app reader and \heldCrossErrFull\% with full alignment.
On the \heldTemplates{} templates of the pre-specified test on task success, the held-out-app reader's task success shows no measurable difference from the original reader's (\heldSuccHeld\% vs.\ \heldSuccOrig\%; \cref{tab:supp_intervals}); its remaining errors may fall on the elements the tasks need, which we did not test. With so few templates the interval is wide, and the original reader already reaches \heldSuccOrig\% on them, against \succRealClean\% on the whole confirmation set.
Full alignment is \nosign{\dFullHeld} points above the held-out-app reader on these templates (descriptive; \cref{tab:supp_intervals}).
Episodes that cross apps are left out of this test. On each of the \heldCrossTemplates{} cross-app templates, success is \heldCrossSuccOthers\% for the clean-text policy with the held-out-app reader or the original reader and for direct screenshot training, while full alignment reaches \heldCrossSuccFullA\% on \texttt{Alipay\allowbreak Monthly\allowbreak Spend\allowbreak ToWechat}, \heldCrossSuccFullB\% on \texttt{Calendar\allowbreak Event\allowbreak ToWechat} and \heldCrossSuccFullC\% on \texttt{Calendar\allowbreak Earliest\allowbreak ToAlarm}. 

\section{Text view}
\label{supp:textview}
\begin{figure*}[t]
  \centering
  \includegraphics[width=\linewidth]{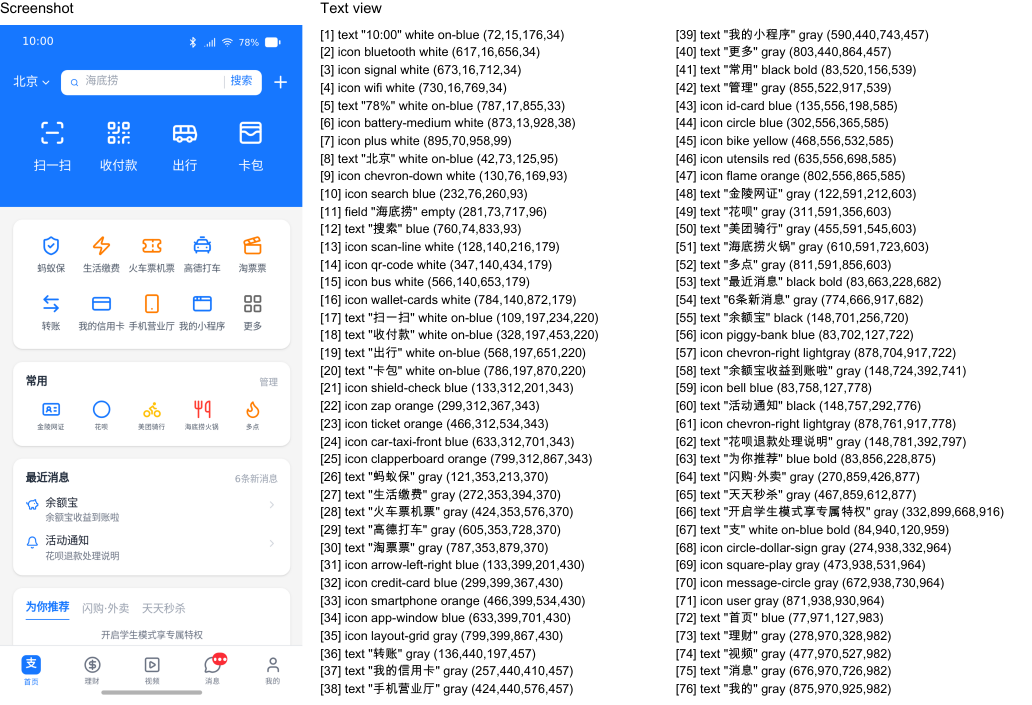}
  \caption{\textbf{The text view lists each visible element on one line, in the layout order.} The screen of \cref{fig:errors} as a screenshot (left) and its text view (right), as the policy reads it in clean-text practice. Each line gives the index, role, text or icon name, the attributes that apply (color, background color and flags such as \texttt{bold} or \texttt{empty}) and the box on the 0--\genericJitterScale{} scale. Interface text is Chinese, as on the screen.}
  \label{fig:supp_textview}
\end{figure*}

\paragraph{Extraction rules.}
The text view contains only the elements visible on the screen, in the layout order of \cref{sec:setup}, and has no MobileGym task markup (\texttt{data-trigger*} attributes), ARIA accessibility attributes, CSS classes, routes (URL paths), hidden state or list of legal actions. An extractor reads the text view from the page that the app renders in the browser. It writes each element as one line with an index, its role, an icon name, its text in quotes, a switch state (on or off), a mark state (filled or empty), its color, a background color, the flags bold and dim, the flag empty for an input field that shows only a placeholder, and its box \texttt{(x1,y1,x2,y2)} normalized to 0--\genericJitterScale{} by the screen's width and height; attributes that do not apply are left out. The roles are \texttt{text}, \texttt{icon} (a small graphic without text), \texttt{mark} (a check mark: a checkbox, radio circle or small dot), \texttt{field} (an input field), \texttt{switch} and \texttt{image} (a photo or large picture). The color is one of twelve names (black, gray, lightgray, white, red, orange, yellow, green, cyan, blue, purple, pink), and a background color is given only when it is neither white nor black. Icon names come from two icon-name tables, one built from the shapes of the Lucide icon library and one with names assigned by hand to the remaining icons; both are shared by all groups. The extractor also records whether a mark is round and whether an icon is a bitmap, which the policy never sees.
Elements are ordered by position on the screen, not by page structure. The screen is cut into fixed horizontal bands of \textviewRowPx{} CSS pixels, a row holds the elements whose top edges fall in the same band, and each row runs from left to right. Adjacent text pieces in a row with the same color and background are merged into one element. \cref{fig:supp_textview} shows the screenshot and the text view of the screen in \cref{fig:errors}. At each step the policy receives the task, the step number and the step limit, its recent actions and notes (\cref{supp:practice}) and this list; in the real pipeline the list is the parsed output of the reader, in the same format.

\paragraph{Visible elements.}
The extractor walks the rendered page from its root and turns a node into an element only if it is one of five kinds and visible on the screen. The kinds are a non-empty text node, an icon or picture, an input field, a check mark and a switch. Text drawn in an icon font counts as an icon. An icon or picture is an SVG with a colored shape, a loaded image, video, canvas or background picture, a shape cut out by a CSS mask, or a cluster of at least \visDotsMin{} small dots. A check mark is a native checkbox or radio button, or a small near-square box with a colored border or a fill that differs from what lies behind it, and a switch is a rounded pill with one round knob. Size limits in screen pixels of the \viewportW{}\,$\times$\,\viewportH{} viewport, fixed in the extractor, separate icons from pictures, and check marks and switches from other boxes.
A node is visible if it is not hidden, its accumulated opacity is at least \visOpacityMin, its part inside the screen and inside every enclosing element that clips its content is at least \visMinSide{} pixels wide and tall, and no opaque layer covers it. Text is kept character by character, so a partly covered or truncated label keeps only its visible characters, and text whose luminance contrast with the solid color behind it is below \visContrastMin{} is dropped.
Other shapes drawn only with a background or a border, such as panels, dividers and plain badges, are not elements; their color appears only as the background color of text drawn on them. For example, the red unread badge on the Messages tab in \cref{fig:supp_textview} is a rounded box with three dots, too few for a dot cluster, and no text, so the text view has no line for it.
Text generated by CSS is not read; in the MobileGym source it occurs only as empty strings. Iframes are not entered; in MobileGym only the system browser app uses them, and none of our templates belongs to it.

\paragraph{Leakage checks.}
The automated test suite of the extractor grew from \leakTestsOne{} tests to \leakTestsTwo{} after a fix to the extractor and to \leakTestsThree{} with regression tests for bitmap icons; every version of the suite passed in full, and the last one ran on the extractor and icon-name tables used in all experiments. The tests scan the extractor's source for reads of \texttt{aria-*}, \texttt{data-*}, \texttt{class}, \texttt{id}, \texttt{role}, \texttt{title}, \texttt{alt}, \texttt{name}, \texttt{href} or \texttt{src} attributes, routes, app state, scroll information or legal actions. They check that hidden labels, interactivity and tag names planted in a test page are absent from the output while all visible content is present, and they check display details (ellipses, masked passwords, colors of shapes, states of marks and switches, rotated icons, omission of input boxes that show nothing). Other tests confirm that a bitmap icon missing from the icon-name tables is written as \texttt{icon} without a name, that its image address, used only to look up the name, never enters the output, and that extraction changes neither the page structure nor the click targets nor the style sheets.
A pixel audit hides each element in turn on screens reached by random walks and checks that the screenshot changes. On \pixAuditScreens{} screens with \pixAuditElements{} elements, hiding \pixAuditUnchanged{} elements (\pixAuditUnchangedPct\%), all in a weather app that the study does not use, left the screenshot unchanged, and hiding \pixAuditFaint{} (\pixAuditFaintPct\%), all low-contrast text, changed it only faintly. After the fix of bitmap icon names, a recheck of \pixAuditRecheckScreens{} screens found that hiding \pixAuditRecheck{} elements (\pixAuditRecheckPct\%) left the screenshot unchanged and hiding \pixAuditRecheckFaint{} (\pixAuditRecheckFaintPct\%) changed it only faintly. Finally, screens were inspected against their lists, in \manualCheckRounds{} rounds of \manualCheckScreens{} during development and \manualAuthorScreens{} in an independent check by one of the authors: one screen of each app, one random screen and five hard ones (an answer-entry page with the keyboard open, a calendar month view, a long list that needs scrolling, a toolbar of icons only and a pop-up). For each screen the check asked whether the list contains anything that is not visible, misses anything visible (counting elements as in \emph{Visible elements} above) or has wrong text, and the answer was no for all \manualAuthorScreens. 

\section{Practice environments and policy training}
\label{supp:practice}

\paragraph{Generic noise.}
At the start of each episode, generic-noise practice draws five rates uniformly: an omission rate of up to \genericOmitMax\%, a single-character error rate of up to \genericCharErrMax\%, box jitter of up to \genericJitterMax{} units on a coordinate scale of \genericJitterScale, a flip rate of up to \genericFlipMax\% for switch and mark states, and a list-truncation probability of up to \genericTruncMax\%. It applies these rates at every step of that episode. The ranges were fixed in advance and are not fitted to the reader. 

\paragraph{Rate-matched noise.}
The error model is fitted on the \calibShots{} calibration screenshots (\calibEpisodes{} episodes), each with its true text view and the original reader's output; combined training uses only the first \directRoundOneShots. Each true element is matched to what the reader wrote for it. Elements are sorted into categories by role, text length, whether the text contains digits, box size, whether the background is colored, the region of the screen and the number of elements on the screen, and a category with fewer than \rateMinObs{} observations (\rateMinObsState{} for states) falls back to a coarser category. During practice, each true element draws one outcome from what the reader made of the elements of its category. Any element may be missed, or its box shifted by an observed offset, so that a click goes to the center of the shifted box and the simulated app decides what it hits. Text may be misread by an observed substitution, deletion or insertion, or merged with its right neighbor, and text on a colored button may become a filled mark of the button's color, losing the text. An icon may be renamed as observed, possibly to no name, or read as another role, and the state of a switch or mark may be flipped as observed for its role, true state and color.
Extra elements are added, as many as the reader added for text of the same category. They are derived from the screen's own text (a repeat, a part, a piece joined with its neighbor or a near spelling) or drawn from the extra elements that the original reader wrote on screens of the same app. Lists are truncated as the original reader's lists were on screens with as many elements, and the result is sorted into the layout order, like every text view in practice. An unchanged screen gets the same errors throughout an episode, because the random draw is seeded by the episode and the screen content. Before training, the episodes were split \rateFitSplit{} for a fit check; on the episodes left out of the fit, the rates of missed elements, of colored buttons read as marks and of flipped states were each within \rateFitPoints{} points of the original reader's. 

\paragraph{Adapting the policy with the same labels.}
For each training seed, the clean-text policy is the teacher. At every step of the calibration episodes it sees the true text view, in the layout order and format of practice, and gives its answer. A student sample pairs the teacher's answer with the original reader's list that the start policy received at that step, keeping the same history and notes; the element the teacher clicks is mapped to its index in that list, and the sample is dropped when the original reader missed the element.
This keeps \sameLabelKeptA{} and \sameLabelKeptB{} samples for the two seeds and drops \sameLabelDroppedA{} and \sameLabelDroppedB, or \sameLabelDroppedPctA\% and \sameLabelDroppedPctB\% of element actions.
The policy is retrained with LoRA from the base model on the start policy's demonstrations, the seed's practice samples from practice rounds 1 and 2, and the student samples (\sameLabelRowsA{} and \sameLabelRowsB{} samples in all), with the hyperparameters shared by all groups, and it is evaluated with the original reader.
The comparison with direct screenshot training points toward the adapted policy ($\dSameLabelDirect$ points; descriptive, \cref{tab:supp_intervals}).

\paragraph{Untrained base policy.}
The untrained base policy is Qwen3-8B without LoRA, with the prompt of the other groups. In text it completes \baseTextSucc\% of the \confInstances{} confirmation instances, above the pre-specified threshold of \baseTextThreshold\% for running it in the real pipeline; \baseOverlongEpisodes{} of the \confInstances{} text episodes count as failures because the prompt exceeded the context length, which happened in none of its real-pipeline episodes. In the real pipeline it completes \succRealBase\% of the confirmation instances with the original reader and \succAlignBase\% with the aligned reader; the two readers show no measurable difference ($\dAlignBase$ points, aligned minus original; descriptive, \cref{tab:supp_intervals}). Its success with the original reader is thus no lower than in text, so it has no text-to-pixels gap with that reader (\cref{sec:reuse}). The missing gap is not explained by which templates this policy completes: on the \baseTextTemplates{} templates that it completes in text at least once, the clean-text policy still loses \baseSubsetCleanGap{} points from text to the original reader (\baseOtherCleanGap{} on the others; \baseWeightedCleanGap{} when templates are weighted by the base policy's successes in text).
Its outcomes on the same instance move in both directions between text and the original reader (\baseLost{} lost, \baseGained{} gained), whereas the clean-text policy's move almost only downward (\cleanLostA{} and \cleanLostB{} lost, \cleanGainedA{} and \cleanGainedB{} gained, per training seed).
Each instance was run once with sampling, so we cannot tell how much of this movement is sampling noise.
We added this analysis after seeing the result and did not test why the gap is missing. 

\paragraph{Policy.}
At each step the policy sees the task, its last \policyHistActions{} actions, its last \policyHistNotes{} notes (any action may carry a note of up to \policyNoteChars{} characters) and the current element list, within a \policyContextTokens-token context. Qwen3's thinking mode, which writes free-form reasoning before the action, is enabled, up to \policyThinkTokens{} tokens per step, but no training data contain thinking: the scripted demonstrations have none, and the practice episodes come from policies trained on them. The thinking that the trained policies output is therefore empty. An episode ends when the policy declares the task done, when the same action is repeated \policyRepeatStop{} times in a row on an unchanged screen, or at MobileGym's step limit for its difficulty level: \stepLimitLevelOne, \stepLimitLevelTwo, \stepLimitLevelThree{} or \stepLimitLevelFour{} steps, plus \stepLimitAnswerExtra{} when the task asks for an answer (\stepLimitConfMin{} to \stepLimitConfMax{} steps on the confirmation set). 

Every practice round trains a new LoRA from the base model instead of continuing the previous one. The LoRA has rank \policyLoraRank, alpha \policyLoraAlpha{} and dropout \policyLoraDropout{} on all \policyLoraTargets{} attention and MLP projections. It is trained with AdamW at learning rate $\policyLR$ without weight decay, with a cosine schedule, \policyWarmupPct\% warm-up and gradient clipping at \policyGradClip, for \policyEpochs{} epochs with a batch size of one and gradients accumulated over \policyGradAccum{} samples, in bf16 with gradient checkpointing. Samples longer than \policyMaxSampleTokens{} tokens are dropped (\policyDroppedTooLong{} in all policy training), and the loss covers only what the policy wrote (thinking and action). Apart from the student samples of the adapted policy above, training samples come only from episodes that the judge accepted and in which the policy itself declared the task done. Each step of such an episode is one sample (the system prompt and the user message at that step, and the policy's answer), except steps whose action could not be parsed, was invalid or hit the length limit.

The scripted demonstrations, shared by all groups, are \demoEpisodes{} episodes recorded in the text view without any reader call, one on each of \demoPerTemplate{} training instances of each of the \confTemplates{} templates. Of these, \demoPassed{} succeeded and give \demoSamples{} samples covering \demoTemplates{} templates; \texttt{Send\allowbreak Message\allowbreak ToContact} has no successful demonstration. In each practice round the current policy (in the first, the start policy) runs \practiceSamplesPerTemplate{} sampled episodes on one new training instance of each template (\practiceEpisodes{} episodes) in its group's practice environment, and a new LoRA is trained on the judge-accepted episodes, the demonstrations and the episodes kept in earlier practice rounds. Direct screenshot training instead learns in its first practice round from the \directRoundOnePassed{} judge-accepted episodes of the start policy among the first \directRoundOneShots{} calibration screenshots (\directRoundOneSamples{} samples), and in its second practice round runs \directRoundTwoSamplesPerTemplate{} sampled episodes per template in the real pipeline with the original reader. Taking the first sampled episode of every template, then the second, and so on, it keeps the whole episodes that contain its first \directRoundTwoShots{} reader calls and trains on the judge-accepted ones among them, together with the earlier data.
The direct-training policy with the aligned reader (\cref{sec:scope}) thus rests on about \directAlignedShotsApprox{} screenshots, about \directAlignedDistinctApprox{} of them distinct. In its first practice round, combined training practices in a rate-matched-noise environment fitted on these \directRoundOneShots{} calibration screenshots and also trains on the same \directRoundOneSamples{} samples; its second practice round is that of direct screenshot training. During practice the policy samples with temperature \policyTemp, top-p \policyTopP{} and top-k \policyTopK, with one seed per episode. A training seed sets the order of the data, the initialization of the LoRA and the offset of the episode seeds.

\section{Cost and screen-reader accounting}
\label{supp:accounting}

An episode in text costs \costTextGpuSec{} GPU-seconds and an episode on screenshots \costPixelGpuSec{}, about \costRatioEpisode$\times$ more (\costRatioStep$\times$ per step). The cost is GPU time. The policy and the reader share one A6000 and \parallelWorkers{} episodes run together, and we divide a run's total GPU time by its number of episodes, so the cost of an episode on screenshots already includes its reader calls, about \costReaderCalls{} per episode. A reader call takes \readerSecAligned{}\,s of wall-clock time after alignment and \readerSecOrig{}\,s before; these are waiting times under batching and do not add to this cost. The aligned reader's outputs are \readerLongerPct\% longer, and on the confirmation set they reach the reader's limit of \readerMaxTokens{} tokens on \truncAligned{} of \confScreensAligned{} screens, against \truncOrig{} of \confScreensOrig{} for the original reader. Every screen-reader call is logged by purpose (totals in \cref{tab:supp_accounting}).
Development calls were made before the study; no call was spent on selecting checkpoints, because every policy and every aligned reader is used as it stands at the end of its training (\cref{sec:method}). 

\begin{table}[!tp]
  \centering
  \footnotesize
  \caption{\textbf{Most screen-reader calls are spent on evaluation.} Calls by purpose. Collection: calibration and held-out screenshots. Training: real-pipeline practice rounds of direct training (second to fourth) and combined training (second). Development: building the pipeline and choosing the reader, before the study. Audit: diagnostic re-reads, held-out reads by the aligned readers and the unaligned 8B reader, and cost measurements.}
  \label{tab:supp_accounting}
\begin{tabularx}{\linewidth}{@{}Xr@{}}
    \toprule
    Purpose     & Calls \\
    \midrule
    Collection  & \callsCollection \\
    Training    & \callsTraining \\
    Development & \callsDevelopment \\
    Evaluation  & \callsEvaluation \\
    Audit       & \callsAudit \\
    \midrule
    Total       & \callsTotal \\
    \bottomrule
  \end{tabularx}
\end{table}

\section{Limitations and discussion}
\label{supp:limitations}

\paragraph{Same tasks, new instances.}
Task success is measured on the templates that the policies practiced, with new instances. The clean-text policy, which learned only the practiced templates, completes \otherTmplTextSucc\% of the tasks of the \otherTmplCount{} unpracticed templates even in text, so task success could not be compared there; alignment lowers the element error rate on them as elsewhere (\cref{sec:reader_side}). 

\paragraph{Labels.}
Alignment needs the text view of each calibration screenshot.
A simulated platform provides it for free; on real devices it must come from accessibility trees, which can be incomplete, or from annotation. With \dropPctHigh\% of label elements deleted at random, the aligned reader falls below the original reader in task success (\cref{supp:reader}). 

\paragraph{Simulated platform and repeated screens.}
All task runs use one platform, MobileGym, whose simulated apps have Chinese interfaces and are rendered in a browser; the reader and the policy both come from the Qwen family, and we have not run tasks on real devices. MobileGym renders the same screen identically across instances, so \lookupHitPct\% of the screens encountered on the confirmation set exactly repeat a calibration screenshot, and a lookup of those screenshots recovers \lookupGainPct\% of the gain. Real devices repeat screens less exactly, so there the gain would depend more on reading new screens correctly; on the new screens of MobileGym, alignment cuts the element error rate from \errGateNewOrig\% to \errGateNewAligned\%. \cref{supp:further} tests how far the gain in task success depends on exact repeats.
On real-app screenshots from AndroidControl (\cref{supp:second_platform}), misreads also recur, but alignment lowers the element error rate of all held-out screenshots by \awErrDropPct\% instead of \errGateReductionPct\%. 

\paragraph{When to align the reader.}
Aligning the reader was cheap and effective on MobileGym and helped far less on AndroidControl (\cref{supp:second_platform}).
The alternative, noise injected into practice, would have to reproduce where the reader's errors fall as well as how often they occur; the task-level predictivity test can check such an environment before policies are trained in it.

\section{A second platform: AndroidControl}
\label{supp:second_platform}
\begin{table}[!tp]
  \centering
  \footnotesize
  \setlength{\tabcolsep}{4pt}
  \caption{\textbf{On AndroidControl, misreads also recur, but alignment lowers the element error rate far less than on MobileGym.} MobileGym values are from \cref{tab:persist,tab:reader}. Recurrence: share (\%) of misread elements that are misread again in another episode, original reader at temperature \readerTemp{} on the alignment set; interval: app groups resampled. Error rates (\%) on held-out screenshots; new screens: text view in no alignment screenshot. Both criteria were fixed before the runs.}
  \label{tab:supp_androidcontrol}
\begin{tabularx}{\linewidth}{@{}Xrr@{}}
    \toprule
                                               & MobileGym                             & AndroidControl \\
    \midrule
    Alignment set: screenshots (apps)          & \calibShots{} (\confApps)             & \awAlignShots{} (\awAlignApps) \\
    Held-out screenshots                       & \gateShots                            & \awHeldShots \\
    \addlinespace[2pt]
    \tabblock{3}{Recurrence of misreads (\%)} \\
    Original reader                            & \persistSmall                         & \awRecurReader \\
    Independent errors                         & \persistSmallIndep                    & \awRecurIndep \\
    Difference (points)                        & \vup{\persistSmallDiff}               & \vup{\awRecurDiff} \\
    Interval                                   & \ci{\persistSmallLo}{\persistSmallHi} & \ci{\awRecurLo}{\awRecurHi} \\
    Criterion: $\geq$\,\awLineRecurPoints{} points, lower bound $>0$ & \cmark   & \cmark \\
    \addlinespace[2pt]
    \tabblock{3}{Element error rate (\%)} \\
    Original reader                            & \errGateOrig                          & \awErrOrig \\
    Aligned reader                             & \errGateAligned                       & \awErrAligned \\
    Relative drop                              & \errGateReductionPct                  & \awErrDropPct \\
    New screens, original reader               & \errGateNewOrig                       & \awErrNewOrig \\
    New screens, aligned reader                & \errGateNewAligned                    & \awErrNewAligned \\
    Unparsable screenshots, aligned reader     & \gateUnparsableAligned                & \awUnparsableAligned \\
    Criterion: drop $\geq$\,\awLineDropPct\%, unparsable $\leq$\,\awLineUnparsablePct\% & \cmark & \xmark \\
    \bottomrule
  \end{tabularx}
\end{table}

\makeatletter\setlength{\@dblfptop}{0pt}\makeatother \begin{figure*}[!t]
  \centering
  \begin{tikzpicture}
\pgfplotsset{acbars/.style={paperplot, ybar=1.5pt, bar width=0.44cm, width=4.3cm, height=3.0cm, ymin=0,
  xmin=-0.55, xmax=1.55, xtick={0,1}, xticklabel style={align=center}, axis x line*=bottom,
  nodes near coords, point meta=explicit symbolic,
  every node near coord/.append style={font=\sffamily\fontsize{6.5}{7.5}\selectfont, inner sep=1.2pt},
  title style={font=\sffamily\bfseries\fontsize{7}{8.2}\selectfont, yshift=-2pt, align=center}}}
\begin{axis}[acbars, name=pa, ymax=100, ytick={0,25,50,75,100}, ylabel={Misread again (\%)},
  xticklabels={MobileGym, AndroidControl}, title={(a) Misreads recur}]
  \addplot[fill=black!42, draw=none] coordinates {(0,\persistSmall) [\persistSmall] (1,\awRecurReader) [\awRecurReader]};
  \addplot[fill=white, draw=black!50, line width=0.5pt] coordinates {(0,\persistSmallIndep) [\persistSmallIndep] (1,\awRecurIndep) [\awRecurIndep]};
\end{axis}
\begin{axis}[acbars, name=pb, at={($(pa.east)+(1.25cm,0)$)}, anchor=west, ymax=40, ytick={0,10,20,30,40},
  ylabel={Element error rate (\%)}, xticklabels={MobileGym, AndroidControl}, title={(b) Alignment, held-out screenshots}]
  \addplot[fill=black!42, draw=none] coordinates {(0,\errGateOrig) [\errGateOrig] (1,\awErrOrig) [\awErrOrig]};
  \addplot[fill=ourblue, draw=none] coordinates {(0,\errGateAligned) [\errGateAligned] (1,\awErrAligned) [\awErrAligned]};
  \draw[basered, densely dashed, line width=0.7pt]
    ([xshift=0.03cm]axis cs:0,{(100-\gateMinReductionPct)/100*\errGateOrig}) -- ([xshift=0.47cm]axis cs:0,{(100-\gateMinReductionPct)/100*\errGateOrig})
    ([xshift=0.03cm]axis cs:1,{(100-\awLineDropPct)/100*\awErrOrig}) -- ([xshift=0.47cm]axis cs:1,{(100-\awLineDropPct)/100*\awErrOrig});
\end{axis}
\begin{axis}[acbars, name=pc, at={($(pb.east)+(1.25cm,0)$)}, anchor=west, ymax=50, ytick={0,10,20,30,40,50},
  ylabel={Element error rate (\%)},
  xticklabels={{No alignment\\screenshots\\(\awUnseenApps{} apps)}, {Most alignment\\screenshots\\(\awManyApps{} apps)}},
  title={(c) AndroidControl apps (descriptive)}]
  \addplot[fill=black!42, draw=none] coordinates {(0,\awUnseenOrig) [\awUnseenOrig] (1,\awManyOrig) [\awManyOrig]};
  \addplot[fill=ourblue, draw=none] coordinates {(0,\awUnseenAligned) [\awUnseenAligned] (1,\awManyAligned) [\awManyAligned]};
\end{axis}
\node[anchor=north, font=\sffamily\fontsize{7}{8.2}\selectfont, inner sep=0pt] at ($(pb.south)+(0,-1.25cm)$) {\tikz[baseline=-0.6ex]{\fill[black!42] (0,-0.09) rectangle (0.26,0.09);}~original reader\quad
  \tikz[baseline=-0.6ex]{\draw[black!50, fill=white, line width=0.5pt] (0,-0.09) rectangle (0.26,0.09);}~independent errors at its rate\quad
  \tikz[baseline=-0.6ex]{\fill[ourblue] (0,-0.09) rectangle (0.26,0.09);}~aligned reader\quad
  \tikz[baseline=-0.6ex]{\draw[basered, densely dashed, line width=0.7pt] (0,0) -- (0.36,0);}~pre-specified threshold (drop of \awLineDropPct\%)};
\end{tikzpicture}   \caption{\textbf{Misreads recur on both platforms, but on AndroidControl alignment lowers the element error rate far less, and not on apps without alignment screenshots.} (a)~Share of misread elements that are misread again in another episode, for the original reader and for independent errors at the same rate. (b)~Element error rate on held-out screenshots before and after alignment; dashed: the pre-specified threshold. (c)~AndroidControl held-out apps without alignment screenshots and the apps with the most alignment screenshots; this breakdown was made after the results. MobileGym values are from \cref{tab:persist,tab:reader}.}
  \label{fig:supp_androidcontrol}
\end{figure*}
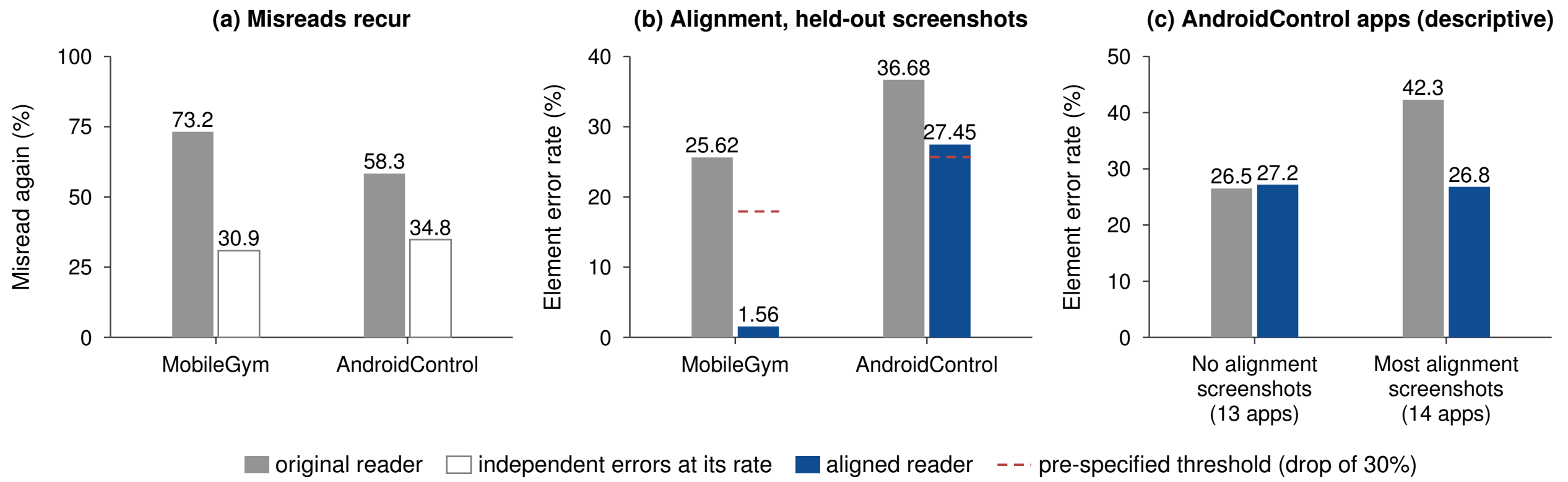

\providecommand{\androidcontrolsource}{ (W.~Li et al., \emph{On the Effects of Data Scale on UI Control Agents}, NeurIPS 2024 Datasets and Benchmarks Track)}

\paragraph{Setting.}
AndroidControl\androidcontrolsource{} records human demonstrations on real Android apps, with a screenshot and an accessibility tree at every step.
Because the dataset is offline, we repeat only the two reader-side measurements of the main paper, the recurrence of misreads (\cref{tab:persist}) and the effect of alignment on held-out screenshots (\cref{tab:reader}).
The text view, which is also the training label, comes from the accessibility tree, with each element checked against the screenshot so that only visible content remains; screenshots whose tree misses much of the screen are dropped.
Unlike MobileGym's text view, it gives no colors, does not name icons, and leaves out keyboards and web views.
From \awShardsUsed{} shards of the official training split, we drew, by a rule fixed before sampling, \awAlignShots{} alignment screenshots from \awAlignApps{} apps and \awHeldShots{} held-out screenshots from \awHeldApps{} apps, split by episode; \awAppsBoth{} apps appear in both sets.
The reader, its prompt, decoding, parser, matching and alignment recipe are those of the main paper; the prompt stays in Chinese, because an English translation failed a parsing check before the runs.
Alignment took \awTrainMinutes{} minutes on one GPU.
The \awCallsTotal{} reader calls (\awCallsSmoke{} for checks before the runs, \awCallsOriginal{} by the original reader and \awCallsAligned{} by the aligned reader) are counted apart from \cref{tab:supp_accounting}.

\paragraph{Results.}
Of the elements that the original reader misreads on the alignment screenshots, \awRecurReader\% are misread again in another episode, \nosign{\awRecurDiff} points more than under independent errors (\nosign{\persistSmallDiff} on MobileGym), so the recurrence criterion is met (\cref{fig:supp_androidcontrol}a).
Alignment lowers the element error rate of the held-out screenshots from \awErrOrig\% to \awErrAligned\%, a relative drop of \awErrDropPct\%, short of the \awLineDropPct\% threshold (\cref{fig:supp_androidcontrol}b); no aligned read is unparsable.
The drop is the same on the \awHeldNewShots{} new screens.
On MobileGym, alignment met the same criterion with a drop of \errGateReductionPct\%.

\paragraph{Where alignment helps.}
The analyses below were made after the results and are descriptive.
On the \awUnseenApps{} held-out apps without alignment screenshots, the element error rate does not fall (\awUnseenOrig\% before and \awUnseenAligned\% after alignment); on the \awManyApps{} held-out apps with the most alignment screenshots, it falls from \awManyOrig\% to \awManyAligned\% (\cref{fig:supp_androidcontrol}c).
Held-out elements that also occur in the alignment set drop by \awSeenElDropPct\%, the others by \awUnseenElDropPct\%.
The reader also fits these labels poorly, with a mean training loss of \awTrainLoss{} in the second epoch against \awPaperTrainLoss{} on MobileGym.
On AndroidControl, alignment thus helps most on apps and elements it has seen.
On MobileGym, a reader aligned without an app still misread far fewer of that app's elements, but task success there showed no measurable difference (\cref{sec:scope}).
On both platforms, the reader should therefore be aligned on screenshots of the apps the agent will use.

\paragraph{Labels and remaining misses.}
Part of the error rate comes from the labels.
For \awBoxShareOrigPct\% of the texts that the original reader misses, it reads the text correctly, but its box differs in width from the label's box, which often spans the whole row, so the element counts as missed.
Counting such reads as correct lowers the two readers' error rates by \awBoxPointsOrig{} and \awBoxPointsAligned{} points, leaves the relative drop at \awBoxDropPct\%, and keeps recurrence just above its threshold ($\awBoxRecurDiff$ points).
Of the \awSpotElements{} elements that the aligned reader still misses in \awSpotShots{} spot-checked held-out screenshots, \awSpotMissedPct\% are on the screen but not output at all, mostly small icons; the rest are label problems or the same element written in another form.

\section{Further tests of reader alignment}
\label{supp:further}
This section tests whether the main comparison depends on screens seen during alignment, on the alignment seed or on the size of the reader, and breaks down the failures that remain.
Alignment stays clearly higher than direct screenshot training in episodes whose screens are mostly pixel-new, with each of three alignment seeds and with a reader twice the size.
Aligned without the \slDedupRemovedPct\% of calibration screenshots that also appear in earlier runs on the confirmation set, and thus on about a sixth of them, a reader shows no measurable difference from direct training.
With the aligned reader, half of the remaining failures begin in the reader's output.
Except where stated, the procedures were fixed before the analyses were run, and new readers are aligned with the unchanged recipe of \cref{sec:method}; the comparisons marked $^\ast$ in \cref{tab:supp_more_readers} are pre-specified, and all others are descriptive.

\begin{table}[!tp]
  \centering
  \footnotesize
  \setlength{\tabcolsep}{3pt}
  \caption{\textbf{In episodes whose screens are mostly pixel-new, the aligned reader is still clearly higher than direct screenshot training.} Episodes of the aligned reader's run on the confirmation set, binned as in \cref{tab:supp_newscreens} by their share of pixel-new screens (identical to no calibration screenshot); full-task success (\%) of four runs on the same instances and training seeds. $\Delta$: aligned minus direct training (points), colored as in \cref{tab:supp_wording}, with both bootstraps of \cref{supp:stats} run within each bin; descriptive.}
  \label{tab:supp_bins}
\begin{tabularx}{\linewidth}{@{}Lrrr>{\color{ourblue}}rrr@{}}
    \toprule
                               &                    & \multicolumn{3}{c}{Clean-text policy}           & Direct           & \\
    \cmidrule(lr){3-5}
    Pixel-new share            & Episodes           & Original            & Lookup           & \textbf{Aligned} & training         & $\Delta$ \\
    \midrule
    Below 1/3                  & \slPixLowEpisodes  & \slPixLowCleanOrig  & \slPixLowLookup  & \slPixLowAlign   & \slPixLowDirect  & \vgray{\slPixLowGap} \\
    1/3 to 2/3                 & \slPixMidEpisodes  & \slPixMidCleanOrig  & \slPixMidLookup  & \slPixMidAlign   & \slPixMidDirect  & \vup{\slPixMidGap} \\
    Above 2/3                  & \slPixHighEpisodes & \slPixHighCleanOrig & \slPixHighLookup & \slPixHighAlign  & \slPixHighDirect & \vup{\slPixHighGap} \\
    \bottomrule
  \end{tabularx}
\end{table}

\begin{table*}[!tp]
  \centering
  \footnotesize
  \setlength{\tabcolsep}{5pt}
  \caption{\textbf{Every reader aligned on all calibration screenshots is clearly higher than direct screenshot training; a reader aligned on the sixth of them that are not exact repeats shows no measurable difference from it.} Screenshots used for alignment (one over-long sample skipped per set; \cref{supp:reader}), element error rate on the \gateShots{} held-out screenshots, full-task success on the confirmation set (mean\,$\pm$\,std over the policy's two training seeds) and differences in points, colored as in \cref{tab:supp_wording}; intervals only for the pre-specified comparisons ($^\ast$).}
  \label{tab:supp_more_readers}
\begin{tabularx}{\textwidth}{Xrrlrrrr}\toprule
                                   & Alignment            & Held-out error           & Success                                         & \multicolumn{2}{c}{$X -$ direct training}                                & \multicolumn{2}{c}{$X -$ alignment seed 1} \\
    \cmidrule(lr){5-6}\cmidrule(l){7-8}
    $X$                            & screenshots          & rate (\%)\,$\downarrow$  & (\%)\,$\uparrow$                                & $\Delta$                       & Interval                                & $\Delta$                     & Interval \\
    \midrule
    Clean-text policy, original 4B reader & --                  & \errGateOrig             & \msd{\succRealClean}{\succRealCleanStd}         &                                &                                         &                              & \\
    Direct screenshot training, original 4B reader & --                  & \errGateOrig             & \msd{\succRealDirect}{\succRealDirectStd}       &                                &                                         &                              & \\
    \addlinespace[2pt]
    \tabblock{8}{Clean-text policy with an aligned reader} \\
    \rowcolor{rowblue}4B, alignment seed 1 (main) & \alignTrainShots & \errGateAligned     & \msd{\succAlignClean}{\succAlignCleanStd}       & \vup{\dMain}$^\ast$            & \ci{\dMainLo}{\dMainHi}                 & --                           & \\
    4B, alignment seed 2           & \alignTrainShots     & \errGateAlignSeedB       & \msd{\succAlignSeedB}{\succAlignSeedBStd}       & \vup{\dSeedBDirect}            &                                         & \vgray{\dAlignSeedB}         & \\
    4B, alignment seed 3           & \alignTrainShots     & \errGateAlignSeedC       & \msd{\slSeedThreeSucc}{\slSeedThreeSuccStd}     & \vup{\slSeedThreeVsDirect}     &                                         & \vdown{\slSeedThreeVsAlign}  & \\
    8B                             & \slLargeTrainUsed   & \slLargeGate             & \msd{\slLargeSucc}{\slLargeSuccStd}             & \vup{\slLargeVsDirect}$^\ast$  & \ci{\slLargeVsDirectLo}{\slLargeVsDirectHi} & \vgray{\slLargeVsAlign}$^\ast$ & \ci{\slLargeVsAlignLo}{\slLargeVsAlignHi} \\
    4B, without exact repeats      & \slDedupTrainUsed    & \slDedupGate             & \msd{\slDedupSucc}{\slDedupSuccStd}             & \vgray{\slDedupVsDirect}$^\ast$ & \ci{\slDedupVsDirectLo}{\slDedupVsDirectHi} & \vdown{\slDedupVsAlign} & \\
    \bottomrule
  \end{tabularx}
\end{table*}

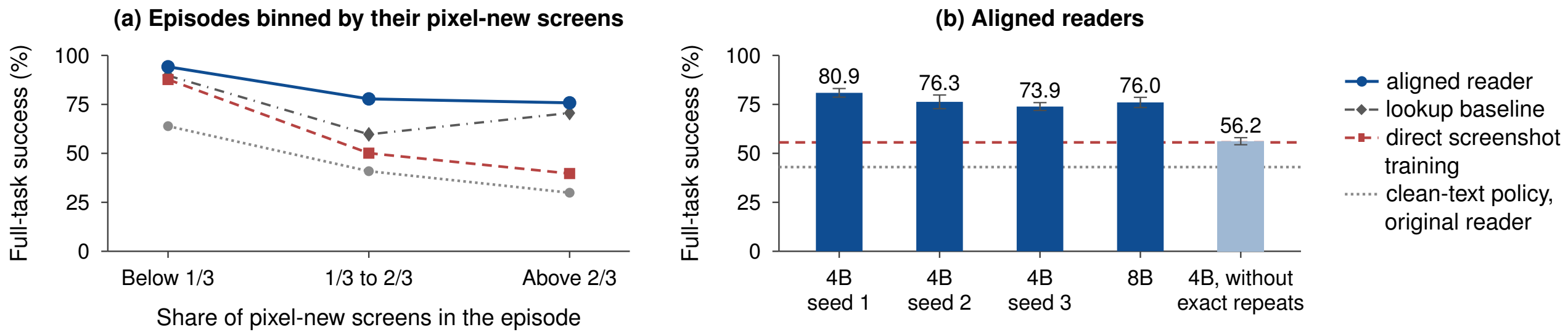
\begin{figure*}[!tp]
  \centering
  \begin{tikzpicture}
\tikzset{flabel/.style={font=\sffamily\fontsize{6.5}{7.5}\selectfont, inner sep=1.2pt}}
\pgfplotsset{ftitle/.style={title style={font=\sffamily\bfseries\fontsize{7}{8.2}\selectfont, yshift=-2pt, align=center}}}
\begin{axis}[paperplot, ftitle, name=fa, width=5.6cm, height=2.1cm, ymin=0, ymax=100, ytick={0,25,50,75,100},
  xmin=-0.3, xmax=2.3, xtick={0,1,2}, xticklabel style={align=center},
  xticklabels={{Below 1/3}, {1/3 to 2/3}, {Above 2/3}},
  xlabel={Share of pixel-new screens in the episode}, ylabel={Full-task success (\%)},
  title={(a) Episodes binned by their pixel-new screens}]
  \addplot[refgray, densely dotted, line width=0.8pt, mark=*, mark size=1.1pt, mark options={solid}]
    coordinates {(0,\slPixLowCleanOrig) (1,\slPixMidCleanOrig) (2,\slPixHighCleanOrig)};
  \addplot[black!65, dash dot, line width=0.7pt, mark=diamond*, mark size=1.7pt, mark options={solid}]
    coordinates {(0,\slPixLowLookup) (1,\slPixMidLookup) (2,\slPixHighLookup)};
  \addplot[basered, densely dashed, line width=0.8pt, mark=square*, mark size=1.3pt, mark options={solid}]
    coordinates {(0,\slPixLowDirect) (1,\slPixMidDirect) (2,\slPixHighDirect)};
  \addplot[ourblue, line width=0.9pt, mark=*, mark size=1.6pt]
    coordinates {(0,\slPixLowAlign) (1,\slPixMidAlign) (2,\slPixHighAlign)};
\end{axis}
\begin{axis}[paperplot, ftitle, name=fb, at={($(fa.east)+(1.6cm,0)$)}, anchor=west, width=5.6cm, height=2.1cm,
  ymin=0, ymax=100, ytick={0,25,50,75,100}, xmin=-0.6, xmax=4.6, xtick={0,1,2,3,4}, xticklabel style={align=center},
  xticklabels={{4B\\seed 1}, {4B\\seed 2}, {4B\\seed 3}, {8B}, {4B, without\\exact repeats}},
  ylabel={Full-task success (\%)}, title={(b) Aligned readers}]
  \draw[basered, densely dashed, line width=0.8pt] (axis cs:-0.6,\succRealDirect) -- (axis cs:4.6,\succRealDirect);
  \draw[refgray, densely dotted, line width=0.8pt] (axis cs:-0.6,\succRealClean) -- (axis cs:4.6,\succRealClean);
  \addplot[ybar, bar width=0.5cm, bar shift=0pt, fill=ourblue, draw=none, error bars/.cd, y dir=both, y explicit,
    error bar style={black!70, line width=0.45pt}]
    coordinates {(0,\succAlignClean) +- (0,\succAlignCleanStd) (1,\succAlignSeedB) +- (0,\succAlignSeedBStd)
                 (2,\slSeedThreeSucc) +- (0,\slSeedThreeSuccStd) (3,\slLargeSucc) +- (0,\slLargeSuccStd)};
  \addplot[ybar, bar width=0.5cm, bar shift=0pt, fill=ourblue!40, draw=none, error bars/.cd, y dir=both, y explicit,
    error bar style={black!70, line width=0.45pt}]
    coordinates {(4,\slDedupSucc) +- (0,\slDedupSuccStd)};
  \node[flabel, anchor=south] at (axis cs:0,\succAlignClean+\succAlignCleanStd) {\succAlignClean};
  \node[flabel, anchor=south] at (axis cs:1,\succAlignSeedB+\succAlignSeedBStd) {\succAlignSeedB};
  \node[flabel, anchor=south] at (axis cs:2,\slSeedThreeSucc+\slSeedThreeSuccStd) {\slSeedThreeSucc};
  \node[flabel, anchor=south] at (axis cs:3,\slLargeSucc+\slLargeSuccStd) {\slLargeSucc};
  \node[flabel, anchor=south] at (axis cs:4,\slDedupSucc+\slDedupSuccStd) {\slDedupSucc};
\end{axis}
\node[anchor=west, font=\sffamily\fontsize{7}{8.6}\selectfont, inner sep=0pt, align=left] at ($(fb.east)+(0.45cm,0)$) {\tikz[baseline=-0.6ex]{\draw[ourblue, line width=0.9pt] (0,0) -- (0.36,0); \fill[ourblue] (0.18,0) circle[radius=1.6pt];}~aligned reader\\
  \tikz[baseline=-0.6ex]{\draw[black!65, dash dot, line width=0.7pt] (0,0) -- (0.36,0); \fill[black!65] (0.18,0) +(0,1.7pt) -- +(1.7pt,0) -- +(0,-1.7pt) -- +(-1.7pt,0) -- cycle;}~lookup baseline\\
  \tikz[baseline=-0.6ex]{\draw[basered, densely dashed, line width=0.8pt] (0,0) -- (0.36,0); \fill[basered] (0.15,-0.045) rectangle (0.21,0.045);}~direct screenshot\\\hphantom{\tikz{\draw (0,0) -- (0.36,0);}~}training\\
  \tikz[baseline=-0.6ex]{\draw[refgray, densely dotted, line width=0.8pt] (0,0) -- (0.36,0);}~clean-text policy,\\\hphantom{\tikz{\draw (0,0) -- (0.36,0);}~}original reader};
\end{tikzpicture}   \caption{\textbf{Full-task success by the share of pixel-new screens and by aligned reader.} (a)~Four runs on the same instances and training seeds, binned by the share of pixel-new screens in the aligned reader's episodes (descriptive; \cref{tab:supp_bins}). (b)~The clean-text policy with the aligned readers of \cref{tab:supp_more_readers} (light blue: without exact repeats), mean\,$\pm$\,std over its two training seeds.}
  \label{fig:supp_further}
\end{figure*}

\paragraph{Episodes with mostly pixel-new screens.}
In episodes where more than two thirds of the screens are pixel-new, alignment is still clearly higher than direct screenshot training ($\slPixHighGap$ points; \cref{tab:supp_bins}). We bin the episodes of the aligned reader's run by their share of pixel-new screens and compare four runs on the same instances and training seeds.
In this run, \slPixNewPct\% of the screens are pixel-new and \newScreenListPct\% have a text view that appears in no training label (\cref{supp:reader}), so most pixel-new screens also differ in content.
Screens are matched as JPEG images, the only form that run stored. On the lookup baseline's run, this matching judged \slPixCheckMissed{} of the \slPixCheckShots{} screens pixel-new although they exactly repeat a calibration screenshot, and none the other way (\cref{supp:reader}). The rule for this analysis, written after that check and before the binning, accepts this error, which can only raise an episode's share.
As the share grows, direct screenshot training falls from \slPixLowDirect\% to \slPixHighDirect\% and alignment from \slPixLowAlign\% to \slPixHighAlign\% (\cref{fig:supp_further}a).
The comparison is descriptive, because the bins come from one run and differ in template mix and in difficulty; the clean-text policy with the original reader, which used no calibration screenshot, falls as well.
Runs also differ in the screens they reach: the lookup baseline's run has \newScreenPixelPct\% pixel-new screens, and with its own bins the baseline reaches \lookupBinHighSucc\% in the top bin (\cref{tab:supp_newscreens}) instead of \slPixHighLookup\%.

\paragraph{Alignment without exact repeats.}
Aligned only on the calibration screenshots that appear in no earlier run on the confirmation set, a reader shows no measurable difference from direct screenshot training ($\slDedupVsDirect$ points; \cref{tab:supp_more_readers}). Matching JPEG versions against the screenshots all these runs stored finds \slDedupRemoved{} of the \slDedupRows, because MobileGym renders a screen identically across instances. We align on the remaining \slDedupKept{} without adding new screenshots, so the test also shrinks the alignment set to about a sixth and cannot separate the missing repeats from the smaller set. In the run with this reader, which stored lossless screenshots, exact matching finds that \slDedupResidKeptPct\% of its \slDedupResidShots{} screens repeat one of its alignment screenshots (\slDedupResidAllPct\% repeat one of the full calibration set).
The clean-text policy reaches \slDedupSucc\% with this reader; full alignment is clearly higher, by \magn{\slDedupVsAlign} points, and the comparison with the lookup baseline points toward the baseline ($\slDedupVsLookup$ points; both descriptive).
After seeing the results, we compared it with the readers of \cref{fig:data_efficiency}a, aligned on the first calibration screenshots without removal. On held-out screenshots that repeat none of their alignment screenshots, it errs on \slDedupGateNew\% of elements and the reader aligned on the first \curveShotsTwoFifty{} on \curveNewErrTwoFifty\%, so it reads about as well as a reader aligned on a set of similar size. 

\paragraph{A third alignment seed.}
Over three alignment seeds of the 4B reader, success ranges from \slSeedThreeSucc\% to \succAlignClean\% (mean \slSeedMeanThree\%), against \succRealDirect\% for direct screenshot training, and each seed is clearly higher than direct training (\cref{tab:supp_more_readers}). The main paper uses alignment seed 1, which was aligned and evaluated before the other two were aligned; it is the highest of the three and clearly higher than seed 3, by \magn{\slSeedThreeVsAlign} points. 

\paragraph{An aligned 8B reader.}
As a control for reader capacity, we align the 8B reader of \cref{sec:scope}, on half-size screenshots, with the same recipe and calibration screenshots. With it the clean-text policy reaches \slLargeSucc\%, against \slSeedThreeSucc\% to \succAlignClean\% for the three 4B alignment seeds, with no measurable difference from seed 1 ($\slLargeVsAlign$ points); it is clearly higher than direct screenshot training, which uses the original 4B reader ($\slLargeVsDirect$ points). Alignment lowers its element error rate on the held-out screenshots from \errGateLarge\% to \slLargeGate\%, about that of the aligned 4B reader, and raises its task success from \succLargeClean\% to \slLargeSucc\%. 

\paragraph{All remaining failures.}
Over all \slFailCount{} failed episodes of the clean-text policy with the aligned reader, not only those of the four templates of \cref{tab:supp_failures}, the pre-specified procedure of \cref{supp:per_template} splits the failures evenly between the reader's output and the policy. In \slFailReaderSide{} of them the first discrepancy lies in the reader's output: in \slFailActed{} the policy acts on an element that the reader misread or invented, in \slFailNeeded{} an element that the policy used in the successful text episode is misread or missed, and in \slFailOrder{} the order differs.
In the other \slFailPolicy, the policy acts on no misread or invented element and also fails the same instance in text, so the procedure attributes the failure to the policy.

\paragraph{Cost.}
The runs of this section made \slReadsTotal{} reader calls, counted apart from \cref{tab:supp_accounting} (8B reader: \slLargeGateReads{} on held-out screenshots and \slLargeReads{} in task runs; without exact repeats: \slDedupGateReads{} and \slDedupReads; alignment seed 3: \slSeedThreeReads{} in task runs), and took \slLargeCardHours, \slDedupCardHours{} and \slSeedThreeCardHours{} GPU-hours, including the alignment of the two new readers.

\end{document}